\documentclass[manuscript]{acmart}
\AtBeginDocument{%
  }

\setcopyright{cc}
\setcctype{by}
\acmJournal{TIST}
\acmYear{2026} \acmVolume{1} \acmNumber{1} \acmArticle{}
\acmMonth{1} \acmDOI{10.1145/3844612}
\acmConference[ACM Transactions on Intelligent Systems and Technology]{}{}{}

\usepackage{mathtools} 
\usepackage[normalem]{ulem} 
\usepackage{forest}
\usepackage{adjustbox}

\forestset{
  parse/.style={
    for tree={
      inner sep=3pt,
      s sep=11pt,
      l sep=12pt,
      align=center
    }
  }
}
\begin{document}

\title{Three tiers of computation in transformers and in brain architectures}


\author{Emma Graham}
\authornote{Both authors contributed equally to this research.}
\affiliation{%
  \institution{Thayer School of Engineering, Dartmouth College}
  \city{Hanover}
  \state{NH}
  \country{USA}}
\author{Richard Granger}
\authornotemark[1]
\email{richard.granger@gmail.com}
\affiliation{%
  \institution{Dartmouth College}
  \city{Hanover}
  \state{NH}
  \country{USA}
}

\renewcommand{\shortauthors}{Graham and Granger}

\begin{abstract}
Enabling AI systems to exhibit both fluent natural language and reliable logical reasoning is of great interest, yet the underlying machinery necessary for these capabilities remains poorly understood. Increases in abilities arise from many factors (scaling, architecture, training) yet the core required computational abilities for any particular task still elude understanding. We show that it is the transition between computational tiers in the automata hierarchy from the Theory of Computation, rather than scaling or particular algorithms, that determines a system's capabilities. We provide a rigorous theoretical treatment, and empirical evaluations of both language models and human subjects, culminating in a theoretically grounded pragmatic benchmarking method, that can be used to ascertain the computational tier of an artificial or biological system.  Specifically, we identify the tiers for the transition from prelinguistic to fluent natural language, and the tiers for the subsequent transition from articulate language use to reliable logical reasoning. The resulting principled analyses provide explanatory accounts of the abilities and shortfalls of a range of language models, and suggest actionable insights into the expansion of their logic and reasoning abilities.

\end{abstract}

\ccsdesc[500]{Theory of computation~Formal languages and automata theory}
\ccsdesc[300]{Applied computing~Biological networks}
\ccsdesc[300]{Theory of computation~Models of learning}
\ccsdesc[500]{Computing methodologies~Natural language processing}

\keywords{transformers, large language models, Theory of Computation, logical reasoning, thalamocortical circuitry, automata hierarchy, formal grammars, allometry. 
  }

\maketitle

\section{Introduction}

Since language models abruptly achieved the fluency that the field of natural language processing had sought for more than 70 years \cite{jonesNaturalLanguageProcessing1994}, there now are two entities, biological and artificial, that can exhibit articulate, versatile, and far-ranging natural language use. 
What similarities and differences underlie the mechanisms of these  language foundation model (LFM) abilities and human abilities? 
Can understanding human processing assist in understanding LFM capacities and limitations? 

Notably, baseline LFM systems are highly challenged by higher logical reasoning, in contrast with their expert language skills. Humans also show profound differences in language and commonsense inference, on one hand, versus formal logic and rigorous reasoning, on the other; are these language-vs.-reasoning distinctions coincidental, or related?

We consider these capability differences in terms of hierarchical {\textit {transitions}}, from more restricted up to more expansive capabilities: what must be added to a system to formally enlarge the sets of tasks it can perform?  Early pre-transformer AI systems, although highly useful for many applied tasks, nonetheless exhibited extremely limited language abilities \cite{wuGooglesNeuralMachine2016}; the introduction of transformer architectures \cite{vaswani2017} suddenly and unexpectedly launched fluent and articulate language use \cite{leveque2012WinogradChallenge}.  By contrast, transformer-based LFMs are highly challenged by tasks requiring reliable reasoning, mathematical, or formal logic-based tasks \cite{AppleGSM, huang2024largelanguagemodelsnotreason, McCoy2024llmsshapedbyproblem}, i.e., careful adherence to facts as opposed to arbitrary inference (see, e.g., \cite{raoLargeLanguageModel2026, tillerGenerativeArtificialIntelligencedriven2026, beanReliabilityLLMsMedical2026, ramaswamyChatGPTHealthPerformance2026}). 
Evidence indicates that these more advanced tasks require qualitative architectural changes: augmentations that incorporate distinct new computational mechanisms beyond those of transformers  alone 
(e.g., \cite{llama3.1, claude3, gpt4}).  
 
\begin{figure}[t]
\centering
\includegraphics[width=0.85\textwidth]{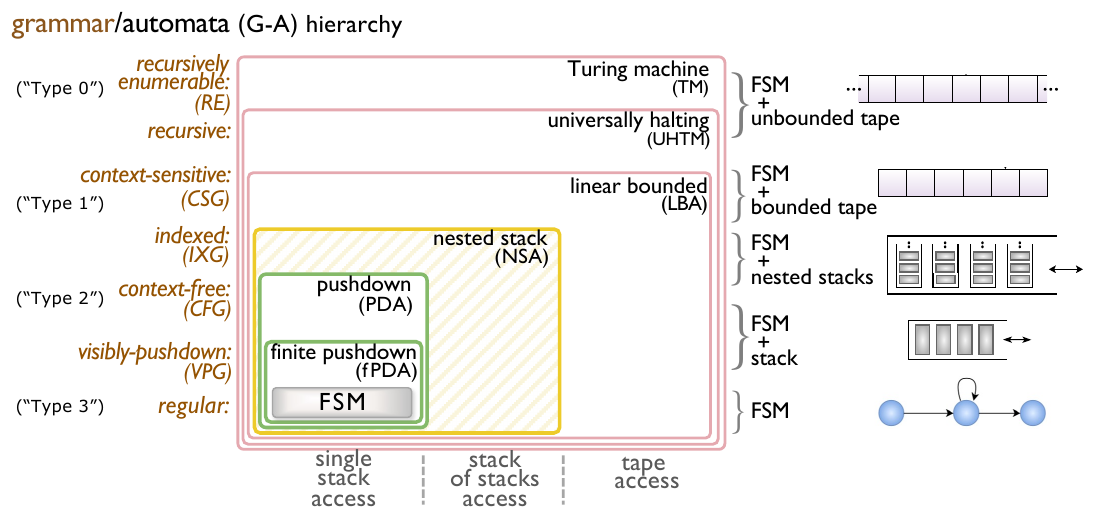}
\caption{The grammar-automata (G-A) hierarchy  pairs classes of machines (black text) with grammars (brown) in order of increasing computational power, from finite state machines (FSM) to Turing machines (TM) \cite{hopcroft-ullman, hopcroftIntroductionAutomataTheory2007}. 
Mechanisms containing only single-stack memory access  (green) can be distinguished from automata with nested stack access (yellow), and from automata with tape access (equivalent to two or more independently accessible stacks) (red). We focus on computational distinctions among these three tiers.
\Description{The grammar-automata hierarchy from the Theory of Computation.}
}
\label{fig:GAhierarchy}
\end{figure}

We study these language and logic abilities with regard to specific transitions in the 
foundational \textit{grammar-automata (G-A) hierarchy} from the Theory of Computation \cite{hopcroftIntroductionAutomataTheory2007}. This well-studied hierarchy of abstract machines (see Figure \ref{fig:GAhierarchy}) formally characterizes the computational powers of the range of possible automata, from least (finite state machines (FSM)) to most powerful (Turing machines (TM)).
  
In light of this formal hierarchy, we ask what computational abilities specifically are introduced by transformers, that enable current AI language abilities; and what is further added in current large AI systems enabling them to surpass their language-only inference, achieving reliable logical reasoning abilities?

Although it was initially conjectured that scaling alone may confer new computational abilities \cite{kaplan2020scalinglawsneurallanguage}, even the largest LFMs that still retain the traditional transformer architecture remain highly challenged by many tasks \cite{AppleGSM, huang2024largelanguagemodelsnotreason, McCoy2024llmsshapedbyproblem, raoLargeLanguageModel2026, tillerGenerativeArtificialIntelligencedriven2026, beanReliabilityLLMsMedical2026, ramaswamyChatGPTHealthPerformance2026}. 
Extant architectures that do enable more robust performance on such tasks do so by incorporating qualitative changes to the foundational transformer architectures of standard large language models, requiring, at minimum, augmented external memories and new methods for how those memories are accessed (see, e.g., \cite{Nguyen2024, jumperHighlyAccurateProtein2021}).  
(Some are referred to as combined ``neuro-symbolic'' systems \cite{mozerTopologicalTroubleTransformers2026, belleFutureNeuroSymbolicWhere2026, garcezNeurosymbolicAI3rd2023}.) 
Such architectural augmentations will be discussed in both empirical and theoretical terms. 
 
As a key to understanding these transitions,  it is notable that humans exhibit two curiously similar transitions: from nonhuman communications systems to human language, and then from (ubiquitous) language-based ``common-sense'' inference to the separate (and far rarer) abilities of rigorous logical reasoning and mathematics.   
Humans and AIs differ in many fundamental respects, and these observed correspondences might have turned out to be spurious or coincidental; but we provide both empirical and theoretic evidence that the correspondences in fact share deep formal computational links, and that these biological transitions actually provide  illuminating hints to further understanding of the corresponding transitions in AI. 
The substantial discrepancies between natural language use and logical reasoning  continue to be heavily studied, yet remain poorly understood. 
It is intriguing that there appears to be no fixed terminology for the divide between statistical pattern-matching-based inference on one hand vs. rigorous reasoning on the other; the resulting confusion has led to countless misunderstandings in the literature \cite{AppleGSM, nikankin2024arithmeticalgorithmslanguagemodels, jiang2024llmnotgenuinereasoner, shi2023llmseasilydistracted, schaeffer2023emergentabilitieslargelanguage, kassner-schutze-2020-negated, li2022largelanguagemodelscontrollable, jones2022capturingfailureslargelanguage, misra2023compsconceptualminimalpair, chaves-richter-2021-look}.
Going forward, we identify the computational nature of the crucial architectural changes in advanced systems, which we refer to as \textit{augmented traditional transformers} (ATT), (see Section 2.3.2).  We provide evidence that these additional architectural mechanisms may be responsible for the growing development of reasoning abilities in language foundation models beyond those of base (decoder) transformers. 
 
We specifically propose to align particular abilities of both AI systems and natural biological systems, with two key transitions in the formal hierarchy from the theory of computation. The first transition, from the context-free grammar (CFG) tier to the indexed grammar (IXG) tier, is linked to the corresponding transition from limited to fluent language use; the second transition, from IXG to the context-sensitive grammar (CSG) tier, corresponds to the transition from limited to robust logical reasoning abilities.

\textit{Section 2} establishes the conceptual framework: the three major computational tiers of the G-A hierarchy; the distinct memory mechanisms that separate the automata tiers; and evidence showing that rules of human natural language reside in the indexed tier, whereas reliable logic and reasoning require higher tiers. 

\textit{Section 3} introduces a new generative benchmarking method that uses formal grammars at particular tiers as symbolic recursive generators to produce test strings incorporating natural, synthetic, and arithmetic language tasks. 
 
\textit{Section 4} applies this novel benchmarking method, reporting statistically significant empirical evaluations of fifteen models, and of human subjects, on  sequences generated from grammars of three fine-grained tiers, specifically at  transitions where evidence suggests human  language computation is segregated from that of higher logical processing tasks (see Figure \ref{fig:empircal}).
 
\textit{Section 5} provides both computational and neurobiological evidence relating the underlying computations of these three tiers to mechanisms in traditional transformer-based LFMs, to distinct ATT (augmented) architectures, and to the differential scaling of biological brain circuit architectures. 

Taken together, the framework and empirical results
establish a coherent foundation for the systematic interpretation of a wide expanse of behaviors  identifiable across tiers of the G-A hierarchy, identifying candidate explanatory correspondences between the computations of large language models and of human brain circuitry.

\section{Tiers of computational capabilities}

Although language fluency and template-matching abilities exhibit impressively powerful and persuasive common sense inference, these abilities do not substitute for reliable logical processing or robust reasoning  \cite{benderClimbingNLUMeaning2020}. 
Much current work focuses on tests to ascertain the presence or absence of these more comprehensive reasoning capacities in existing systems.
However, this often takes the form of sets of narrowly specialized tests, which can be misleading when it comes to understanding broader system capabilities \cite{rajiAIEverythingWhole2021}. Current benchmarks still struggle to assess language models; this shortfall is sometimes referred to as ``the benchmark crisis'' \cite{mahmoodBenchmarkingCrisisBiomedical2025, rajiAIEverythingWhole2021, reuelBetterBenchAssessingAI2024, banerjeeVulnerabilityLanguageModel2024, erikssonCanWeTrust2025}.

Some standard benchmarks (e.g.,  \cite{wangGLUEMultiTaskBenchmark2019,wangSuperGLUEStickierBenchmark2020}), evaluate general natural language understanding, but may conflate linguistic competence with commonsense inference
\cite{rajiAIEverythingWhole2021}, 
making it challenging to isolate measures of the models' linguistic capabilities, as those benchmarks may confound syntax with broader reasoning tasks. 
Other benchmarks target abstract reasoning or problem-solving. The Abstraction and Reasoning Corpus (ARC) \cite{cholletMeasureIntelligence2019} presents grid-based puzzles to assess multi-step problem solving. 
It evaluates pattern recognition and solution planning, but is limited in its effectiveness to directly evaluate underlying computational ability. Such benchmarks may provide valuable insights into certain capabilities, but do not directly measure a model's inherent computational class, in terms of the mathematical classes of abstract machines. They complement but do not replace formal analyses. To gain deeper insight into what language  models can compute, we turn to the rigorous tools of formal languages and the theory of computation.

\subsection{Theory of Computation: formal grammars and automata}

The expressivity of language models, examined from the formal language perspective (predominantly via automata or circuit complexity analyses), 
\cite{ackermanSurveyNeuralNetworks2020, merrillSequentialNeuralNetworks2019, stroblWhatFormalLanguages2024}  has examined what models are able to do 
\cite{siegelmannComputationalPowerNeural1995, weissPracticalComputationalPower2018, suzgunLSTMNetworksCan2019, perezTuringCompletenessModern2019, ackermanSurveyNeuralNetworks2020, bhattamishraAbilityLimitationsTransformers2020, bhattamishraComputationalPowerTransformers2020, merrillSequentialNeuralNetworks2019, merrillLinguisticCapacityRealTime2021, yaoSelfAttentionNetworksCan2021, JMLR:v22:20-302, chiangOvercomingTheoreticalLimitation2022} and what models are not able to do 
\cite{hahnTheoreticalLimitationsSelfAttention2020, haoFormalLanguageRecognition2022, merrillSaturatedTransformersAre2022, merrillParallelismTradeoffLimitations2023, stroblAverageHardAttentionTransformers2023}, i.e., computational lower and upper bounds.  

Automata and formal grammars
can be used as an abstract computational model for sequential data processing, providing an analytical framework for assessing the expressive power of a system \cite{sarrofExpressiveCapacityState2024}. This lens has offered a principled approach used to determine the (in)abilities of transformers (e.g., \cite{perezTuringCompletenessModern2019, hahnTheoreticalLimitationsSelfAttention2020, bhattamishraComputationalPowerTransformers2020, yaoSelfAttentionNetworksCan2021, ackermanSurveyNeuralNetworks2020, liu2023transformerslearnshortcutsautomata, liuExposingAttentionGlitches2023, deletangNeuralNetworksChomsky2023, stroblWhatFormalLanguages2024, sanfordRepresentationalStrengthsLimitations2023, pengLimitationsTransformerArchitecture2024}), and state-space models (e.g., \cite{siegelmannComputationalPowerNeural1992, horneBoundsComplexityRecurrent1993, indykOptimalSimulationAutomata1995, hahnVisiblyCounterLanguages2015, weissPracticalComputationalPower2018}). 

There are theoretical guarantees on one hand, and  empirical realities on another. 
Empirical analyses of the expressivity of massive pre-trained language models are sparse yet striking. Crafted sets of synthetic and arithmetic task evaluations of LFMs from the formal grammar perspective are found in \cite{daveInvestigatingSymbolicCapabilities2024}, \cite{deletangNeuralNetworksChomsky2023}, and others, together repeatedly indicating that tasks that require linear bounded automata or more are not reliably solved by LFMs.  
Beyond these partial, example-specific tests, 
proposed benchmarks of computational capability are either not generative \cite{zahraeiTuringQBenchmarkingAI2024} or have inherent verification issues of correctness and task difficulty levels  \cite{fanNPHardEvalDynamicBenchmark2024}, limiting the benchmarks' assurance of separate training and testing data.  
We make use of the computational equivalence between classes of formal grammars and automata to bridge this gap.
 
Grounded in the theory of computation, formal grammars can produce sets of strings that are situated within specific tiers in computational hierarchies (e.g.,  \cite{jagerFormalLanguageTheory2012});  (Figure \ref{fig:GAhierarchy}). 
Formal grammars are fundamental to (neuro)symbolic artificial intelligence (AI), and their symbolic relations are  often expressed recursively. For a system to \textit{reliably} recognize the strings from a given tier, it must have the computational power of that tier.
The computational tiers are \emph{equivalence classes} 
in terms of their computational power: classes of algorithms (distinct architectures) that can parse (automata) or generate (grammars) a specific tier or level of formal languages.

\subsection{A note on stack-based memory classes}

The IXG class is situated neither at the bottom nor top of the G-A
hierarchy, raising the natural question of why human languages are repeatedly shown to empirically reside in this seemingly arbitrary middle tier (see discussion in Section 2.3.1).
All systems in the lower sub-hierarchy (Figure \ref{fig:GAhierarchy}) are limited to one memory stack that may in turn push objects or (nested) stacks \cite{Aho1968indexed, hopcroft-ullman, maslov1974, JOSHI1975}. By contrast, all more-powerful (linear bounded automata (LBA) to TM) systems require access to a tape memory structure, which is computationally equivalent to configurations of multiple accessible stack structures \cite{hopcroftIntroductionAutomataTheory2007}. 
 
Climbing tiers in the hierarchy requires these distinctly different memory mechanisms. The shifts, from single-stack to stack-of-stacks to tape memory access, separates the hierarchies into sub-hierarchies containing pushdown automata (green), the mid-tier of nested stack automata (yellow), and the more powerful classes containing Turing machines (red), in Figure \ref{fig:GAhierarchy}. Specifically, nested stack automata -- a single stack-based memory system wherein the stack itself has access to nested stacks -- are the most powerful systems that can be constructed without a tape memory structure. The nested stack automata tier can therefore be seen as a natural ceiling of computational power within the G-A hierarchy.

\subsection{Transitions among three tiers}

Any given tier of the hierarchy demarcates computational capabilities that strictly exceed, and encompass, those of all lower tiers. Some complex computational processes cross some of these tier boundaries; it is not always possible to circumscribe a particular system’s capabilities in terms of specific tiers of the G-A hierarchy, and indeed other candidate formal hierarchies are also studied \cite{stroblWhatFormalLanguages2024, yang2024masked}. Nonetheless this computational hierarchy is highly applicable to a broad range of formal and real-world tasks in computation, finance, biology, and many other fields \cite{Datta2013, Zhang2022, Nehaniv2000, Duenas-Diez2019, Sassaman2013, Flood2015}.

\subsubsection{Human abilities, from language to higher logic}

Notably, all humans (barring severe neurological deficits) acquire their native language, even without formal training: natural language acquisition is universal and seemingly (perhaps deceptively) effortless. Humans can generate unlimited meaningful utterances from a fixed vocabulary, whereas other animals produce reduced conceptual repertoires \cite{berwickSongsSyntaxLinguistics2011, berwickEvolutionBrainNature2013, beckersWhatAnimalsLearn2017}; even rich and informative animal vocalizations typically denote relatively circumscribed sets of possible descriptions, opportunities, or risks (e.g., a particular type of food / predator / obstruction / conspecific is nearby) whereas humans can convey entirely novel messages (``the key is hanging behind the red panel just past the kitchen door''). Syntax provides the means for codifying finite sets of words into potentially infinite distinct expressions \cite{Fitch_2010, Scott-Phillips2014, Berwick2017}.

Formal grammars can be designed to include  specific syntactic rules, cases, and dependencies found in a human natural language, 
such as subject-verb agreements or past tense phrasal verb rules, thus generating strings that, while nonsensical, adhere to the syntactic structures that require a specific tier of computational power for reliable recognition.

Decades of human language analyses have endeavored to ascertain the computational class of syntactic dependencies in natural languages in terms of the grammar-automata hierarchy \cite{Chomsky1975, ahoTheoryParsingTranslation1972, Joshi1990TheCO, Steedman1987, weirCombinatoryCategorialGrammars1988}.  Context-free grammars have been found insufficient for the processing of many dependencies that occur in natural languages.  This led to the development of new formal grammar definitions that can express these natural language dependencies, including tree-adjoining grammars \cite{Joshi1997}, combinatory categorial grammars \cite{weirCombinatoryCategorialGrammars1988}, and head grammars \cite{Pollard1984}.  Moreover, it has been shown that these various formulations of grammars, all empirically found to correspond with natural human language utterances, all reside within the encompassing indexed-grammar tier of the G-A hierarchy, supra-context-free and sub-context-sensitive (\cite{VijayShanker1994}; see also \citet{ahoNestedStackAutomata1969}).  Meanwhile, no syntactic dependencies in any human natural language have been found to require supra-IXG tiers.

Conversely, extensive studies of corpora of nonhuman communication (calls of nonhuman primates, cetaceans, avians, etc.), have thus far found none classified above the context-free tier \cite{suzukiInformationEntropyHumpback2006, berwickSongsSyntaxLinguistics2011, stobbeVisualArtificialGrammar2012, fitchComputationalConstraintsSyntactic2004, fitchArtificialGrammarLearning2012, gentnerRecursiveSyntacticPattern2006, hermanComprehensionSentencesBottlenosed1984, vanheijningenSimpleRulesCan2009, kershenbaumAcousticSequencesNonhuman2016, kershenbaumAnimalVocalSequences2014, lefebvreBrainsInnovationsEvolution2004, berwickEvolutionBrainNature2013, beckersWhatAnimalsLearn2017}.  
 
This unexpectedly powerful confluence between empirical natural language abilities on one hand, and formal computation on the other, allows these human abilities to be analyzed in terms of the specific indexed tier, above  context-free and below  context-sensitive, suggesting that the mechanisms for natural language processing may 
reside within the indexed tier.  Consequently,  indexed grammars are the computational  focus in this work, central  among the three described tiers. Analyzing empirical  language abilities in terms of computations within this tier, enables principled evaluations of the abilities and limits of mechanisms  of natural language processing. 
 
By contrast with the ease of acquisition and evaluation of natural language, humans do not naturally or effortlessly acquire
logic and reasoning methods necessary for fields ranging from arithmetic to engineering to propositional logic to scientific pursuits.
Many advanced tasks require computational classes of higher tiers.  Examples abound, 
including most forms of engineering and math;
a few well studied illustrative examples include DNA Computing \cite{karlDNAComputingArrival1997}, $\lambda$-calculus \cite{TuringComputabilityLdefinability1937}, square root computation \cite{deletangNeuralNetworksChomsky2023}, and decimal arithmetic \cite{daveInvestigatingSymbolicCapabilities2024}.

There are, of course, simpler forms of inference that do not use or depend on logical steps. 
``Common sense'' inference can refer to simple statistical pattern matching (if I drive a car, I  sit in a seat and handle a steering wheel), without any need for logical reasoning; typicality need not equate to necessary or sufficient conditions. 
For humans to use advanced logical reasoning beyond ``common sense'' inference, they require extensive specialized training, as opposed to tasks of natural language acquisition or statistical pattern-matching that appear with surprising ease in humans.

Notably, logic or math may be impossible for an unaided human, yet can become trivial to someone trained in the specific use of external memory (even something as deceptively simple as particular uses of paper and pencil): multiplying two eight-digit numbers in your head will be untrustworthy and error-prone, but straightforward (perhaps trivial) if performed by following an appropriate sequence of carefully learned rules on paper. 
Many logical reasoning tasks require both external memory, and extremely specific learned rules for using that memory.

\subsubsection{AI systems, from language to higher logic}

Before transformer-based natural language processing, language models displayed strikingly limited language abilities (although useful for specific tasks; e.g., \cite{wuGooglesNeuralMachine2016}), but the introduction of the architecture of transformer-based LFMs abruptly launched fluent and communicative language use \cite{devlin2019BERT, vaswani2017, tenney2019bertrediscoversclassicalnlp}.
Transformer-based LFMs still nonetheless struggle with mathematics and formal logic \cite{AppleGSM, nikankin2024arithmeticalgorithmslanguagemodels, jiang2024llmnotgenuinereasoner, shi2023llmseasilydistracted, schaeffer2023emergentabilitieslargelanguage, kassner-schutze-2020-negated, li2022largelanguagemodelscontrollable, jones2022capturingfailureslargelanguage, misra2023compsconceptualminimalpair, chaves-richter-2021-look,kambhampatiCanLargeLanguage2024, kambhampatiLLMsCantPlan2024}, and it is posited that corresponding training and specific augmentations, not simply scaling the model parameters of current architectures, are required to overcome these limitations (e.g., \cite{yang2024masked, ahuja2024, wei2022emergentabilitieslargelanguage, Kilpatrick_BasuMallick_Kofman_2024, openai-reasoning-post}).
In this light it is notable that many new large language model architectures indeed appear to incorporate augmentations of the base transformer systems.  

Qualitative changes to traditional transformer architectures include incorporation of mixture-of-experts \cite{loCloserLookMixtureofExperts2025}, chain-of-thought steps \cite{wei2023chainofthoughtpromptingelicitsreasoning, chenReasoningEraSurvey2025}, and memory-augmentations \cite{wuMemformerMemoryAugmentedTransformer2022}.
Such mechanisms are not simply appended, they are intercalated into the transformer architecture, resulting in a new type of transformer-variant architectures which we  refer to as augmented traditional transformers (ATT). 
We conjecture that current frontier models are instances of the ATT architectures that we describe here \cite{IntroducingGPT552026, IntroducingClaudeOpus, Gemini31Pro}. 
After initial claims that scaling alone may enable logic-based reasoning, system developers appear to have taken to heart the need for augmentation rather than solely scaling \cite{merrillIllusionStateStateSpace2024, mozerTopologicalTroubleTransformers2026, belleFutureNeuroSymbolicWhere2026, garcezNeurosymbolicAI3rd2023}.

Related work further shows that without augmentations, LFMs suffer from brittleness (e.g., inability to adapt to small deviations from memorized templates \cite{tamLetMeSpeak2024}; debilitating sensitivity to wording and phrasing \cite{McCoy2024llmsshapedbyproblem}), hallucinations \cite{huangSurveyHallucinationLarge2025, xuHallucinationInevitableInnate2024}, and weak generalizations (e.g., the inability to perform on unfamiliar (even trivial) problems \cite{nezhurina2024alicewonderlandsimpletasks}).
 
These sometimes-covert limitations appear starkly in ongoing ``struggle[s] with implied meaning and pragmatics'' \cite{changLanguageModelBehavior2024},  and findings that LFM models ``cannot perform genuine logical reasoning''~\cite{AppleGSM}, that they ``struggle with counterfactual reasoning'' \cite{yaminCanLLMsReconcile2025}, and that planning and out-of-distribution task performance is fragile \cite{yuWordFrequencyDoes2020, saparovLanguageModelsAre2023, misra2023compsconceptualminimalpair, kassner-schutze-2020-negated, kwonWhyMaskedNeural2019, caoCanPromptProbe2022, elazarMeasuringImprovingConsistency2021, chaves-richter-2021-look}.
Additionally, although syntactic abilities are considered to have been robustly learned in LFMs, nonetheless it is found that subject-verb agreement performance degrades as more infrequent or complex strings are seen, and predictions can be over-sensitive to token position information \cite{yuWordFrequencyDoes2020, changLanguageModelBehavior2024, chaves-richter-2021-look}.

We provide empirical findings that a broad range of language models can be sorted by generative testing on the evaluation of strings produced from rule-based grammars at three distinct tiers of the G-A hierarchy in Section 4. The results are shown to be statistically significant (see Supplementary Material A) and situate the empirical abilities of each system within these tiers.
We will see that these LFM evaluation results conform to three categories, corresponding to model size and architecture. The results suggest that size alone only gets you so far; only particular architectures are able to accept the productions of grammars at supra-IXG tiers. 
These informative outcomes are discussed further in Section 4.

\section{Benchmarking computational capabilities}

A novel benchmarking method is presented that combines insights from formal language theory with the capabilities of generative AI to evaluate pre-trained large language models. By utilizing symbolic recursive generators -- specifically, formal grammars -- and leveraging the known correspondence between formal grammar classes and automata hierarchies, a generative paradigm for testing the computational limits of language models can be created.

The core idea is to assess a model’s ability to process complex structural patterns by subjecting it to tasks  grounded in formal grammar constructs of varying computational difficulty.  We emphasize that we  test existing systems without providing any additional training, fine-tuning, or other modification.  We test systems exactly as they are.  
 
The approach begins with the observation that language models are trained on enormous corpora of natural language text from the internet. This fact makes it non-trivial to craft evaluation examples that the model has not encountered in some form.
To address this, formal grammars can be relied upon to systematically generate strings (synthetic sentences) that have not previously been experienced even by models trained on trillions of sentences, i.e., the new strings are out of distribution. These strings may be composed of numbers, natural language words or tokens, or other symbols.
If composed of words (or tokens) of a natural language, these strings may look superficially like valid sentences, yet are constructed to be semantically nonsensical and structurally driven by specific formal grammar rules. 
(It might have been expected that the extensive training datasets of language models may have contaminated such testing, yet the benchmarking methods ensured that the sentences were highly likely to be out of distribution by generating syntactically valid but novel and nonsensical strings (see Supplementary Material A, B, C, and D).) 
 
The set of all strings from a grammar constitutes a formal language; common notation gives $\Sigma^* \text{ or } \Gamma^*$ as the set of finite-length strings over the alphabet $\Sigma \text{ or } \Gamma$, and $[\:] \text{ or } \epsilon$ denotes the empty string  \cite{sipserIntroductionTheoryComputation2006}.

\begin{definition}
    A \textit{formal grammar} is defined by a tuple $G = \left( V, \Sigma, R, S \right)$, where $V$ is a finite set of variables giving the \textit{non-terminal symbols}, $\Sigma$ is a finite set of \textit{terminal symbols} such that $\Sigma \cap V = \emptyset$, $R$ is a finite relation in $V \times (V \cup \Sigma)^*$, where $^*$ represents the Kleene star operation, which gives a set of \textit{rules}, and $S \in V$ is the \textit{start symbol}.
    \label{def:grammar}
\end{definition}

If the terminal symbols are the words of a natural language, then the generated strings can be seen as \textit{grammatical nonsense}: well-formed according to a specific formal grammar but devoid of coherent meaning. By using natural language tokens as the terminals of these grammars, the tasks remain in the realm of natural language; thus, the model’s syntactic abilities are engaged, even though there may be no real-world meaning.

Crucially, sets of formal grammars can be employed to represent a particular tier of the G-A hierarchy. For example, one might generate sentences that require an understanding of  context-free patterns involving nested dependencies, or even more complex context-sensitive structures. Each of these sets of sentences tests whether a system can handle the specific type of dependency or recursive structure characteristic of the corresponding grammar generator(s).

Because the sentences are generated \textit{de novo} by these grammars, the likelihood that the model has seen them in its training data is significantly reduced.  Evidence of this is seen in Section 3.1: many models, although trained on English corpora, are found to not accept these generated well-formed English sentences as grammatical.

The nonsensical nature of the content (despite using common words) ensures that the model cannot rely on world knowledge or semantic priors to \textit{reliably} answer correctly -- it must recognize the formal structure.
In essence, a set of formal grammars  provide symbolic recursive generators that encompass wide ranges of a specific computational tier.

\subsection{Evaluation Methodology}

It is clarifying to look at the benchmark tasks in the following (although overlapping) categories of formal languages: natural language, synthetic language, and arithmetic language.

\subsubsection{Natural Language} Natural language tasks are the most restrictive, a consequence of the fact that the grammars of natural languages are unknown. Predominantly, tasks will inquire about inclusion. These may ask, in terms of a specified natural language: \textit{Is the given string grammatical?}

Natural language tasks  crucially offer a direct assessment of complex combinations of (small) building-block rules that comprise the input domain (human-generated text). This is a largely unexplored method for evaluating language models, and consequently, the example grammar creation and  empirical evaluations are of this type.

Appearing as natural language at first glance leads to a response of \textit{grammatical acceptance} reliant on formal linguistic competence -- recognition of types and their relational dependencies -- that is not affected by a statistically improbable length or number of sub-rule repetitions from the recursive symbol expansions.

\subsubsection{Synthetic Language} Formal grammars  offer the strength and diversity of possible synthetic language tasks. Sufficiently powerful grammar rules offer a rigorous way to encode the syntax of expressions of any algebraic rule system \cite{backusSyntaxSemanticsProposed1959,postFormalReductionsGeneral1943, backusReportAlgorithmicLanguage1960}.
Synthetic language tasks offer both a greater range of tasks and the ability to make use of broader input domains, instead of limiting terminal symbols to natural language tokens  or to numbers.  
Providing a grammar and generations from that grammar enables more expansive inquiries, such as the following:
{\it
Given the grammar $G$, is the following string in the grammar's language?
Provide a valid next token given the following partial string generated by the grammar $G$?
Reverse the following string.
}

Examples of implemented synthetic language tasks include: associative recall tasks used in language model \emph{primitive} development, e.g., \citet{aroraZoologyMeasuringImproving2023}, \citet{fuHungryHungryHippos2022},  \citet{olssonIncontextLearningInduction2022}; string manipulation tasks, e.g., ``Reverse String'' and ``Duplicate String'' \cite{deletangNeuralNetworksChomsky2023}; and monoidal word problems, e.g., $S_5$ \cite{merrillIllusionStateStateSpace2024}. 

A central problem for algebraic structure is the word problem:
``given a structure in terms of generators and relations, deciding whether two given words over the generators represent the same element of the structure'' \cite{nyberg-broddaWordProblemSpecial2020}.

\subsubsection{Arithmetic Language} Arithmetic rules are in the training data and have been a focus of empirical model evaluations. For example, assessments in \citet{deletangNeuralNetworksChomsky2023} and \citet{daveInvestigatingSymbolicCapabilities2024} include parity checks, modular arithmetic, and square root computations, which are in the regular, context-free, and context-sensitive grammar tiers, respectively. 

Formal grammars offer a method to easily produce interpretable problems beyond those of simple addition, logic gate, or scaling properties. This can simply arise from the hierarchical relations of the grammar rules, stacking logical operators, or deep compositions of arithmetic rules.

Previous work have evaluated LFMs from the formal language perspective but testing has so far has only been performed in the synthetic and arithmetic language task categories \cite{daveInvestigatingSymbolicCapabilities2024, deletangNeuralNetworksChomsky2023, sarrofExpressiveCapacityState2024}.

\subsection{Indexed grammar generator}

To demonstrate the feasibility of the newly introduced benchmarking method, an example indexed grammar for synthetic string generation that satisfies the conditions is defined and described below.

The \emph{indexed sentences} are strings in the language of the grammar $L(G_{ix})$ (i.e., the set $L = \{\sigma \in \Sigma \:|\: S \xRightarrow{*} \sigma \}$), where $G_{ix}$ is a nondeterministic indexed grammar. (See Definition \ref{def:grammar}.)

Using \textit{productions} (or \textit{phase structure rules}) as defined by \citet{Chomsky1975}, a generator $G_{ix}$ can be given by $V = \{S, NP, VP, N, Cnj, D, Adj, V_{\text{intr}}, V_{\text{tr}} \}$, $\Sigma$ given by English words, $S$ is the start symbol, and the rules $R$ are as follows.

Let $G_{ix}$ be defined by
\begin{center}
\small
    \begin{tabular}{l l}
        {$S\rightarrow NP^s \;\; VP^s \;|\; NP^p \;\; VP^p \;|\; [\:]$} & $D^s \rightarrow \sigma_{D^s} \in \{$ ``a'', ``the'', ``this'', \dots $\}$ \\[0.5ex]
        
        $NP \rightarrow NP^s \;\; NP^p$ &
        $pN \rightarrow \sigma_{pN} \in \{$ ``Queen'', ``Tolstoy'', \dots $\}$ \\[0.5ex]
        
        $NP^s \rightarrow D^s \;\; N^s \;\; T^s \;|\; D^s \;\; Adj \;\; N^s \;\; T^s$ &
        $N^s \rightarrow \sigma_{N^s} \in \{$ ``machine'', ``idea'', \dots $\}$ \\[0.5ex]
        
        $NP^p \rightarrow D^p \;\; N^p \;\; T^p \;|\; D^p \;\; Adj \;\; N^p \;\; T^p$ &
        $Adj \rightarrow \sigma_{Adj} \in \{$ ``nice'', ``green'', \dots $\}$ \\[0.5ex]
        
        $VP^s \rightarrow V_{intr}^s \;|\; V_{tr}^s \;\; NP \;\; T^s$ &
        $V_{\text{intr}}^s \rightarrow \sigma_{V_{\text{intr}}^s} \in \{$ ``sings'', ``laughs'', \dots $\}$ \\[0.5ex]
        
        $VP^p \rightarrow V_{intr}^p \;|\; V_{tr}^p \;\; NP \;\; T^p$ &
        $V_{\text{tr}}^s \rightarrow \sigma_{V_{\text{tr}}^s} \in \{$ ``joins'', ``enjoys'', ``says'', \dots $\}$ \\[0.5ex]
        
        $T \rightarrow  [\:] \;|\; P \;\; NP \;|\; pN \;\; VP^s$ &
        $P \rightarrow \sigma_{P} \in \{$ ``with'', ``in'', ``from'', \dots $\}$ \\[0.5ex]

        $D^p \rightarrow \sigma_{D^p} \in \{$ ``these'', ``those'', \dots $\}$ &
        $N^p \rightarrow \sigma_{N^p} \in \{$ ``machines'', ``whales'', \dots $\}$ \\[0.5ex]
        
        $V_{\text{tr}}^p \rightarrow \sigma_{V_{\text{tr}}^p} \in \{$ ``join'', ``like'', ``need'', ``say'', \dots $\}$ & $V_{\text{intr}}^p \rightarrow \sigma_{V_{\text{intr}}^p} \in \{$ ``sing'', ``laugh'', \dots $\}$ \\[0.5ex]
    \end{tabular}
\end{center}

\noindent where $[\:]$ denotes the \textit{skip} or the empty string.

\begin{figure}
    \centering \includegraphics[width=0.95\linewidth]{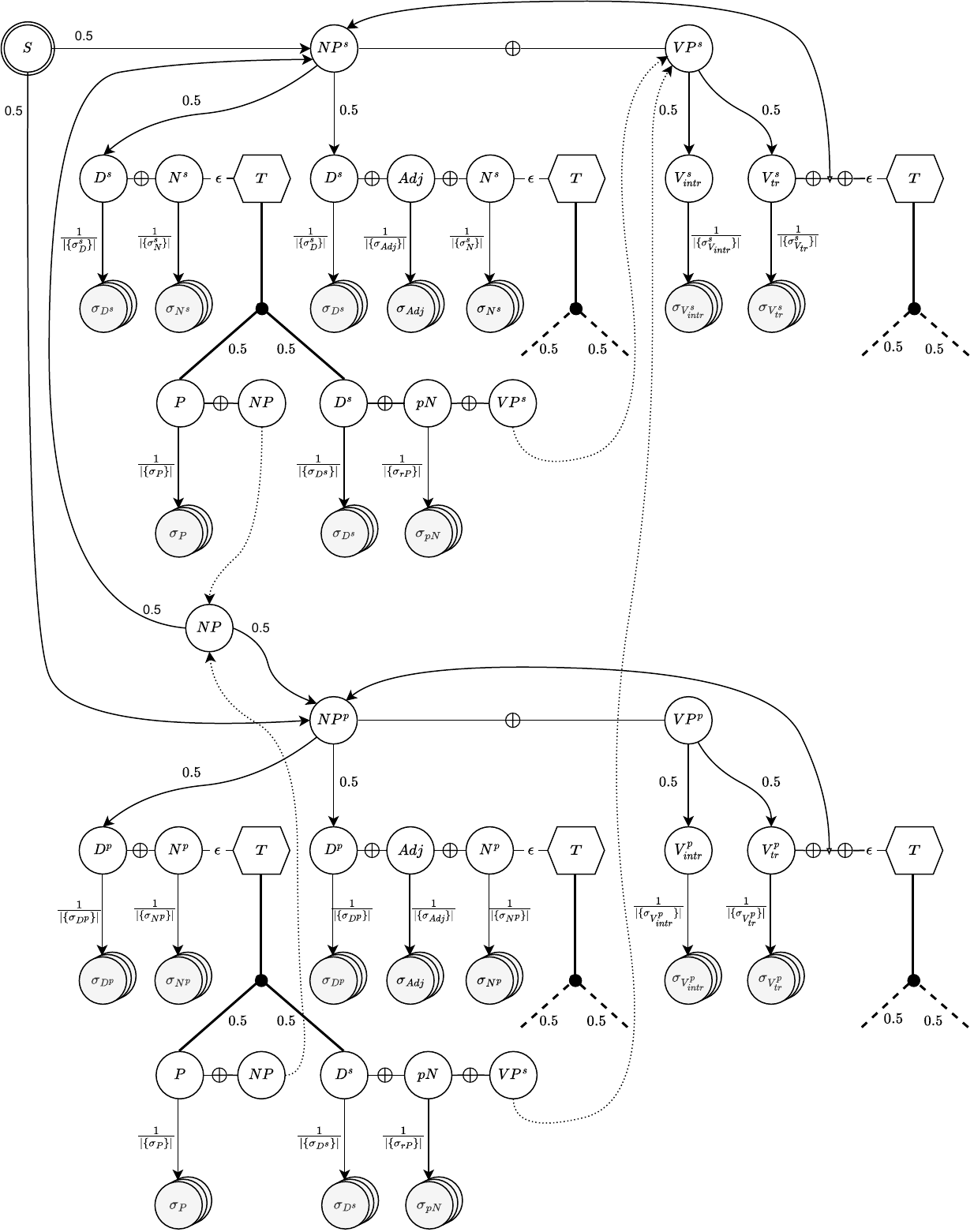}
    \caption{\small Generative syntax expansion diagram of (nontrivial) $G_{ix}$ rules, where nodes with the symbols ($V \cup \Sigma$), each directed edge depict probabilistic transitions ($R$) to the next node. The subtrees, hexagonal $T$ nodes, are expanded by the bold edges and repeated for the trees with to dashed edges.
    An $\oplus$ is the addition of sequentially connected symbols. The $\epsilon$-edges denote the $\epsilon$-probability of skipping the next node. The non-terminal symbols $V$ are given by white nodes and terminal symbol $\Sigma$ nodes are light grey.}
    \label{fig:Gix}
\end{figure}

Specifically, $G_{ix}$ is a tree-adjoining grammar (TAG), as defined by \citet{Joshi1997}, which is a (supra-\textit{context-free}) class of grammars that are a subset of the indexed grammars class \cite{Aho1968indexed}. The \textit{embedded} nested dependencies that can be constructed by subtrees $T \in V$ of $G_{ix}$ can not be handled by context-free grammars. The expansion of $G_{ix}$, with its complex embedded structure of nested dependencies is graphically depicted in Figure \ref{fig:Gix}.

\subsubsection{Language}
To express the language of $G_{ix}$, let singular combinations of terminals be defined by $\dot{\alpha} \coloneqq \sigma_{\text{D}^s} \: \sigma_{\text{N}^s}$,  $\dot{\beta} \coloneqq \sigma_{\text{D}^s} \: \sigma_{\text{Adj}} \: \sigma_{\text{N}^s}$,  $\dot{\xi} \coloneqq \sigma_{\text{V}_{\text{tr}}^p} \dot{\alpha}$,  and $\dot{\vartheta} \coloneqq \sigma_{\text{V}_{\text{tr}}^p} \dot{\beta}$, 
and plural combinations of terminals by
$\ddot{\alpha} \coloneqq \sigma_{\text{D}^p} \: \sigma_{\text{N}^p}$, 
$\ddot{\beta} \coloneqq \sigma_{\text{D}^p} \: \sigma_{\text{Adj}} \: \sigma_{\text{N}^p}$, 
$\ddot{\xi} \coloneqq \sigma_{\text{V}_{\text{tr}}^p} \ddot{\alpha}$, and 
$\ddot{\vartheta} \coloneqq \sigma_{\text{V}_{\text{tr}}^p} \ddot{\beta}$.

Then, the sets of possible subtrees $T$ is
\begin{align*}
    \mathbb{T} \triangleq \:& \left\{ \sigma_{\text{P}} \: \dot{\alpha}\: T \right\} \cup \left\{ \sigma_{\text{P}} \: \ddot{\alpha} \: T \right\} \cup
    \left\{ \sigma_{\text{P}} \: \dot{\beta} \: T \right\} \cup \left\{ \sigma_{\text{P}} \: \ddot{\beta} \: T \right\}
    \cup \left\{ \sigma_{\text{D}^s} \: \sigma_{\text{pN}} \: \sigma_{\text{V}_{intr}^s} \right\} \cup \left\{ \sigma_{\text{D}^s} \: \sigma_{\text{pN}} \: \dot{\xi} \: T \right\} 
    \cup \left\{ \sigma_{\text{D}^s} \: \sigma_{\text{pN}} \: \dot{\vartheta} \: T \right\} \cup \{[\:]\},
\end{align*} 
recursively defined where $T \in \mathbb{T}$.

Traversing the graph of $G_{ix}$ in Figure \ref{fig:Gix}, it can be seen that the language expressed ($\mathcal{L}(G_{ix})$) may be represented by the closed form
\begin{align*}
\mathcal{L}\!\bigl(G_{ix}\bigr)=
&\Bigl\{
      \dot{\alpha} T \sigma_{V_{\text{intr}}^{s}}
      \cup
      \dot{\beta}\, T  \sigma_{V_{\text{intr}}^{s}}
      \cup
      \ddot{\alpha} T  \sigma_{V_{\text{intr}}^{p}}
      \cup
      \ddot{\beta}\, T \sigma_{V_{\text{intr}}^{p}}
      \Bigm| T \in \mathbb{T}
  \Bigr\} \\ 
 & \cup
\Bigl\{
      \lambda_{1}^\iota  T 
      \sigma_{V_{\text{tr}}^{\iota}}\,
      \lambda_{2}^\iota  T
      \Bigm|
      T \in \mathbb{T}, 
      \iota\in\{s,p\},
      \lambda_{1}^\iota, \lambda_{2}^\iota\in
      \begin{cases}
           \{\dot{\alpha},\dot{\beta}\} & \text{if } \iota=s\\[2pt]
           \{\ddot{\alpha},\ddot{\beta}\} & \text{if } \iota=p
      \end{cases}
  \; \Bigr\} \cup \left\{ [\:] \right\}.
\end{align*}

Strings of this language, $\ell \in \mathcal{L}(G_{ix})$, include \textit{The machine with a green idea sings.} and \textit{``This nice person, the Queen, enjoys the song from the band,'' says the musician.} (without punctuation and with added ``person'', ``band'', ``musician'' $\in \{\sigma_{\text{N}^s}\} \subset \Sigma$). It is both the rules and string generation process that require additional enforcement to create the indexed-level dependencies.

\paragraph{In General}
Evaluating a trained language system offers an initial indication of whether the system will be capable of different levels of computational skills.
The full symbolic evaluation method presented here probes the fundamental intrinsic computational ability of LFMs, enabling a comprehensive evaluation across  input domains, generative datasets, and a range of complex rule-based tasks:  
a holistic measure of computational class.

\section{Empirical performance of humans and LFMs on three tiers of grammars}

Human subjects, and fifteen language foundation models (LFMs) of different sizes and compositions, were presented with candidate sentences from the CFG class, the IXG class, and the CSG class, as shown in Figure \ref{fig:GAhierarchy}. (For generation details see Supplementary Material C and D). Simple illustrative examples from each of the grammars are here:

\begin{description}
    \item[CFG] The defective items chase a blueberry and a person or an intricate lion.
    \item[IXG] The dogs which chase Lagrange who moves Euler who falls find the trees that believe in the fair theorems.
    \item[CSG] The functions Ramanujan silently laugh a university to the intricate building conjectures desperately votes.
\end{description}

The biological and artificial systems were asked to judge the strings to be either grammatically acceptable or unacceptable (see Supplementary Material A). 
Importantly, all LFMs were evaluated off the shelf via standard inference (zero-shot), with no task-specific fine-tuning nor retraining.
 
The artificial systems are classified as small LM (Llama 3.1 8b \cite{llama3.1}, Mistral 7b \cite{mistral7b}), mid-sized LM (Gemini 2.0 flash \cite{deepmindGemini20Flash2024, blogIntroducingGemini20}, Llama 3.1 70b \cite{llama3.1}, Claude 3 Haiku \cite{claude3}, Gemini 1.5 flash \cite{gemini1.5}, GPT-3.5 Turbo \cite{gpt3.5-turbo}), large LLM (Claude 3 Opus \cite{claude3}, DeepSeek R1 \cite{deepseek-aiDeepSeekR1IncentivizingReasoning2025}, GPT 4 \cite{gpt4}, Llama 3.1 405b \cite{llama3.1}), or augmented (ATT) LFMs (o1, o1- preview, o1-mini \cite{o1, o1-preview, o1-mini}, Gemini 1.5 Pro \cite{Kilpatrick_BasuMallick_Kofman_2024}) (Supplementary Material A details  model specs and  size classes.)
(Our ongoing research has tested many more systems, which are being incorporated into futher upcoming findings.)

\begin{figure}
    \centering
    \includegraphics[width=0.65\linewidth]{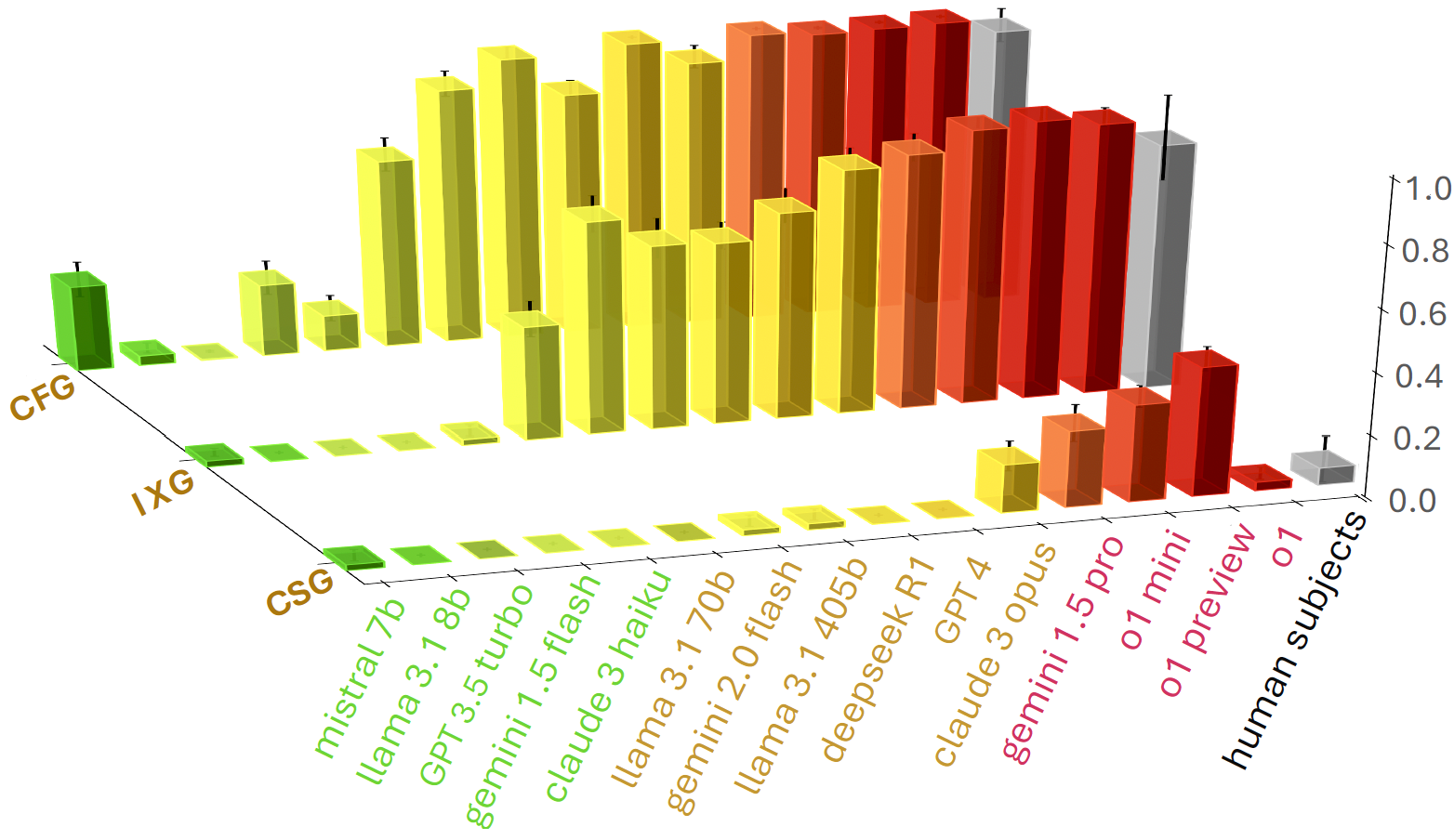}
    \caption{Empirical percent acceptance (\emph{y axis}) of three formal grammar class tiers (CFG, IXG, CSG; \emph{z axis}) by small language foundation models (LFMs) (\emph{green}), large LFMs (\emph{yellow}) and augemented transformer-based LFMs (ATTs) (\emph{red}); and human subjects (\emph{gray bars}) (\emph{x-axis}). (Error bars are std.err.). The differences between small (green), large (yellow), and augmented (red) systems are rigorously found significant (p < 0.01) both by ANOVA and by Kruskal-Wallis tests (see Supplementary Material A.3)}
    \label{fig:empircal}
\end{figure}

Figure \ref{fig:empircal} summarizes the results: the responses of various transformer decoder language models (ordered by estimated sizes, from 7B to 2T parameters), to sentences from these three tiers of the G-A hierarchy. All results shown are statistically significant; detailed results and statistical analyses are given in Supplementary Material A. 
It can be seen that:

\begin{description}
    \item[Small LFMs (under tens of billions of parameters)] did not reliably succeed at processing any of the inputs; some recognize a subset of CFG strings but no higher-level grammars
    \item[Mid-sized LFMs (from tens of billions to hundreds of billions of parameters)] exhibited somewhat 
    more reliable recognition of CFG grammars but still unreliable on higher levels.
    \item[Large LFMs (over hundreds of billions of parameters)] and humans, exhibited robust recognition of both CFG and IXG strings; no tested language models (up through $2$ trillion parameters), nor humans, exhibit reliable recognition of CSG sentence strings. 
    \item[ATT LFMs (systems with intercalated augmentation of  traditional transformer-based LFMs)] showed overwhelming acceptance of CFG and IXG strings, plus a limited yet significant portion (up to 44\%) of CSG sentences. 
\end{description}
 
It is notable, and perhaps surprising, that the LFMs’ acceptability judgments obey a clearly recognizable progression with respect to model parameter size and model architecture. 
The clear distinctions between model abilities, even within (corporate) developers, support that 
the nonsensical generated sentences are evidently not found in any existing training corpus. The generated sentences can be seen as being out of distribution, strongly indicating that they can only be recognized by 
extrapolation of
the abstract grammatical rules of automata at specific tiers. 

(Importantly, since all test sentences are newly produced for each trial via the generative grammars given here, we eliminate the need to 
``hold back'' test samples.  Instead, the inherently generative method consistently composes arbitrary new tests that are out of distribution.)

These empirical findings accord with extensive analyses from the computational linguistics literature: small traditional transformer-based LFMs’ abilities are less than those of humans, whereas large traditional transformer-based LFMs correspond with human judgments; and no language foundation models reliably process sentences at tiers higher than those of human language. The intercalated incorporation of additional advanced algorithms (reinforcement learning, mixtures of experts, etc.) into the transformer architecture enables some behaviors that transformer-based language models alone do not readily produce, at any scale \cite{blogIntroducingGemini20, openai-reasoning-post}.
 
The CSG-generated sentences offer perhaps the most striking results. These sentences are not only ``out of distribution,'' they are technically ungrammatical in terms of human natural languages, and thus not acceptable according to human judgments (both from the literature and from the present human subjects' performance). 
Humans cannot naturally parse these in the seemingly automatic manner of IXG-tier natural language sentences; the CSG-generated sentences can in principle be parsed, far more arduously, by a human who can formally work through the abstract grammar rules of the sentence structures. Correspondingly, neither human subjects, nor any traditional transformer-based LFMs, accepted these CSG-generated sentences, yet the ATT systems did accept a fraction of these CSG-generated sentences:  enough to be statistically significant (see Supplementary Material A for  extensive statistical analyses).

The binary  response from a model is not deep evidence that the model has actually parsed the test sentence.  Advanced systems (ATT models) were separately asked to provide an explicit parse  of each sentence after its yes/no response; if the binary answer does  reflect the model's parsing ability, then a `yes' would be accompanied by a correct parse, and a `no' would yield an attempted but failed parse.  This is ongoing work, and preliminary data are provided in Supplementary Material G.   

The ATT systems are evidently accepting supra-IXG patterns via abstraction mechanisms that are out of reach for LFMs \cite{kambhampatiCanLargeLanguage2024, kambhampatiLLMsCantPlan2024}. 
The natural question is raised: what are the relations between sheer (quantitative) scaling of LFM size (from
small to large LFMs), on one hand, versus (qualitatively) distinct algorithms (LFMs to ATT LFMs), on the other?

\section{The nature of scaling in brains and in language models}
 
\subsection{Scaling and computation in brain architectures.}

Although humans can perform both natural language
processing and advanced logical reasoning tasks,  evidence indicates that the
acquisition and performance of these respective
abilities engage very different brain mechanisms.

As discussed (see Section 3.1), 
extensive evidence 
finds connections between human natural language and the IXG tier of the G-A hierarchy, above context-free and below context-sensitive grammars \cite{Pullum1982, Joshi1997, VijayShanker1994, Shieber1985}.
By contrast, evidence indicates that nonhuman communication systems,
despite having rich and useful characteristics, exhibit only sub-IXG language abilities \cite{suzukiInformationEntropyHumpback2006, berwickSongsSyntaxLinguistics2011, stobbeVisualArtificialGrammar2012, fitchComputationalConstraintsSyntactic2004, fitchArtificialGrammarLearning2012, gentnerRecursiveSyntacticPattern2006, hermanComprehensionSentencesBottlenosed1984, vanheijningenSimpleRulesCan2009, kershenbaumAcousticSequencesNonhuman2016, kershenbaumAnimalVocalSequences2014, lefebvreBrainsInnovationsEvolution2004, berwickEvolutionBrainNature2013, beckersWhatAnimalsLearn2017}.
Moreover, in stark contrast to humans’ universal and seemingly effortless natural language acquisition, no humans learn mathematics, formal logic, or other advanced reasoning abilities without extensive specialized training. 

In other words, humans appear to effortlessly attain capabilities associated with the IXG tier but require extensive training and tools to obtain capabilities associated with higher tiers.

Differential task-specific brain activity has found language processing to be distributed through a seemingly constrained subset of particular cortical sites, often termed the language network \cite{Fedorenko2024, Blank2016}.
By contrast, substantial evidence suggests that mathematical and logic tasks engage a separable set of structures beyond those engaged solely in the language network \cite{tuckuteLanguageBrainsMinds2024, Duncan2010, amalricOriginsBrainNetworks2016, montiThoughtLanguageNeural2012}.

\begin{figure}
    \centering
    \includegraphics[width=0.45\linewidth]{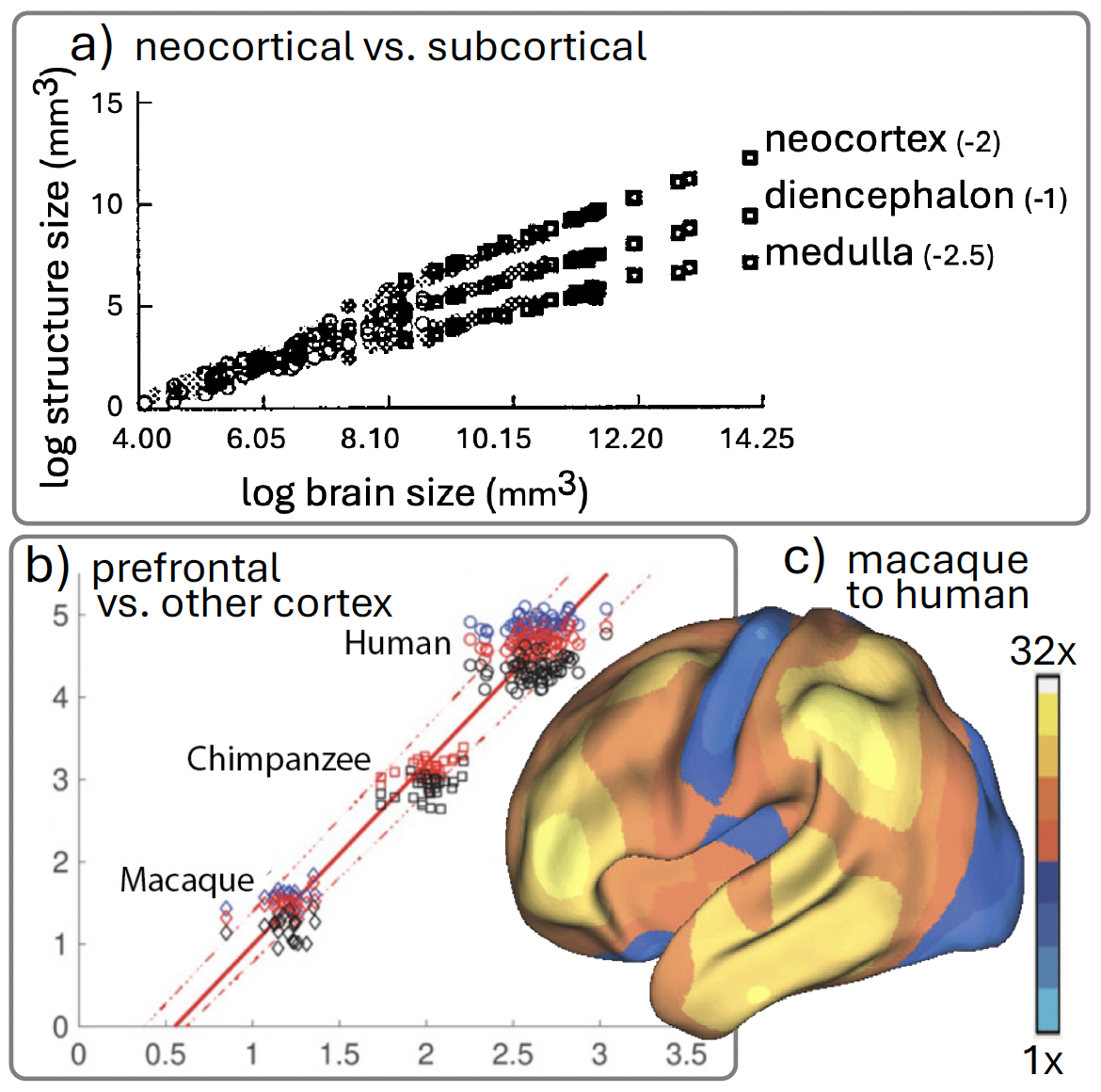}
    \caption{a) As brain
size increases, each structure allometrically expands by
different factors; for instance, neocortex
disproportionately increases faster than subcortical
structures (adapted from \citet{Finlay2001}). b) Within
neocortex, distinct regions also expand differently as
brains get larger; prefrontal areas show the most
expansion whereas primary sensory areas such as visual
cortex expand very little. Shown are the relative sizes of
prefrontal vs visual cortex (colors show different
measurement methods), for three different primate brain
sizes (adapted from \citet{Donahue2018}. c) Estimated
factors of expansion from macaques to humans. Frontal,
parietal and temporal areas are by far the most
disproportionately expanded (orange and yellow regions)
(adapted from \cite{bucknerEvolutionDistributedAssociation2013}).}
    \label{fig:brainsizes}
\end{figure}

Those differential math- and logic-activated brain areas are frontal, parietal, and anterior temporal areas,  all  regions that are disproportionately increased in relative size in humans \cite{Finlay2001, bucknerEvolutionDistributedAssociation2013}
(see Figure \ref{fig:brainsizes}). As
absolute and relative brain sizes (brain to body ratio) increase among the primates, all of neocortex  increases, but the frontal, parietal, and anterior temporal regions increase far more than the rest \cite{Herculano-Houzel2009, Striedter2005, Finlay1995}. Figure \ref{fig:brainsizes}a shows that neocortical areas are systematically increased in size more than subcortical structures (y axis) as overall brain size increases (x axis). 
(Note that the axes are logarithmic, i.e., even seemingly small changes are considerable). 
Figure \ref{fig:brainsizes}b further shows that, within the neocortex, the most anterior regions (prefrontal)  differentially increase  most in size.  
Figure \ref{fig:brainsizes}c illustrates these differential growth effects in humans compared to nonhuman primates:  brain area sizes in macaques are compared with those in humans, highlighting the most differential growth in frontal, parietal, and temporal cortical areas. 
For instance, an average macaque monkey’s total neocortex volume is roughly 34cc, of which roughly 5cc ($\sim$ 15\%) is frontal cortex, whereas a human neocortex is roughly 513cc, of which roughly 130cc ($\sim$ 25\%) is frontal cortex \cite{Donahue2018, Glasser2014, VanEssen2012Parcellations}.

\citet{Friederici+Singer2015} highlight the pluripotency of cortical computation, noting that ``cortical areas supporting language processing should operate according to principles similar to those cortical areas dealing with sensory and executive processes in the non-language domain'' 
\cite{Friederici2011, Friederici+Singer2015, friedericiNeuralBasisHuman2018}. Taken together, there is substantive evidence for a consistent repertoire of underlying computations and algorithms throughout neocortex, responsible for general processing of those inputs that reach given areas. The language network may be performing processing of a kind much like that for non-linguistic information, all of which may be rendered in brain networks in the form of grammars, whether representing language or other environmental data \cite{granger2020quantificationcognition, Rodriguez_2016}.

If similar computations are being carried out in brain
areas that process language, and those that process
logic, how might scaling affect such computations and
abilities? The answer is that scaling a system may not
simply be uniformly increasing all aspects of that
system. Rather, scaling very often disproportionately
scales parts of systems more than other
parts, unveiling abilities that were small and limited
in the small version of the system but are prominent
in the large system, even though those abilities were already
present in both small and large systems. We provide
two brief illustrative examples of specific brain
scaling that highlights this underlying principle.

\begin{description}
    \item[Motor system scaling] Motor control is operated by
combinations of cortical and subcortical (striatal and
brainstem) systems; the former are evolutionarily
newer circuits than the latter, and are composed of
quite distinct cell types and circuit designs, so motor
behaviors in small- and large-brained animals (e.g.,
mice vs. humans) operate by substantially different
rules. These components scale very differently as
brains grow large. If a mouse’s motor cortex is
damaged, the mouse exhibits relatively minor motor
defects, whereas damage to its striatal complex can
cripple its motor abilities \cite{delongRoleBasalGanglia1984, dudmanBasalGangliaMotor2016}.
By contrast, human 
motor cortex damage may cause irreparable paralysis,
whereas striatal damage typically causes relatively less
motor impairment. In sum, although the mouse and human
brain differ primarily  via  scaling, 
nonetheless motor operations are almost completely
different: they are predominantly striatally-run in a mouse but
predominantly cortical in a human. The numeric size
change begets qualitative 
striatal vs cortical
dominance, resulting in 
different motor abilities and limits.

\item[Visual system scaling] 
Visual cortex damage induces blindness in humans, but a similarly lesioned rodent  exhibits largely intact visual abilities. This is largely due to an intact highly conserved superior colliculus, which confers considerable visual ability to mammals, and to reptiles via its homologue, the optic tectum \cite{steinSensoryRepresentationReptilian1981, steinReceptivefieldPropertiesReptilian1983, hoySuperiorColliculus2025}.  As in the motor example above, visual processing is overwhelmingly cortical in large-brained mammals but largely collicular in small-brained mammals, again  a qualitative change arising from not from simple scaling but from rebalancing the sizes of two largely independent (and sometimes competing) systems. 
(It is worth noting that the relative speed of collicular visual processing privileges its use in extreme circumstances such as a second baseman catching a line drive -- visuomotor behavior too fast for normal cortical processing.  The colliculus is also considered largely responsible for residual visual abilities present in human `blindsight' \cite{ptitoNeuralSubstratesBlindsight2007,tamiettoCollicularVisionGuides2010,bassoUnravelingCircuitsVisual2021}.)
\end{description}

Related effects occur throughout brain and behavior: many functions engage both cortical and subcortical components.
The voluminous matrix-like synaptic layout of much of association cortex allows general encoding and storage of elaborate combinations of inputs (see \cite{rollsNeuralNetworksBrain1999, wallisSingleNeuronsPrefrontal2001, freedmanCategoricalRepresentationVisual2001, raoIntegrationWhatWhere1997})
differing greatly
from most subcortical areas which each engage more specialized machinery, not affording the breadth or flexibility that is enabled by cortical
function. Thus, the scaling-based emergence of
novel, qualitatively distinct behaviors may be
understood as arising not from scaling per se, but
directly due to corticalization, i.e., a changeover
between primarily subcortical processing in small
brains vs. predominantly cortical processing in large
brains, even though cortical and subcortical areas
continue to 
interact during almost all
known behaviors.

Brains are constructed from probabilistic
components (synapses, cells), and indeed much of our
behavior (perception, decision, memory) is
correspondingly statistical. How, then, can we add
two arbitrary numbers and arrive reliably at the correct sum?
By intentionally suppressing our natural (statistical)
intuitive operators and instead ``following the rules''
of arithmetic. If we instead use natural common
sense, or intuition, we will usually
arrive at incorrect answers in the realms of math and
logic.  Rules for logic are unnatural operations to us, acquired
only via careful  instruction.
 
\subsection{Scaling and computation in LFMs.}

Without a large enough model, many of the less (computationally) costly tasks are, empirically, not reliably within the capabilities of a language model (e.g., Supplementary Material E).

In the empirical evaluations small LFMs did not recognize indexed grammars (IXG) whereas large LFMs did, yet large LFMs did not appear to further scale up to higher tiers (context-sensitive grammars) or beyond. Why might sheer scaling of size enable the transition from sub-IXG to IXG, yet not enable the next transition from IXG to CSG? Effects of scaling transformers are well documented \cite{bidermanPythiaSuiteAnalyzing2023, kaplan2020scalinglawsneurallanguage}, sometimes incorporating apparently entirely new features that only appear in the scaled-up system but not in the smaller system \cite{wei2022emergentabilitieslargelanguage}. Such emergent abilities are often described as qualitative changes, i.e., entirely new phenomena, that arise from solely quantitative or numeric changes, i.e., mere scaling; the underlying mechanisms for these changes remain poorly understood \cite{schaeffer2023emergentabilitieslargelanguage, wei2022emergentabilitieslargelanguage, kaplan2020scalinglawsneurallanguage, lubana2024percolationmodelemergenceanalyzing}.

\citet{cui2024phasetransitionpositionalsemantic} provides an instance in which a qualitative change appeared seemingly from scaling up alone: in sum, they carefully detailed a sudden shift of the transformer attention mechanism, from position-based to seemingly position-independent attention as the quantity of training data increased.

This shift seems to reflect a qualitative change in the underlying algorithmic strategy that the system is using: there appears to be a changeover from one form of processing (relying on position-encoded information) to another (relying on positional-encoding-independent, token-embedded content).

But importantly, both processing abilities are inherently present in their mechanism throughout, so it is not that a new ability arises from the scaling on input. Rather, the scaling triggers a shift between reliance on the two mechanisms that were both already available, as illustrated in brain architectures. As in those instances, rather than scaling per se, this may  be characterized as one internal mechanism supplanting another, in part as an effect of increased training.

Transformers' abilities to capture complex patterns are closely tied to the architecture's depth and to the number of attention heads. \citet{weiss2021thinkingliketransformers} empirically show that reducing the number of layers, or layer width (heads), reduces accuracy on most tasks. These numeric measures of transformers (number of heads, layers, number of parameters) may figure crucially in their ability to process complex grammatical data.
In general, 
expressivity of transformer variants depend on many detailed aspects of the attention mechanism and architecture \cite{stroblWhatFormalLanguages2024},
and the balance between position-encoded information versus token-embedded, position-independent information
(\citet{cui2024phasetransitionpositionalsemantic} and \citet{allen-zhuPhysicsLanguageModels2023}), 
is crucial for the transition from one computational tier to the next.  
Where the empirical computational upper bound currently resides for today's large language models has been a topic of extensive and ongoing study; many researchers have provided empirical evidence of transformers (as well as RNN- and LSTM- based systems) performing differently on tasks that correspond to lower vs. higher tiers of the G-A hierarchy 
(\cite{deletangNeuralNetworksChomsky2023, bhattamishraComputationalPowerTransformers2020, zhouWhatAlgorithmsCan2023, yang2024masked, stroblWhatFormalLanguages2024, daveInvestigatingSymbolicCapabilities2024}).

\subsection{Beyond scaling of the transformer architecture.}
The findings lead to the
hypothesis that the absence of ``extendable accessible memory'' 
in transformers may impose strict limits on their ability to benefit from scaling. Although it seems obvious that the largest LFMs have ample
available access to memory, the key is in the use that
the architecture makes of that memory. 
Transformers are configured to
treat their internal representations as a unified
reservoir of positional embeddings. Notably, Nye et
al. (2021) showed that ``scratchpads,'' literally
allowing models to read and write from a simple
external tape memory, somewhat expanded the
transformers' ability to perform more complex logic-like
tasks 
\cite{nye2021, gravesNeuralTuringMachines2014}.
This approach, typically termed
``chain of thought'', has been widely explored in
attempts to advance large LFMs to logic-based tasks \cite{feng2023revealingmysterychainthought, wu2024rethinkingchainofthoughtperspectiveselftraining}. The ability to separately track and compare
alternate candidate inference steps may be a crucial
component in the expansion of transformers from
IXG-tier to supra-IXG tiers of abilities.

Attempts to adapt any abilities, such as chain-of-thought
capabilities, from large systems into smaller
models remain unsolved, leading to a ``small model
learnability gap'' \cite{liSmallModelsStruggle2025}.  We specifically propose that it is the
lower computational tier of the model (and the
corresponding absence of crucial computational
mechanisms), rather than the model size per se, that
prevents small LFMs from attaining the powers of
higher-tier systems.

It is crucial to test LFMs via out-of-distribution data,
yet evaluations that use completely non-linguistic
test material (such as arithmetic tests) may only
indirectly be making contact with the LFMs’ abilities.
It is equally important that test materials be
independently validated, rather than ``self-evaluation;''
such external validation is often difficult
but is integral to the generative language material
used in the assays we present here.

Solving even seemingly simple reasoning tasks  can require the 
rigorous exploration of multiple alternatives before
any solution can be settled on. In a crossword puzzle,
any candidate answer may affect subsequent entries,
so many possible alternatives must be considered,
in concert, all before any individual answer can be
finalized. This requires that all intermediate steps
follow rigorous rules and can store results intact for
subsequent testing. 
Even if a solution is not found for a given problem, adhering to logical rules at least prevents incorrect solutions from being proffered.

In sum, the rules governing many advanced tasks have been shown to
correspond to at least the CSG tier \cite{daveInvestigatingSymbolicCapabilities2024, deletangNeuralNetworksChomsky2023, pan2023logiclmempoweringlargelanguage},
requiring distinct architectures beyond LFMs
alone \cite{kambhampatiCanLargeLanguage2024, kambhampatiLLMsCantPlan2024,
moussad2023, Nguyen2024},   augmented 
with additional machinery for multiple separate
networks that can communicate with each other,
specifically incorporating ``new mechanisms to
exchange information within the ... representations that enable direct reasoning about the
spatial and evolutionary relationships'', i.e.,
separately evaluating different conformation
hypotheses and comparing them with each other \cite{Nguyen2024, jumperHighlyAccurateProtein2021}.)

In some architectural extensions of LFMs, the added
machinery either precedes or appends to
transformer-based foundational models (as in
current multi-modal architectures; see ``Supplementary Material A. LFMs and ATT LFMs in the G-A Hierarchy,''). By contrast, intercalated augmentation of
traditional transformer-based large language models (that is, ATT LFMs) 
can be seen, for example, in certain instances of chain-of-thought (CoT) \cite{wei2023chainofthoughtpromptingelicitsreasoning, chenReasoningEraSurvey2025}, mixture-of-experts (MoE) \cite{loCloserLookMixtureofExperts2025}, or memory-augmentations \cite{wuMemformerMemoryAugmentedTransformer2022}
integrated into transformer-based large LFMs. 
These intercalated augmentations appear to
measurably increase successes on particular formal
logic tasks, but it is not known whether these
represent a fundamental ascent from the IXG level
up to the 
complex logical reasoning
abilities of supra-IXG
tiers (LBAs to TMs).
 
It was shown that GPT-4 could generate acronyms from the first letter of
each word in a phrase (International Business
Machines to IBM) but struggled using the second
letter (NUA).  
These are equivalently easily achieved
by following a logical rule, but quite distinct if the
system is attempting to use statistically familiar
examples. By contrast, o1 reportedly could do both
more tokens for the rare task (2nd letters) than the
statistically familiar one. The evidence indicates that
the o1 suite of models expended extra work to get the
answer right -- as indeed we ourselves may do when
carrying out learned logical rules as opposed to
``intuiting'' an answer \cite{openai-reasoning-post}.
 
Analyses show that language foundation models (LFMs) fundamentally lack genuine basic formal logic abilities \cite{AppleGSM, nikankin2024arithmeticalgorithmslanguagemodels, jiang2024llmnotgenuinereasoner, shi2023llmseasilydistracted, schaeffer2023emergentabilitieslargelanguage, kassner-schutze-2020-negated, li2022largelanguagemodelscontrollable, jones2022capturingfailureslargelanguage, misra2023compsconceptualminimalpair, chaves-richter-2021-look,kambhampatiCanLargeLanguage2024, kambhampatiLLMsCantPlan2024}. The wide discrepancy between the continued lack of reasoning abilities and mastery of language is increasingly investigated, and may be closely related to the differential computational power of the distinct tiers of the G-A hierarchy. Careful generation of novel, out-of-distribution strings from distinct G-A tiers may provide unusually strict and compelling measures of the real abilities of AI architectures.

\subsection{Possible confluence of biological and artificial computations} 

Multiple findings in computational neuroscience identify possible confluences between biological and artificial language systems; these largely occur not at the mechanistic level, but at a computational or specification level.
\citet{schrimpfNeuralArchitectureLanguage2020} showed that the best analyses of neural predictivity in empirical human language processing data were from transformers.
Transformers' ability to `predict the next word' arises from their internal representation of the grammar information entailed in language processing \cite{tenney2019bertrediscoversclassicalnlp}.
\citet{tangSemanticReconstructionContinuous2023} used the same classes of models to decode a range of structured sequence information from recordings across multiple brain areas; \citet{mischlerContextualFeatureExtraction2024} found that the better the performance of a language model on benchmark tasks, the better it predicted neural responses and aligned with hierarchical feature extraction pathways of the brain. It also is notable that brain regions identified for language tasks and logic tasks correspond to the areas most increased in relative size in humans (see Figure \ref{fig:brainsizes}) \cite{goelAnatomyDeductiveReasoning2007}.

\begin{figure}[tb]
    \centering
    \includegraphics[width=0.8\linewidth]{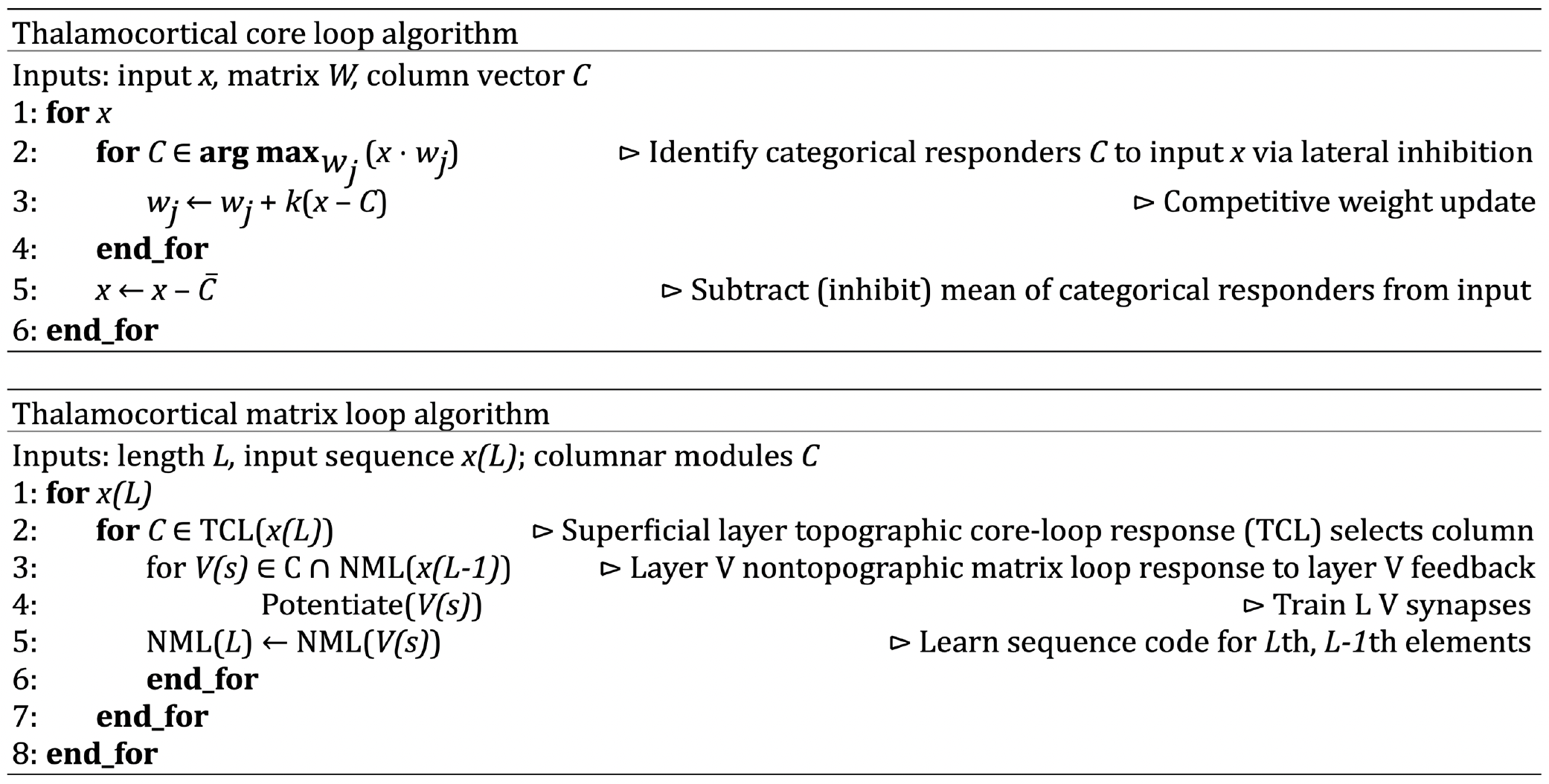}
    \caption{Algorithms derived from the two predominant components of
thalamocortical loops. Core thalamic cells (parvalbumin reactive) project
topographically to layer IV and III of specific cortical areas; these project to
superficial layer cells (II-III) $\rightarrow$ deep layers (VI) $\rightarrow$ core and nucleus
reticularis. These produce emergent hierarchical categorization. Matrix
thalamic cells (calbindin reactive) project broadly and non-topographically
to all cortical areas, synapsing in layer I on the apical dendrites of layer II,
III and V cells $\rightarrow$ matrix (non-topographic); these in turn produce emergent
hierarchical sequential chaining (from \cite{Rodriguez_2016}).}
    \label{fig:TCalgo}
\end{figure}

Analyses of the  circuitry of these brain regions have provided formulations across the implementation, algorithm, and computational levels 
(\cite{MarrVisionComputationalInvestigation1982}), illustrating where different substrate-level mechanisms may nonetheless be carrying out computations within the same 
equivalence class as 
those of certain artificial and formal systems.  
\citet{Rodriguez_2016} extrapolated from a substantial set of known constraints in 
physiology and  anatomy to present a specific hypothesis of both algorithms and computations arising from mammalian thalamocortical circuitry. The circuit consists of two well studied subcircuits: the core (or ``specific'') loop and the matrix (or ``nonspecific'') loop \cite{Rodriguez_2016}. These use topographic and nontopographic circuitry, respectively, to compute i) categories, i.e., embedding-like categorical encodings of individual cues, and ii) sequences, i.e., positional-encoding representations of successive cues (see Fig. \ref{fig:TCalgo}).

Recall our tree-adjoining grammar $G_{ix}$ defined in Section 3.2.
From the perspective of sequences of categories, we can view the $G_{ix}$ production rules $R$ as functions mapping from a category (non-terminal symbol $s \in V$) to a sequence of categories (non-terminal and terminal symbols $\{s_0, \dots, s_n\} \in \{V \cup \Sigma\}^n$ in order). Such symbolic processing and 
successive \emph{nesting} of category to sequence-of-category mappings offers a comparable procedure to the hypothesized iterative brain circuit update process described in algorithms of Figure \ref{fig:TCalgo}. The formal language perspective of both of the grammar generators for natural language tasks presented here, and the thalmocortical loop algorithm, may be viewed abstractly as nested production rule expressions, yielding successively nested sequences of categories
\cite{Rodriguez_2016}. 
(See also the relevant discussion in Supplementary Material F).  
 
These identified computations emerge not from individual neurons in isolation, nor from simplified mass operation of grouped neurons of similar types (as in most abstract neural networks). Rather, the computations derived by \citet{Rodriguez_2016} arise from complex interactions among multiple distinct neuron types arranged within the very well-studied anatomical circuit architectures of thalamocortical loops. Extensive evidence for the resulting emergent categorization and sequencing operations show up both in microscale and macro-scale readouts of brain activity \cite{pathakBiomimeticModelCorticostriatal2025}, as well as yielding to formal treatment to produce tractable algorithms (Figure \ref{fig:TCalgo}).

These results are highly suggestive of a
computation-level correspondence between brain architectures and
generative AI architectures.   
From detailed interactions at the wetware or ``implementation'' level of neurons and microassemblies of neurons, emerges a set of algorithm- and computation-level characterizations 
\cite{MarrVisionComputationalInvestigation1982}, that enables us to enlist the G-A hierarchy directly in the study of both biological and artificial systems.

\section{Conclusions}

Language models become not just syntactically fluent but broadly conversant across extensive domains. Such models appear to bear strong resemblances to human notions of ``everyday common sense knowledge'', which is surprising since the models have trained solely on natural language data, not on any mechanistic models nor embodied experience from which such common-sense inferential knowledge could have been educed. 
Much research investigates how this seemingly real-world knowledge could arise from mere text.  
Beyond the surface token stream, the deeper levels of a transformer accrue information into the residual stream of successive high dimensional internal states that autoregressively acquire rich semantic information (e.g., 
\cite{gevaTransformerFeedForwardLayers2022}).
They predict not just the ``next word'' but also (surprisingly) the next concept, as seen in mechanistic interpretability analyses (e.g., \cite{elhage2021mathematical}).
These fertile generative internal representations may converge to a stable statistical internal model of reality 
(\cite{pmlr-v235-huh24a}).  

Yet the resulting internal representations often differ from human knowledge: language models may fail to notice everyday physics such as gravity, object permanence, containment, or other world characteristics (see, e.g., \cite{AppleGSM, song2026largelanguagemodelreasoning}).  
Efforts to construct ``world models''
(e.g., \cite{HaWorldModels2018, matchinCorticalOrganizationSyntax2020, hafnerMasteringDiverseControl2025})
aim to overcome some of these shortfalls,
and also should be evaluated 
to see whether they rise above the powers of 
statistical models, to attain the logical reasoning capabilities of higher computational tiers. 
 
In general, the lack of established rigorous concepts and terminology for such divides exacerbates misunderstandings in the literature of how to characterize particular model shortcomings 
(e.g., \cite{AppleGSM, nikankin2024arithmeticalgorithmslanguagemodels, jiang2024llmnotgenuinereasoner, shi2023llmseasilydistracted, schaeffer2023emergentabilitieslargelanguage, kassner-schutze-2020-negated, li2022largelanguagemodelscontrollable, jones2022capturingfailureslargelanguage, misra2023compsconceptualminimalpair, chaves-richter-2021-look,kambhampatiCanLargeLanguage2024, kambhampatiLLMsCantPlan2024}). 
The incorporation of the  tier-based benchmarks proposed here from the theory of computation may be a step toward formalizing some of these crucial distinctions among capabilities.

In sum, 
\begin{itemize}
    \item Evidence indicates that reliable logical reasoning abilities across many tasks require higher G-A tiers than natural language rules do  \cite{deletangNeuralNetworksChomsky2023, pan2023logiclmempoweringlargelanguage, daveInvestigatingSymbolicCapabilities2024, hopcroft-ullman, hopcroftIntroductionAutomataTheory2007}. 
    \item Humans acquire these extended abilities (arithmetic, logic, etc.) via intensive specialized training, as opposed to the surprising ease with which natural language is acquired; and evidence indicates that these higher reasoning abilities  differentially engage brain areas distinct from those normally engaged for language tasks \cite{Fedorenko2024, Blank2016}.
 
    \item A new method for generative benchmarking of language models is introduced (section 2), encompassing generation of natural language, synthetic sequences, and arithmetic sequences, to rigorously test models' computational abilities.    

    \item The new benchmarks are shown to be directly applicable to existing language systems (section 3), providing  significant direct empirical evidence of where distinct language model capabilities reside in the G-A hierarchy, thereby providing a candidate method for distinguishing among models that can or cannot perform higher computational tasks beyond 
    language-based inference, such as logical reasoning.  Several previous individual tests from the literature are shown to be special cases of this new generative method.  
    
\end{itemize} 

We posit that the assays introduced here provide a principled formal approach to the evaluation of computational power of artificial language systems.  We provide code for symbolic AI generators that can quantitatively evaluate the computational power of a language system, and we apply that code to provide empirical verification
across a range of language models. 
(See the code repository and descriptions provided in Supplementary Material D.)

Further systematic studies of the acquisition of logical and mathematical abilities in humans may provide actionable insights into the ongoing algorithmic augmentation of current transformer-based AI systems from their seemingly native fluency in natural language, up to higher computational tiers of reliable  logic and reasoning abilities.

\begin{acks}
Thanks to Chelsea Joe and  Thang Nguyen. This work was supported in part by funds from  the Office of Naval Research. 
\end{acks}

\newpage
\appendix
\noindent\textbf{\huge Supplementary Material} \\

Section A details the empirical experiments from Section 3 of the manuscript

\ \ \ \ \ \ \ \ \ A.1 Methods 

\ \ \ \ \ \ \ \ \  A.2 Materials  

\ \ \ \ \ \ \ \ \  A.3 Statistical analyses

Section B details the strings used in the empirical experiments. 

Section C defines and describes the full formal grammar generators produced for the benchmark. 

Section D details the code and process used to generate outputs from the grammars defined in C. 

Section E provides empirical examples of  language model failures, and outlines ongoing errors in language models.  

Section F provides notes on other perspectives beyond the formal computational focus of this work. 

\ \ \ \ \ \ \ \ \ F.1 Work relating abstract machines at other analytical levels

\ \ \ \ \ \ \ \ \  F.2 Forms of learning  

\ \ \ \ \ \ \ \ \  F.3 Program induction from formal language strings

\ \ \ \ \ \ \ \ \  F.4 A note on the computational perspective of natural phenomena

Section G provides paired responses from GPT-5.5: binary grammaticality judgments, and then full parse tree brackets for the corresponding strings.

\newpage

\section{Experiments} \label{appendix:experiments}

\subsection{Methods}

From each of the grammars $G_{cf}, G_{ix}, \text{ and } G_{cs}$ defined in Supplementary Material \ref{appendix:defgrammars}, sentence strings were generated, and then $50$ sentences from each grammar were selected based on the criteria provided in \ref{appendix:sentences}. 

\paragraph{Language models}

Each language foundation model was prompted with the following question before the sentences.

\begin{quote}
\it
    Disregarding any lack of punctuation or sensicality, please answer the following with only ``Yes'' or ``No'' : Is the sentence grammatically complete? 
\end{quote}

Each model was called for each prompt-sentence pair, which forms the query. Each model was queried $100$ times for each of the $150$ tested strings ($50$ from each formal grammar generator.

\paragraph{Human participants}

Each of the human participants was given a questionnaire that began with the question:

\begin{quote}
    \it
    Disregarding any lack of punctuation or sensicality, please answer the following: Is the sentence grammatically complete?
\end{quote}

The question was followed by $75$ sentences, $25$ of which were generated from one of the three grammars, respectively, then a \textit{Yes} or a \textit{No} selection option. Later Supplementary Material create, describe, and give examples of the sentences of which the prompts refer. These sentences lack both punctuation and sense; they merely represent a minimal rule-based arrangement of English words, which leads to an effort to avoid thoughts of \emph{correctness} the decision of prompting with ``grammatically complete'' instead of the synonymous yet more ubiquitous phrase ``grammatically correct''. 

Due to the mental effort required to attempt to parse never-seen-before nonsensical sentences, each of the participants was given a sample sentence, not included in the test set, from each of the three grammars and was provided a review of the object-verb agreements and structures that make the sentences grammatically acceptable, or not, in natural English language.

\subsection{Materials}

\subsubsection{Language Foundation Models}

Eleven language foundation models (LFMs) were selected for their varying sizes and different training databases, which vary according to the companies that built and trained the models. Sorted by family, the LFMs are the following: Llama 3.1 (405b, 70b, 8b) \cite{llama3.1}, GPT (4 \cite{gpt4}, 3.5 Turbo \cite{gpt3.5-turbo}), Gemini (1.5 Flash \cite{gemini1.5}, 2.0 Flash \cite{blogIntroducingGemini20}), Claude 3 (Opus, Haiku) \cite{claude3}, Mistral 7b \cite{jiangMistral7B2023}, and DeepSeek R1 \cite{deepseek-aiDeepSeekR1IncentivizingReasoning2025}. The estimated model parameter sizes are presented in Table \ref{tab:lms}.

\begin{table}[htb]
\begin{center}
\renewcommand{\arraystretch}{2}
\begin{tabular}{lccccc}
    \textbf{Models} & Llama 3.1 405b & Llama 3.1 70b & Llama 3.1 8b & GPT-4 & GPT-3.5 Turbo \\
    \hline
    \textbf{Size Estimate} & 405 B & 70 B & 8 B & 1.7 T - 1.8 T & 20 B - 175 B \\
    \hline
\end{tabular}
\vspace{2mm}

\begin{tabular}{lcccccc}
    \textbf{Models} & Gemini 2.0 Flash & Gemini 1.5 Flash & Claude 3 Opus & Claude 3 Haiku & Mistral 7b & DeepSeek R1 \\
    \hline
    \textbf{Size Estimate} & 100 B - 200 B & 20 B - 32 B & 2 T & 20 B - 125 B & 7 B & 600 B - 700 B \\
    \hline
\end{tabular}
\caption{Language model parameter size estimates.}
\label{tab:lms}
\renewcommand{\arraystretch}{1}
\end{center}
\end{table}

The context windows for the models all fall within the range of 10,000 to 2,000,000. That said, the models also have smaller maximum token input sizes, and it is important to note that the limits did not impact the abilities of any of the models, since the total input tokens in all queries are about 60 tokens. While the output of just ``Yes'' or ``No'' was requested of the models in the query itself, the output token max was specified at 128 tokens for all models when queried. Additionally, the temperature parameter was set to 0.2, which is on the lower end of the stochasticity setting generally used by developers, yielding more deterministic outputs. All of the transformer-based language models above are decoder-only transformer architectures.

\subsubsection{Large Language Foundation Models with Augmented Transformer Architectures} 

Extensions through architectural additions that are integrated into the language foundation model (LFM) instead of separated or appended components at training or inference time, as in current multi-modal architectures. For large LFMs with this augmentation, let us call these augmented traditional transformer (ATT) large language foundation models: ATT LFMs.

An ATT LFM is a first definition of a broader space of specific hybrid architectures. The key criteria for an ATT LFM are (i) the \textit{augmentation} adds a mechanism that influences the `direction' of processing, and (ii) the mechanism is \textit{interpolated} at inference time so that it acts as a guide throughout processing (i.e., not preceding or afterward).

\begin{description}

    \item[o1 series]
OpenAI's three released o1 series models, o1, o1-preview and o1-mini, were selected to test language models with an additional mechanism integrated into a \textit{large} transformer-based architecture used to process language. In the case of the o1 series, the addition of large-scale reinforcement learning (RL) was used to guide the path of convergence toward a response
\cite{o1}. 

Due to the additional reasoning tokens, the maximum output tokens were not capped on these models, although the cap on the other models did not affect the requested single word output. The temperature was set to the same lower end of the typical development range, $0.2$.

 \item[Gemini 1.5 series]
Google's Gemini 1.5 Pro model ``is a sparse mixture-of-experts (MoE) Transformer-based'' large LM (LLM), specifically from the 1T Gemini 1.0 Pro language foundation model \cite{gemini1.5}. In this case, the added component is a (sparsely-gated) MoE layer embedded within the large transformer-based language foundation model architecture \cite{lepikhinGShardScalingGiant2020, zophSTMoEDesigningStable2022, shazeerOutrageouslyLargeNeural2017}. 

The temperature was set to the same lower end of the typical development stochasticity range, $0.2$. Note that while Gemini 1.5 Pro qualifies with the additional architectural component, its relative, Gemini 1.5 Flash, is a transformer-based language model that is not as ``large'' (over hundreds of billions of model parameters) and is online distilled from Gemini 1.5 Pro \cite{gemini1.5}.
\end{description}

\paragraph{Cost}

In total, the cost for querying and receiving the outputs of all the models totaled about $3000$ USD. The process was run remotely on an Nvidia server for convenience; however, the entire experiment (assuming the requisite API keys have adequate funds) can be completed in two days, when models are queried in parallel, on a 2022 Macbook Air.

\subsubsection{Human Participants}

The data collection from ten human participants was conducted via a Google Form, the specifics of its setup are described in the Section A.1 Methods. The human participants were undergraduate college students.

\subsection{Results}

\paragraph{Language models}

\begin{table}[htb]
\begin{center}
\renewcommand{\arraystretch}{2}
\begin{tabular}{lccccc}
    \textbf{Models} & Llama 3.1 405b & Llama 3.1 70b & Llama 3.1 8b & GPT-4 & GPT-3.5 Turbo \\
    \hline
    \textbf{CFG} & $ 0.977 \pm 0.02$ & $0.657 \pm 0.058$ & $0.033 \pm 0.018$ & $1.000 \pm 0.0$ & $0.003 \pm 0.002$ \\
    \textbf{IXG} & $0.637 \pm 0.068$ & $0.394 \pm 0.061$ & $0.0008 \pm 0.0006$ & $0.716 \pm 0.058$ & $0.0 \pm 0.0$ \\
    \textbf{CSG} & $0.021 \pm 0.02$ & $0.0 \pm 0.0$ & $0.0 \pm 0.0$ & $0.0 \pm 0.0$ & $0.0 \pm 0.0$ \\
    \hline
\end{tabular}
\vspace{2mm}

\begin{tabular}{lcccccc}
    \textbf{Models} & Gemini 2.0 Flash & Gemini 1.5 Flash & Claude 3 Opus & Claude 3 Haiku & Mistral 7b & DeepSeek R1 \\
    \hline
    \textbf{CFG} & $0.886 \pm 0.043$ & $0.253 \pm 0.059$ & $0.918 \pm 0.038$ & $0.125 \pm 0.106$ & $0.299 \pm 0.061$ & $0.8 \pm 0.2 \; ^{(*)}$ \\
    \textbf{IXG} & $0.736 \pm 0.062$ & $0.0 \pm 0.0$ & $0.845 \pm 0.048$ & $0.02 \pm 0.019$ & $0.020 \pm 0.014$ & $0.6 \pm 0.3 \; ^{(*)}$ \\
    \textbf{CSG} & $0.02 \pm 0.019$ & $0.0 \pm 0.0$ & $0.167 \pm 0.052$ & $0.0 \pm 0.0$ & $0.02 \pm 0.019$ & $0.0 \pm 0.0 \; ^{(*)}$ \\
    \hline
\end{tabular}
\renewcommand{\arraystretch}{1}
\caption{The average and standard error range for the acceptance of the sentences by the LMs, where each column gives the model. The sentences are generated by either the context-free grammar (CFG) $G_{cf}$, the indexed grammar (IXG) $G_{ix}$, and the context-sensitive grammar (CSG) $G_{cs}$, distinguish each row. (See Supplementary Material \ref{appendix:defgrammars} for definitions of $C_{cf}, C_{ix}, C_{cs}$.)}
\label{tab:modresults}
\end{center}
\end{table}

Table \ref{tab:modresults} presents the percentage of accepted sentences for each of the models, over 100 trials, with a precision range given by the standard error. Note that, due to query restrictions, DeepSeek R1 was averaged over 10 trials and is marked with the $^{(*)}$ symbol. The full .csv file of the output statistics used to create the 3D bar figure in the main paper, Figure 3, can be found in the datasets.

\paragraph{Augmented Traditional Transformer (ATT) Large Language models}

Table \ref{tab:o1results} gives the percentage of accepted sentences for each of the o1-series models \cite{o1}, with a precision range provided by the standard error.

\begin{table}[htb]
\begin{center}
\renewcommand{\arraystretch}{2}
\begin{tabular}{lcccc}
    \textbf{Models} & Gemini 1.5 Pro & o1-mini & o1-preview & o1  \\
    \hline
    \textbf{CFG} & $1.000 \pm 0.000$ & $0.987 \pm 0.002$ & $ 0.992 \pm 0.002$ & $0.996 \pm 0.001$ \\
    \textbf{IXG} & $0.88 \pm 0.046$ & $0.946 \pm 0.018$ & $0.960 \pm 0.023$ & $0.934 \pm  0.031$ \\
    \textbf{CSG}  & $0.261 \pm 0.062$ & $0.333 \pm 0.037$ & $0.442 \pm 0.041$ &  $0.029 \pm 0.021$ \\
    \hline
\end{tabular}
\renewcommand{\arraystretch}{1}
\caption{The average and standard error range for the acceptance of the sentences by the o1-series, LLMs with large-scale reinforcement learning integration, and Gemini 1.5 Pro, LLM with sparsely-gated mixture-of-experts integration, where each column gives the model. The sentences are generated by either the context-free grammar (CFG) $G_{cf}$, the indexed grammar (IXG) $G_{ix}$, and the context-sensitive grammar (CSG) $G_{cs}$), distinguish each row.}
\label{tab:o1results}
\end{center}
\end{table}

The statistically significant increase in acceptance of context-sensitive (CS) sentences is notable and brings the Gemini 1.5 Pro, o1-mini, and o1-preview into the distinctive range of \textit{over a fourth} of the CS sentences accepted on average, while o1, interestingly, did not. Note the strong implications of such nonsensical and ungrammatical (in any natural human language) ordering of words with seemingly induced agreements and dependencies between them in the form of a string, which can be seen as a \textit{sentence} of sorts, being accepted as ``grammatically complete'' by these models but not with LMs without the additional mechanism integrated directly into the transformer-based architecture used to process language.

\paragraph{Human participants}

The human participant sentence acceptances for each formal grammar class of the grammar that generated the sentence strings are given in Table \ref{tab:humans}.

\begin{table}
    \centering
    \renewcommand{\arraystretch}{2}
    \begin{tabular}{lccc}
        \textbf{Grammar Class} & CFG & IXG & CSG \\
        \hline
        \textbf{Participant 1} & $0.96 \pm 0.20$ & $1.00 \pm 0.00$ & $0.04 \pm 0.20$ \\
        \textbf{Participant 2} & $1.00 \pm 0.00$ & $0.96 \pm 0.20$ & $0.00 \pm 0.00$ \\
        \textbf{Participant 3} & $1.00 \pm 0.00$ & $0.92 \pm 0.27$ & $0.00 \pm 0.00$ \\
        \textbf{Participant 4} & $1.00 \pm 0.00$ & $0.96 \pm 0.20$ & $0.00 \pm 0.00$ \\
        \textbf{Participant 5} & $1.00 \pm 0.00$ & $0.92 \pm 0.27$ & $0.04 \pm 0.20$ \\
        \textbf{Participant 6} & $1.00 \pm 0.00$ & $0.96 \pm 0.20$ & $0.04 \pm 0.20$ \\
        \textbf{Participant 7} & $0.96 \pm 0.20$ & $0.64 \pm 0.48$ & $0.16 \pm 0.37$ \\
        \textbf{Participant 8} & $0.64 \pm 0.48$ & $0.48 \pm 0.50$ & $0.32 \pm 0.47$ \\
        \textbf{Participant 9} & $0.96 \pm 0.20$ & $0.84 \pm 0.37$ & $0.00 \pm 0.00$ \\
        \textbf{Participant 10} & $1.00 \pm 0.00$ & $0.80 \pm 0.40$ & $0.00 \pm 0.00$ \\
        \textbf{Average Acceptance} & $\mathbf{0.95 \pm 0.05}$ & $\mathbf{0.84 \pm 0.14}$ & $\mathbf{0.06 \pm 0.08}$ \\
        \hline
    \end{tabular}
    \caption{Average sentence acceptance, and standard deviation range, of each human participant for 25 sentences generated by the context-free grammar (CFG) $G_{cf}$, the indexed grammar (IXG) $G_{ix}$, and the context-sensitive grammar (CSG) $G_{cs}$, respectively, are seen in each Participant row. The average acceptance per grammar of all of the participants is given in the final row with the standard error range.}
    \label{tab:humans}
\end{table}

Together, the participants vastly accepted the sentences generated by context-free and indexed grammars and rejected the sentences generated by a context-sensitive grammar, in correspondence with the formal grammar classification of natural human languages. The variations, small but present, can be attributed to the participants' need to ignore the lack of punctuation and the nonsensicality, which consequently makes the sentences more consuming to parse.

\subsubsection{\bf Statistical Analysis}

We treated model size as a four-level between-model acceptance factor (Small, Mid, Large, Augmented) and grammatical tier as a three-level within-model acceptance factor (CFG, IXG, CSG). Tier effects were evaluated using Friedman tests across model-level tier scores, followed by Holm-corrected Wilcoxon signed-rank post hoc tests. Model size-class effects were evaluated separately within each tier using Kruskal–Wallis tests across model-level scores, followed by Holm-corrected Mann–Whitney U tests.

\paragraph{Tier effects}

Table \ref{tab:friedman-tier-across-models} provides the Friedman test results across model responses for each grammar tier. The Friedman test is the non-parametric analogue of a repeated-measures ANOVA: it tests whether, after blocking on the model, the rank distributions differ across the three related tier conditions. A significant result indicates that at least one tier differs systematically from the others. Kendall's \(W\) gives the corresponding effect size, with values closer to \(1\) indicating stronger agreement in the ordering of tiers across models.

\begin{table}[htbp]
\renewcommand{\arraystretch}{1.25}
\centering
\caption{Friedman test for tier effect across all models.}
\label{tab:friedman-tier-across-models}
\begin{tabular}{llclrrr}
\toprule
Test & Blocks & \(n\) models & Tiers & Statistic & \(p\)-value & Kendall's \(W\) \\
\midrule
Friedman & model & $16$ & CFG, IXG, CSG & $31.096774$ & $1.767752e-07$ & $0.971774$ \\
\bottomrule
\end{tabular}
\end{table}

The omnibus Friedman result in Table \ref{tab:friedman-tier-across-models} is highly significant, showing that the three grammar tiers do not elicit equivalent model-level responses. The effect size is also extremely large (\(W = 0.971774\)), indicating consistency in the ranking of tiers across models. In other words, the models do not vary randomly across CFG, IXG, and CSG; they exhibit a strong and systematic tier-dependent response pattern.

To assess whether this tier effect also holds within each model-size class, Table \ref{tab:friedman-tier-within-size} reports separate Friedman tests for Small, Medium, Large, and Augmented models. These tests have the same interpretation as the global Friedman test: significance indicates that, within a given size class, at least one tier differs from the others.

\begin{table}[htbp]
\centering
\renewcommand{\arraystretch}{1.25}
\caption{Friedman tests for grammar generator tier effect within each size class.}
\label{tab:friedman-tier-within-size}
\begin{tabular}{lcrrr}
\toprule
Size class & \(n\) models & Statistic & \(p\)-value & Kendall's \(W\) \\
\midrule
Small & $3$ & $6.000000$ & $0.049787 $& $1.000000$ \\
Medium & $5$ & $9.333333$ & $0.009404$ & $0.933333$ \\
Large & $4$ & $8.000000$ & $0.018316$ & $1.000000$ \\
Augmented & $4$ & $8.000000$ & $0.018316$ & $1.000000$ \\
\bottomrule
\end{tabular}
\end{table}

Table \ref{tab:friedman-tier-within-size} shows significant omnibus tier effects within every size class. The Small class is just below the conventional threshold (\(p = 0.049787\)), whereas the Medium, Large, and Augmented classes are more clearly significant. The effect sizes are very large throughout (\(W = 0.933333\) to \(1.000000\)), implying that the within-class ordering of tiers is highly consistent. At the same time, these within-size omnibus results should be interpreted somewhat cautiously because each size class contains only \(3\) to \(5\) models.

Table \ref{tab:tier-pairwise-wilcoxon} reports the post hoc Wilcoxon signed-rank tests for pairwise tier comparisons across all models. The Wilcoxon signed-rank test is a paired non-parametric test for two related conditions; here it compares two tiers measured on the same set of models. A pairwise contrast is significant when the Holm-adjusted \(p\)-value is below \(\alpha\), indicating a systematic paired difference between the two tiers. The rank-biserial coefficient summarizes the direction and magnitude of that difference.

\begin{table}[htbp]
\centering
\renewcommand{\arraystretch}{1.25}
\caption{Pairwise Wilcoxon signed-rank tests comparing grammar generator tiers across all models.}
\label{tab:tier-pairwise-wilcoxon}
\begin{tabular}{lccrcrr}
\toprule
Comparison & \(n\) models & Wilcoxon \(W\) & Raw \(p\) & Rank-biserial & Median diff. & Holm-adjusted \(p\) \\
\midrule
CFG vs IXG & $16$ & $0.0$ & $0.000031$ & $1.0$ & $0.134844$ & $0.000092$ \\
CFG vs CSG & $16$ & $0.0$ & $0.000031$ & $1.0$ & $0.676651$ & $0.000092$ \\
IXG vs CSG & $16$ & $0.0$ & $0.000982$ & $1.0$ & $0.567679$ & $0.000982$ \\
\bottomrule
\end{tabular}
\end{table}

All three pairwise tier contrasts in Table \ref{tab:tier-pairwise-wilcoxon} remain significant after Holm correction. The positive median differences indicate that the first-named tier in each comparison has a higher score than the second-named tier. The results show the global ordering is \( \mathrm{CFG} > \mathrm{IXG} > \mathrm{CSG} \). The rank-biserial value of \(1.0\) for every contrast indicates complete directional separation in the paired differences across models.

Table \ref{tab:tier-pairwise-wilcoxon-within-size} presents the same Wilcoxon signed-rank comparisons separately within each size class. As above, a result is considered significant when the Holm-adjusted $p$-value is below $\alpha$; significant results would identify which specific tier pairs are responsible for the omnibus within-class Friedman effects.

\begin{table}[htbp]
\centering
\renewcommand{\arraystretch}{1.25}
\scriptsize
\caption{Pairwise Wilcoxon signed-rank tests comparing grammar generator tiers within each size class.}
\label{tab:tier-pairwise-wilcoxon-within-size}
\resizebox{\textwidth}{!}{%
\begin{tabular}{llccrcrr}
\toprule
Size class & Comparison & \(n\) models & Wilcoxon \(W\) & Raw \(p\) & Rank-biserial & Median diff. & Holm-adjusted \(p\) \\
\midrule
Small & CFG vs IXG & $3$ & $0.0$ & $0.2500$ & $1.0$ & $0.278800$ & $0.7500$ \\
Small & CFG vs CSG & $3$ & $0.0$ & $0.2500$ & $1.0$ & $0.279400$ & $0.7500$ \\
Small & IXG vs CSG & $3$ & $0.0$ & $0.2500$ & $1.0$ & $0.000800$ & $0.7500$ \\
Medium & CFG vs IXG & $5$ & $0.0$ & $0.0625$ & $1.0$ & $0.149688$ & $0.1875$ \\
Medium & CFG vs CSG & $5$ & $0.0$ & $0.0625$ & $1.0$ & $0.253200$ & $0.1875$ \\
Medium & IXG vs CSG & $5$ & $0.0$ & $0.2500$ & $1.0$ & $0.020000$ & $0.2500$ \\
Large & CFG vs IXG & $4$ & $0.0$ & $0.1250$ & $1.0$ & $0.256167$ & $0.3750$ \\
Large & CFG vs CSG & $4$ & $0.0$ & $0.1250$ & $1.0$ & $0.901948$ & $0.3750$ \\
Large & IXG vs CSG & $4$ & $0.0$ & $0.1250$ & $1.0$ & $0.648781$ & $0.3750$ \\
Augmented & CFG vs IXG & $4$ & $0.0$ & $0.1250$ & $1.0$ & $0.053613$ & $0.3750$ \\
Augmented & CFG vs CSG & $4$ & $0.0$ & $0.1250$ & $1.0$ & $0.698851$ & $0.3750$ \\
Augmented & IXG vs CSG & $4$ & $0.0$ & $0.1250$ & $1.0$ & $0.616751$ & $0.3750$ \\
\bottomrule
\end{tabular}
}
\end{table}

None of the within-size pairwise contrasts in Table \ref{tab:tier-pairwise-wilcoxon-within-size} survive Holm correction. This is not necessarily evidence against a tier effect as it may mostly reflect the very small number of models available within each size class, which makes the exact signed-rank tests conservative and yields coarse $p$-value resolution. Importantly, the signs of the median differences are still broadly aligned with the global ordering \( \mathrm{CFG} > \mathrm{IXG} > \mathrm{CSG} \), although the Small model size class shows almost no separation between IXG and CSG---corresponding to the lack of acceptance of the strings generated at these tiers.

Taken together, the tier-level analyzes show a strong and highly consistent effect of grammatical tier. Across all models, the tier effect is unequivocal and all pairwise differences are significant, establishing the ordering \( \mathrm{CFG} > \mathrm{IXG} > \mathrm{CSG} \). The within-size omnibus tests further indicate that this pattern is present inside each size class, but the limited number of models per class leaves the within-class pairwise tests underpowered after multiple-comparison correction.

\paragraph{Model size level effects}

Table \ref{tab:kruskal-size-within-tier} provides Kruskal--Wallis tests of model-size class differences within each tier. The Kruskal--Wallis test is the non-parametric analogue of a one-way ANOVA for independent groups. Here it asks whether model-level string acceptances differ across the four model size classes within a fixed tier. A significant result indicates that at least one size class differs from at least one other. The reported $\epsilon^2$ values summarize the effect size, with larger values indicating stronger size-related separation.

\begin{table}[htbp]
\centering
\renewcommand{\arraystretch}{1.25}
\caption{Kruskal--Wallis tests for size-class effect within each grammar generator tier.}
\label{tab:kruskal-size-within-tier}
\begin{tabular}{llcrrr}
\toprule
Tier & Model size classes & \(n\) models & \(H\) & \(p\)-value & \(\epsilon^2\) \\
\midrule
CFG & Small, Medium, Large, Augmented & $16$ & $11.167526$ & $0.010854$ & $0.680627$ \\
IXG & Small, Medium, Large, Augmented & $16$ & $11.292342$ & $0.010246$ & $0.691028$ \\
CSG & Small, Medium, Large, Augmented & $16$ & $10.227127$ & $0.016731$ & $0.602261$ \\
\bottomrule
\end{tabular}

\end{table}

All three Kruskal--Wallis tests in Table \ref{tab:kruskal-size-within-tier} are significant, indicating that model size class matters within CFG, IXG, and CSG alike. The effect sizes are large ($\epsilon^2$ ranging from $0.602261$ to $0.691028$), suggesting that a substantial portion of the between-model variation within each tier is associated with each model size class. Even before looking at pairwise contrasts, there is clear model-level evidence that larger-capacity and augmented transformer-based systems respond differently from smaller ones.

Table \ref{tab:item-blocked-friedman} reports a sensitivity analysis that treats each item (string acceptance response per grammar) as a matched block and compares model size classes across the same complete set of $5000$ items within each grammar generator tier. Recall that there are $5000$ string responses per grammar (items) because each model was queried $100$ times for each of the $50$ selected string generations from each grammar, and models, such as DeepSeek R1 that did not have $5000$ string responses due to the limited number or queries per string, is not included in the analysis of Table \ref{tab:item-blocked-friedman}.
This item-blocked Friedman test controls for item-specific difficulty and asks whether the size classes differ consistently when evaluated on matched items. A significant result indicates a robust size effect that is not merely an artifact of aggregation over models.

\begin{table}[htbp]
\centering
\renewcommand{\arraystretch}{1.25}
\caption{Item-blocked Friedman sensitivity analyses for size-class effect within each grammar generator tier.}
\label{tab:item-blocked-friedman}
\begin{tabular}{lrlrrr}
\toprule
Tier & Items & Model size classes & Statistic & \(p\)-value & Kendall's \(W\) \\
\midrule
CFG &  $5000$ & Small, Medium, Large, Augmented & $13027.261975$ & $0.0$ & $0.868484$ \\
IXG & $5000$ & Small, Medium, Large, Augmented & $12332.466944$ & $0.0$ & $0.822164$ \\
CSG & $5000$ & Small, Medium, Large, Augmented & $7470.965649$ & $0.0$ & $0.498064$ \\
\bottomrule
\end{tabular}%
\end{table}

The item-blocked sensitivity results in Table \ref{tab:item-blocked-friedman} are overwhelmingly significant in all three tiers, with $p$-values reported as effectively zero. The corresponding Kendall's \(W\) values indicate very strong size effects for CFG and IXG, and a still substantial effect for CSG. This confirms that size-class differences are robust at the string acceptance response per grammar level and are not dependent on how the model-level summaries were aggregated.

Table \ref{tab:item-blocked-pairwise-wilcoxon} gives the item-blocked pairwise Wilcoxon signed-rank tests. These are paired post hoc tests on matched items (string acceptance responses), comparing two size classes at a time within a tier. A contrast is significant when the Holm-adjusted $p$-value is below $\alpha$, indicating that one size class systematically produces higher item-level responses than the other. Rank-biserial values near $\pm 1$ indicate near-complete directional separation.

\begin{table}[htbp]
\centering
\renewcommand{\arraystretch}{1.25}
\small
\caption{Pairwise Wilcoxon signed-rank tests for item-blocked size-class comparisons within each grammar generator tier.}
\label{tab:item-blocked-pairwise-wilcoxon}
\resizebox{\textwidth}{!}{%
\begin{tabular}{llrrrrrr}
\toprule
Tier & Comparison & Items & Wilcoxon \(W\) & Raw \(p\) & Rank-biserial & Median diff. & Holm-adjusted \(p\) \\
\midrule
CFG & Small vs Medium & $5000$ & $4531583.0$ & $2.422039e-48$ & $-0.240874$ & $-0.066667$ & $2.422039e-48$ \\
CFG & Small vs Large & $5000$ & $48.0$ & $0.000000e+00$ & $-0.999992$ & $-0.666667$ & $0.000000e+00$ \\
CFG & Small vs Augmented & $5000$ & $0.0$ & $0.000000e+00$ & $-1.000000$ & $-0.666667$ & $0.000000e+00$ \\
CFG & Medium vs Large & $5000$ & $180.0$ & $0.000000e+00$ & $-0.999971$ & $-0.600000$ & $0.000000e+00$ \\
CFG & Medium vs Augmented & $5000$ & $0.0$ & $0.000000e+00$ & $-1.000000$ & $-0.600000$ & $0.000000e+00$ \\
CFG & Large vs Augmented & $5000$ & $7969.0$ & $1.930986e-102$ & $-0.926265$ & $0.000000$ & $3.861972e-102$ \\
IXG & Small vs Medium & $5000$ & $1403220.0$ & $0.000000e+00$ & $-0.678988$ & $-0.200000$ & $0.000000e+00$ \\
IXG & Small vs Large & $5000$ & $18342.0$ & $0.000000e+00$ & $-0.996536$ & $-0.666667$ & $0.000000e+00$ \\
IXG & Small vs Augmented & $5000$ & $2340.0$ & $0.000000e+00$ & $-0.999624$ & $-1.000000$ & $0.000000e+00$ \\
IXG & Medium vs Large & $5000$ & $101217.5$ & $0.000000e+00$ & $-0.982519$ & $-0.600000$ & $0.000000e+00$ \\
IXG & Medium vs Augmented & $5000$ & $2366.0$ & $0.000000e+00$ & $-0.999621$ & $-0.750000$ & $0.000000e+00$ \\
IXG & Large vs Augmented & $5000$ & $123288.5$ & $0.000000e+00$ & $-0.933563$ & $0.000000$ & $0.000000e+00$ \\
CSG & Small vs Medium & $5000$ & $5050.0$ & $2.989053e-10$ & $0.497512$ & $0.000000$ & $2.989053e-10$ \\
CSG & Small vs Large & $5000$ & $57600.0$ & $2.684240e-132$ & $-0.790022$ & $0.000000$ & $5.368480e-132$ \\
CSG & Small vs Augmented & $5000$ & $3.0$ & $0.000000e+00$ & $-0.999999$ & $-0.250000$ & $0.000000e+00$ \\
CSG & Medium vs Large & $5000$ & $0.0$ & $1.009390e-178$ & $-1.000000$ & $0.000000$ & $3.028170e-178$ \\
CSG & Medium vs Augmented & $5000$ & $0.0$ & $0.000000e+00$ & $-1.000000$ & $-0.250000$ & $0.000000e+00$ \\
CSG & Large vs Augmented & $5000$ & $325277.0$ & $0.000000e+00$ & $-0.883959$ & $-0.250000$ & $0.000000e+00$ \\
\bottomrule
\end{tabular}%
}
\end{table}

All item-blocked pairwise comparisons in Table \ref{tab:item-blocked-pairwise-wilcoxon} are significant after Holm correction. For CFG and IXG, the pattern is especially clear: Small and Medium models are systematically below Large and Augmented models, and Augmented models are typically at least as strong as Large models. For CSG, several comparisons have a median difference of $0$, reflecting strong floor effects, but the very small $p$-values and large rank-biserial magnitudes still show systematic distributional differences. In other words, even where medians tie at floor, the full item-level distributions remain separated by size class.

Table \ref{tab:overall-size-kruskal} reports a supplementary Kruskal--Wallis test on overall equal-tier means collapsed across grammar tiers. This omnibus test asks whether average performance differs across size classes after collapsing over CFG, IXG, and CSG.

\begin{table}[htbp]
\centering
\renewcommand{\arraystretch}{1.25}
\caption{Supplementary overall Kruskal--Wallis test across model size classes.}
\label{tab:overall-size-kruskal}
\begin{tabular}{llrrrr}
\toprule
Test & Size classes & \(n\) models & \(H\) & \(p\)-value & \(\epsilon^2\) \\
\midrule
Kruskal-Wallis & Small, Medium, Large, Augmented & 16 & 11.75 & 0.008291 & 0.729167 \\
\bottomrule
\end{tabular}%
\end{table}

This supplementary analysis is also significant, indicating that the model class affects overall performance even when the three grammar generator tiers are averaged together; the large $\epsilon^2$ value suggests a substantial model size-related effect.

Table \ref{tab:pairwise-mannwhitney-within-tier} analyzes which specific pairs of independent size classes differ within each tier at the model level. The Mann--Whitney $U$ test is the non-parametric test for two independent groups. A contrast is significant when the Holm-adjusted $p$-value is below $\alpha$, indicating that the two size classes differ in their rank distributions within that tier. Cliff's $\delta$ summarizes effect direction and magnitude, with values near $\pm 1$ indicating strong separation.

\begin{table}[htbp]
\centering
\renewcommand{\arraystretch}{1.25}
\caption{Pairwise Mann--Whitney \(U\) tests for size-class comparisons within each grammar generator tier.}
\label{tab:pairwise-mannwhitney-within-tier}
\begin{tabular}{llccrrrrr}
\toprule
Tier & Comparison & \(n_a\) & \(n_b\) & \(U\) & Raw \(p\) & Cliff's \(\delta\) & Median diff. & Holm-adjusted \(p\) \\
\midrule
CFG & Small vs Medium & $3$ & $5$ & $8.0$ & $1.000000$ & $0.066667$ & $0.046200$ & 1.000000 \\
CFG & Small vs Large & $3$ & $4$ & $0.0$ & $0.057143$ & $-1.000000$ & $-0.648566$ & $0.228571$ \\
CFG & Small vs Augmented & $3$ & $4$ & $0.0$ & $0.057143$ & $-1.000000$ & $-0.696598$ & $0.228571$ \\
CFG & Medium vs Large & $5$ & $4$ & $1.0$ & $0.031746$ & $-0.900000$ & $-0.694766$ & $0.158730$ \\
CFG & Medium vs Augmented & $5$ & $4$ & $0.0$ & $0.015873$ & $-1.000000$ & $-0.742798$ & $0.095238$ \\
CFG & Large vs Augmented & $4$ & $4$ & $3.5$ & $0.245383$ & $-0.562500$ & $-0.048032$ & $0.490767$ \\
IXG & Small vs Medium & $3$ & $5$ & $8.0$ & $1.000000$ & $0.066667$ & $0.000600$ & $1.000000$ \\
IXG & Small vs Large & $3$ & $4$ & $0.0$ & $0.057143$ & $-1.000000$ & $-0.656300$ & $0.228571$ \\
IXG & Small vs Augmented & $3$ & $4$ & $0.0$ & $0.057143$ & $-1.000000$ & $-0.920971$ & $0.228571$ \\
IXG & Medium vs Large & $5$ & $4$ & $3.0$ & $0.109855$ & $-0.700000$ & $-0.656900$ & $0.228571$ \\
IXG & Medium vs Augmented & $5$ & $4$ & $0.0$ & $0.019451$ & $-1.000000$ & $-0.921571$ & $0.116706$ \\
IXG & Large vs Augmented & $4$ & $4$ & $0.0$ & $0.028571$ & $-1.000000$ & $-0.264671$ & $0.142857$ \\
CSG & Small vs Medium & $3$ & $5$ & $8.5$ & $0.843669$ & $0.133333$ & $0.000000$ & $0.843669$ \\
CSG & Small vs Large & $3$ & $4$ & $3.0$ & $0.359012$ & $-0.500000$ & $-0.012586$ & $0.718024$ \\
CSG & Small vs Augmented & $3$ & $4$ & $0.0$ & $0.049746$ & $-1.000000$ & $-0.297534$ & $0.248730$ \\
CSG & Medium vs Large & $5$ & $4$ & $4.0$ & $0.139996$ & $-0.600000$ & $-0.012586$ & $0.419989$ \\
CSG & Medium vs Augmented & $5$ & $4$ & $0.0$ & $0.015078$ & $-1.000000$ & $-0.297534$ & $0.090471$ \\
CSG & Large vs Augmented & $4$ & $4$ & $1.0$ & $0.057143$ & $-0.875000$ & $-0.284948$ & $0.248730$ \\
\bottomrule
\end{tabular}%
\end{table}

At the model level, none of the pairwise size-class comparisons within each formal grammar tier remains significant after Holm correction. Nevertheless, several raw $p$-values are small and many Cliff's $\delta$ values are large in magnitude, especially for comparisons involving Small or Medium versus Large or Augmented models. The directional pattern strongly suggests better performance for larger-capacity systems, but the small number of models per class limits pairwise power once multiple testing is controlled.

Table \ref{tab:overall-size-pairwise-mann} analyzes the same kind of pairwise Mann--Whitney contrasts, but now using each model's overall equal-tier mean collapsed across tiers. These pairwise tests are significant when the Holm-adjusted $p$-value falls below $\alpha$, which would indicate a reliable overall difference between the two model size classes.

\begin{table}[htbp]
\centering
\renewcommand{\arraystretch}{1.25}
\caption{Supplementary pairwise Mann--Whitney \(U\) tests across size classes using overall equal grammar generator tier means.}
\label{tab:overall-size-pairwise-mann}
\begin{tabular}{lrrrrrrr}
\toprule
Comparison & \(n_a\) & \(n_b\) & \(U\) & Raw \(p\) & Cliff's \(\delta\) & Median diff. & Holm-adjusted \(p\) \\
\midrule
Small vs Medium & $3$ & $5$ & $7.0$ & $1.000000$ & $-0.066667$ & $0.028933$ & $1.000000$ \\
Small vs Large & $3$ & $4$ & $0.0$ & $0.057143$ & $-1.000000$ & $-0.445467$ & $0.228571$ \\
Small vs Augmented & $3$ & $4$ & $0.0$ & $0.057143$ & $-1.000000$ & $-0.622734$ & $0.228571$ \\
Medium vs Large & $5$ & $4$ & $2.0$ & $0.063492$ & $-0.800000$ & $-0.474400$ & $0.228571$ \\
Medium vs Augmented & $5$ & $4$ & $0.0$ & $0.015873$ & $-1.000000$ & $-0.651668$ & $0.095238$ \\
Large vs Augmented & $4$ & $4$ & $0.0$ & $0.028571$ & $-1.000000$ & $-0.177268$ & $0.142857$ \\
\bottomrule
\end{tabular}%
\end{table}

As with the within-tier model-level comparisons, no pairwise overall size contrast survives Holm correction in Table \ref{tab:overall-size-pairwise-mann}. The raw \(p\)-values and effect-size estimates again point in the same direction: Large and Augmented models tend to accept more generated strings than Small and Medium models, with the clearest raw contrasts involving Augmented versus Medium and Large versus Augmented. However, there is a failure to retain significance after correction, although this may be understood as a small $n_a$ and $n_b$ sample size limitation rather than as evidence of no size effect.

Tables \ref{tab:pairwise-mannwhitney-within-tier} and \ref{tab:overall-size-pairwise-mann} show that model-level pairwise size comparisons are directionally consistent with a monotonic model size level effect---larger and augmented systems generally accept more than smaller systems---but most individual pairwise contrasts are underpowered after multiple-comparison correction. The item-blocked analyses provide the strongest evidence that the size effect is real and robust.

\paragraph{Statistical summary.}
Across all 16 models, there was a very strong effect of tier, with a Friedman test showing clear differences among CFG, IXG, and CSG (\(31.10\), \(p = 1.77 \times 10^{-7}\), Kendall's \(W = 0.972\)). Holm-corrected pairwise Wilcoxon tests showed that all three tier contrasts were significant, indicating the overall ordering of \emph{acceptance} of strings generated by grammars of the respective classes: \( \mathrm{CFG} > \mathrm{IXG} > \mathrm{CSG} \). Within each model size class, omnibus Friedman tests were also significant, but the within-class pairwise tests did not survive Holm correction, likely because the number of models per size class was small. Size class also mattered within each tier: Kruskal--Wallis tests were significant for CFG, IXG, and CSG, with large effect sizes (\(\epsilon^2 \approx 0.60\) to \(0.69\)). Although model-level pairwise size comparisons were not significant after multiple-comparison correction, the item-blocked sensitivity analyses over $5000$ items per tier showed strong and consistent size effects, supporting the interpretation that augmented traditional transformer-based (ATT) foundation models and large language foundation models generally accept more generated strings than mid-sized and small language foundation models, especially on CFG and IXG, while CSG is the least accepted tier overall.

\newpage
\section{Sentence Selection from Generated Sentences} \label{appendix:sentences}

To align closer with the sentences found in the specific grammar that generates natural (human) English language, all sentences generated for selection are in the range of 10 to 21 total words. Additionally, it can be seen in the terminal symbols (defined in Supplementary Material \ref{appendix:defgrammars}) that the singular determiners include the words ``the'' and ``a''. Since there are some singular nouns that start with vowels and would be preceded with ``an'' not ``a'' in natural English, ``a'' was changed to ``an'' in the few cases where the noun (e.g., ``ocean'') necessitates the change.

For the context-free sentences, only sentences with tense agreement, between the object(s) and verb(s), were selected. Here we will restate that higher levels of formal grammar and automata classes subsume the lower levels. Consequently, for example, indexed grammars can produce both indexed strings and context-free strings. 

Out of the indexed sentences generated, sentences with enough dependencies to require the recognition of nested stack automata were chosen. Additionally, to better correspond with natural English, only sentences where the relative pronoun ``who'' followed the proper nouns and sentences that had the relative pronouns ``which'' and ``that'' corresponding with their objects were selected. 

For the context-sensitive language, sentences with a maximum recursion depth of ten and a maximum expansion per symbol of twenty, before a forced path to terminals is traversed, were generated to increase the probability that the output strings requires a linear bounded automaton for recognition. Out of these generated strings, sentences with dependencies exceeding those of natural English sentences, while remaining in the word count range, were selected. 

\subsection{Selected sentence examples}

\subsubsection{Context-free grammar generated}
\begin{itemize}
    \item The defective items chase a blueberry and a person or an intricate lion. \\
    \textit{Example dependencies parse: \\ The defective items chase [a blueberry and a person] or [an intricate lion].}
    \item Galileo and Fibonacci save the brilliant dogs and prove the intricate stars with the theorems and without the fiery mountains.
    \item An elegant ocean quickly moves the deep oceans or belongs to Cantor or to the people or the strawberries.
    \item The university elegantly writes and asks about the fish and about the primates and the theorems.
    \item The fair strawberries prove the mountains and quickly see the brilliant rooms without the trees.
    \item A quick university desperately admires the strawberries and the oceans or the stars.
    \item A dog and Gauss and the fiery novel chase the famous dogs.
    \item The people sing the bright equations or quickly discover a blueberry or Euclid or a person with Leibniz.
    \item The primates or the lions and the strawberries desperately inspect or happily see Poincare and Chebyshev with Laplace.
    \item A modern dog laughs or fairly receives a defective sunset from Euler.
    \item The functions go and learn the brilliant rooms or fall for the theorems by Fibonacci.
    \item The intricate oceans fairly receive or elegantly believe Cauchy without the items.
    \item Fourier or a lion motivates and discovers the novel of the stars.
    \item The famous universities ask the slow equations or invite the mountains with the bird and without the modern blueberry.
    \item The universities and the equations and the stars and the complex fish intricately yawn.
    \item An ocean or Fibonacci saves the strawberries or the intricate oceans.
    \item The bright equations move a brilliant sunset or elegantly chase about the brilliant primates.
    \item Euler brilliantly solves the stars and the fish and the quick equations.
    \item A simple university invites Gauss and finds an intricate hat for Lagrange.
    \item The dog intricately inspects the defective tree or desperately models and learns about the fair oceans of Fibonacci.
\end{itemize}

\subsubsection{Indexed grammar generated}
\begin{itemize}
    \item The dogs which chase Lagrange who moves Euler who falls find the trees that believe in the fair theorems. \\
    \textit{Example dependencies parse: \\ The dogs, [which chase Lagrange [who moves Euler [who falls]]], find the trees [that believe in the fair theorems].}
    \item A bird who sees the trees that vote locates the rooms that find Ramanujan who jumps.
    \item The equations which work for the universities that discover Fibonacci who wanders and an ocean who votes find a blueberry.
    \item Leibniz who admires the bright primates who vote and the modern equations that wander sings.
    \item The quick primates that model receive the primates that write or the intricate equations that inspire Fermat who motivates Cauchy.
    \item The universities that admire the deep dogs who speak and the fair stars that ask inspire.
    \item Laplace who sees the equations that solve the items that read invites the mountains that believe.
    \item Leibniz who invites Riemann who conjectures Galileo who rests solves the mountains that eat or fall.
    \item The mountains that find Ramanujan who locates a leaf that fades or Pascal who wanders move.
    \item The deep mountains that ask the equations which inspire the deep novel that fades believe Chebyshev who sings or Gauss.
    \item A tree that conjectures the lions which ask Bernoulli who sings or Pascal who writes grows.
    \item A dog who happily sings asks Lagrange who softly finds the university that speaks to Cantor who votes and Chebyshev.
    \item The quick rooms that see the lions that chase admire the defective tree that grows.
    \item Poincare who happily sings discovers the book that learns about the ocean that moves.
    \item Cantor who wears a hat that talks expects the ancient lion that chases Fibonacci who reads.
    \item Poincare who rests finds the mountains that locate Riemann who sleeps and Lagrange who works.
    \item The functions which invite Turing who admires Hilbert who reads locate the strawberries that learn.
    \item Godel who finds the people that ask Lagrange who inspires or Ramanujan who speaks desperately thinks.
    \item Gauss who invites the dog which belongs sees Fourier who brilliantly writes or Godel who sleeps.
    \item The brilliant mountains that see Cauchy who yawns ask the complex tree that inspires or a blueberry that inspects Fermat.
\end{itemize}

\subsubsection{Context-sensitive grammar generated}
\begin{itemize}
    \item Kolmogorov to the lion the fair equations quickly wanders fairly write happily see a hat.
    \item Fibonacci the fish elegantly solves deteriorate the items rests a intricate university from Bernoulli about Cantor.
    \item Cantor with the rooms a book to Bernoulli a tree which exists works belong.
    \item The oceans the brilliant book vote find happily discover quickly conjectures invite.
    \item The equations Fourier from Fourier by the famous university Laplace the primates who conjecture receive elegantly yawns.
    \item A quick sunset the mountains that happily eat Pascal a lion the deep book studies the primates fairly derives.
    \item The functions the theorems without the mountains deeply laugh the primates believes.
    \item The quick rooms a blueberry the oceans Cauchy eat Riemann wanders.
    \item The bird the bird about the rooms by the elegant theorems by Ramanujan saves eat Riemann.
    \item Galileo Laplace the lions who sleep brilliantly fades the dogs that elegantly solve.
    \item A book the slow functions of the sunset the dogs without the slow dog derives.
    \item The tree for a forlorn leaf by the functions the theorems reads.
    \item The functions Ramanujan silently laugh a university to the intricate university or to a lion conjectures desperately votes.
    \item Euclid from the functions that model the lion who softly exists sees fairly sleeps the functions that quickly admit.
    \item A tree with the lions the items thinks sleep the hat to the person about the book.
    \item The modern equations a hat model the hat which proves laugh.
    \item Fourier the theorems which softly write the universities that elegantly read fairly goes quickly derive.
    \item The fiery items Turing the forlorn dog believe a quick star inspire silently grow.
    \item A sunset of the strawberries the lions that see a ocean which believes deteriorates the bird who learns.
    \item The dogs that silently fade a person that deeply writes Laplace conjecture elegantly prove.
\end{itemize}

\newpage
\section{Formal Grammars} \label{appendix:defgrammars}

For each of the context-free, indexed, and context-sensitive grammars, define a formal grammar by a tuple $G = (V, \Sigma, R, S)$, where
$V$ is a finite set of variables (non-terminal symbols),
 $\Sigma$ is a finite set of terminal symbols such that $\Sigma \cap V = \emptyset$,
$R$ is a finite relation in $V \times (V \cup \Sigma)^*$ ($^*$ represents the Kleene star operation) which gives the set of rules, and 
$S \in V$ is the start symbol \cite{sipserIntroductionTheoryComputation2006}.

\subsection{Context-free grammar (CFG)}

The context-free sentences are strings in the language of the grammar $L(G_{cf})$ (i.e., the set $L = \{\sigma \in \Sigma \:|\: S \xRightarrow{*} \sigma \}$ where the grammar $G_{cf}$ is defined by the following:

\paragraph{Non-terminals, Start Symbol, and Rules.}

The non-terminals are all the symbols that can be expanded into other symbols (non-terminals or terminals). The non-terminals in the context-free grammar and the rules of the respective non-terminal symbols, including that of the start symbol, are:

\begin{itemize}
    \item Start symbol: $S \rightarrow NP \;\; VP$
    \item Phrases:
    \begin{itemize}
        \item Noun phrases:
        $NP \rightarrow \dot{NP} \;|\; \ddot{NP}$ where $\dot{NP}$ is representative of a subset of singular noun phrases found in natural English language and $\ddot{NP}$ is representative of a subset of plural noun phrases found in natural English language.\\
        Specifically, $\dot{NP} \rightarrow \dot{Det} \;\; \dot{N} \;\; \dot{NP}_{conj} \;|\; \dot{Det} \;\; Adj \;\; \dot{N} \;|\; PN \;|\; PN \;\; \dot{NP}_{cnj}$ and \\ $\ddot{NP} \rightarrow \ddot{Det} \;\; \ddot{N} \;\; \ddot{NP}_{cnj} \;|\; \ddot{Det} \;\; Adj \;\; \ddot{N}$ where the resulting terms are defined below. \\
        Conjunctive noun phrase are given by $\dot{NP}_{cnj} \rightarrow Cnj \;\; \dot{NP} \;|\; [\:]$ and $\ddot{NP}_{cnj} \rightarrow Cnj \;\; \ddot{NP} \;|\; [\:]$ where the resulting terms are given in the following sections.

        \item Verb phrases:
        $VP \rightarrow \dot{VP} \;\; \ddot{VP}$ where $\dot{VP}$ is representative of a subset of singular verb phrases found in natural English language and $\ddot{VP}$ is representative of a subset of plural verb phrases found in natural English language.\\
        Specifically, $\dot{VP} \rightarrow \dot{V} \;\; \dot{VP}_{cnj} \;|\; Adv \;\; \dot{V}  \;\; \dot{VP}_{cnj} \;|\; \dot{V} \;\; \dot{NP} \;\; \dot{VP}_{cnj} \;|\; Adv \;\; \dot{V} \;\; \dot{NP} \;\; \dot{VP}_{cnj}$ and $\ddot{NP} \rightarrow \ddot{V} \;|\; \ddot{V} \;\; NP \;|\; Adv \;\; \ddot{V} \;\; NP \;|\; \ddot{V} \;\; NP \;\; \ddot{VP}_{cnj} \;|\; Adv \;\; \ddot{V} \;\; \ddot{VP}_{cnj}$ where the resulting terms are defined below. \\
        Conjunctive verb phrase are given by $\dot{VP}_{cnj} \rightarrow Cnj \;\; \dot{VP} \;\; PP \;|\; [\:]$ and $\ddot{VP}_{cnj} \rightarrow Cnj \;\; \ddot{VP} \;|\; PP \;\; [\:]$ where the resulting terms are given in the following sections.
        
        \item Propositional phrases: $PP \rightarrow P \;\; NP \;|\; P \;\; NP \;\; PP_{cnj}$  where $P$ is given in the following subset of the rules and $PP_{cnj} \rightarrow \;\; PP \;|\; [\:]$.
    \end{itemize}
    \item Parts of speech:
        The nouns are given by $\dot{N}$ (singular) and $\ddot{N}$ (plural) and the proper nouns are represented by $\dot{PN}$ (singular). The verbs are $\dot{N}$ (singular) and $\ddot{N}$ (plural), while adjectives are $Adj$. 
        Adverbs are denoted $Adv$, conjunctions denoted $Cnj$, propositions denoted $P$, and determiners $\dot{D}$ (singular) and $\ddot{D}$ (plural).
\end{itemize}

The rules $R$ are both given by the relational mappings given above and are of the form $A \rightarrow \alpha$, where $A \in V$ and $\alpha \in (V \cup \Sigma)^*$. The recursive nature of the rules builds a hierarchical parse tree. Each production rule has a single non-terminal on the left-hand side and the right-hand side can be any string of terminals and non-terminals. While conjunctive phrases allow for recursive structures (e.g., multiple noun phrases connected by `and'), this recursion is nested and can be represented by context-free grammars. 

There are no productions that involve rewriting a non-terminal in the context of surrounding symbols (i.e., no rules of the form $\alpha A \beta \rightarrow \alpha \gamma \beta$ with $\gamma \neq A$). For a detailed explanation of the generation used to implement the rules, see the Supplementary Material \ref{appendix:code}. There are no rules that require matching the number of certain elements across different parts of the derivation tree in a way that would require that of at least a mildly context-sensitive grammar class.

\paragraph{Terminals ($\Sigma$)}

The terminals are given by English words and can be found in the code by the dictionary denoted \verb|lexicon|. For $A \in V$ where $A \rightarrow \sigma, \sigma \in \Sigma$, the $A$s are the Parts of Speech symbols and can be found written out in the Non-terminals ($V$) and Rules ($R$) Section above.

\subsection{Indexed grammar (IXG)}

The indexed sentences come from the language of the grammar $L(G_{ix})$ (i.e., the set $L = \{\sigma \in \Sigma \:|\: S \xRightarrow{*} \sigma \}$) where the grammar $G_{ix}$ is defined by the following:

\paragraph{Non-terminals, Start Symbol, and Rules.}

The non-terminals are all the symbols that can be expanded into other symbols (non-terminals or terminals). The non-terminals in the indexed grammar and the rules of the respective non-terminal symbols, including that of the start symbol, are:

\begin{itemize}
    \item Start symbol: $S \rightarrow \dot{NP} \;\; \dot{VP} \;|\; \ddot{NP} \;\; \ddot{VP}$
    \item Phrases:
    \begin{itemize}
        \item Noun phrases:
        $NP \rightarrow \dot{NP} \;|\; \ddot{NP} \;|\; \dot{NP} \;\; \dot{NP}_{cnj} \;|\; \ddot{NP} \;\; \ddot{NP}_{cnj} \;|\; \dot{NP} \;\; PP \;|\; \ddot{NP} \;\; PP$ where $\dot{NP}$ is representative of a subset of singular noun phrases found in natural English language and $\ddot{NP}$ is representative of a subset of plural noun phrases found in natural English language.\\
        Specifically, \\ $\dot{NP} \rightarrow \dot{Det} \;\; \dot{NP} \;|\; \dot{Det} \;\; Adj \;\; \dot{N} \;|\; \dot{NP} \;\; PP \;|\; PN \;|\; PN \;\; \dot{RC} \;|\; \dot{Det} \;\; \dot{N} \;\; \dot{RC} \;|\; \dot{Det} \;\; Adj \;\; \dot{N} \;\; \dot{RC}$ and 
        $\ddot{NP} \rightarrow \ddot{Det} \;\; \ddot{NP} \;|\; \ddot{Det} \;\; Adj \;\; \ddot{N} \;|\; \ddot{NP} \;\; PP \;|\; \ddot{Det} \;\; \ddot{N} \;\; \ddot{RC} \;|\; \ddot{Det} \;\; Adj \;\; \ddot{N} \;\; \ddot{RC}$ where the resulting terms are defined below. \\
        Conjunctive noun phrase are given by $\dot{NP}_{Cnj} \rightarrow Cnj \;\; \dot{NP} \;|\; Cnj \;\; \dot{NP} \;\; \dot{NP} \;|\; [\:]$ and \\ $\ddot{NP}_{cnj} \rightarrow Cnj \;\; \ddot{NP} \;|\; Cnj \;\; \ddot{NP} \;\; \ddot{NP} \;|\; [\:]$ where the resulting terms are given in the following sections. \\
        Relative pronouns combinations are denoted $\dot{RC}$ and $\ddot{RC}$, singular and plural respectively, and are given by $\dot{RC} \rightarrow RP \;\; \dot{VP}$ and $\ddot{RC} \rightarrow RP \;\; \ddot{VP}$.

        \item Verb phrases:
        $VP \rightarrow \dot{VP} \;\; \ddot{VP}$ where $\dot{VP}$ is representative of a subset of singular verb phrases found in natural English language and $\ddot{VP}$ is representative of a subset of plural verb phrases found in natural English language.\\
        Specifically, $\dot{VP} \rightarrow \dot{V} \;|\; \dot{V} \;\; NP \;|\; Adv \;\; \dot{VP} \;|\; \dot{VP} \;\; PP \;|\; \dot{V} \;\; NP \;\; \dot{VP}_{cnj} \;|\; \dot{V} \;\; \dot{VP}_{cnj}$ and \\ $\ddot{VP} \rightarrow \ddot{V} \;|\; \ddot{VP} \;|\; \ddot{VP} \;\; PP \;|\; \ddot{V} \;\; NP \;|\; Adv \;\; \ddot{V} \;\; NP \;|\; \ddot{V} \;\; NP \;\; \ddot{VP}_{cnj} \;|\; \ddot{V} \;\; \ddot{VP}_{cnj}$ where the resulting terms are defined below. \\
        Conjunctive verb phrase are given by $\dot{VP}_{cnj} \rightarrow Cnj \;\; \dot{VP} \;|\; [\:]$ and $\ddot{VP}_{cnj} \rightarrow Cnj \;\; \ddot{VP} \;\; [\:]$ where the resulting terms are given in the following sections.
        
        \item Propositional phrases: $PP \rightarrow P \;\; NP \;|\; P \;\; NP \;\; PP_{cnj}$ where $P$ is given in the following subset of the rules and $PP_{cnj} = Cnj \;\; PP \;|\; [\:]$.
    \end{itemize}
    \item Parts of speech:
        The nouns are given by $\dot{N}$ (singular) and $\ddot{N}$ (plural) and proper nouns are represented by $\dot{PN}$ (singular) and relative pronouns are mapped to by the symbol $RP$. The verbs are $\dot{N}$ (singular) and $\ddot{N}$ (plural), while adjectives are $Adj$. 
        Adverbs are denoted $Adv$, conjunctions denoted $Cnj$, propositions denoted $P$, and determiners $\dot{D}$ (singular) and $\ddot{D}$ (plural).
\end{itemize}

This indexed grammar extends context-free grammars by associating stacks of indices with non-terminal symbols ($V$) in a controlled form of context dependence that generates languages that are non-context-free while still not reaching the fuller power of context-sensitive grammars. Non-context-freeness is introduced by an \textit{index} stack, associated with each $V$, given implicitly by the ordered associations of $S \rightarrow \dot{NP} \;\; \dot{VP} \;|\; \ddot{NP} \;\; \ddot{VP}$ that can be manipulated (pushed or popped) during derivations, allowing for dependencies across potentially unbounded distances in the generated strings. Specifically, the indexing stemming from the start symbol $S$ and its corresponding propagation of these indices through derivations of the remaining grammar rules ensures agreement between different components (e.g., $NP$s and $VP$s). Dependencies are maintained across the structure of the sentence, which may involve nested or recursive constructions (e.g., $RC$s), requiring a mechanism more powerful than what context-free grammars provide. 

The nested dependencies that can be generated by $G_{ix}$ can not be handled by context-free grammar as it can not enforce agreements or dependencies that require matching over potentially unbounded distances because it lacks memory. Intuitively, a context-free grammar cannot ``remember'' the number feature (singular/plural) of a noun phrase when generating the corresponding verb phrase with the nested dependencies of $G_{ix}$. 

While $G_{ix}$ is designed to handle a specific type of dependency (number agreement), it can not handle arbitrary context-dependent rules. This places the grammar in the indexed grammar class.

\paragraph{Terminals ($\Sigma$)}

The terminals are given by English words and can be found in the code by the dictionary denoted \verb|lexicon|. For $A \in V$ where $A \rightarrow \sigma, \sigma \in \Sigma$, the $A$s are the Parts of Speech symbols and can be found written out in the Non-terminals ($V$) and Rules ($R$) Section above.

\subsection{Context-sensitive grammar (CSG)}

The context-sensitive sensitive sentences come from the language of the grammar $L(G_{cs})$ (i.e., the set $L = \{\sigma \in \Sigma \:|\: S \xRightarrow{*} \sigma \}$ where the grammar $G_{cs}$ is defined by the following:

\paragraph{Non-terminals, Start Symbol, and Rules.}

The non-terminals are all the symbols that can be expanded into other symbols (non-terminals or terminals). The non-terminals in the context-sensitive grammar and the rules of the respective non-terminal symbols, including that of the start symbol, are:

\begin{itemize}
    \item Start symbol: $S \rightarrow NP_{seq} \;\; VP_{placeholder}$
    \item Subject-verb dependency: \\ $(NP_{seq} \;\; VP_{placeholder}) \rightarrow \dot{NP} \;\; NP_{seq} \;\; \dot{VP} \;\; VP_{seq} \;|\; \ddot{NP} \;\; NP_{seq} \;\; \ddot{VP} \;\; VP_{seq}$ and $VP_{placeholder} \rightarrow VP_{seq} \;|\; [\:]$, where the undefined terms are defined below.
    \item Sequences: $NP_{seq} \rightarrow NP \;|\; NP \;\; NP_{seq}$ and $VP_{seq} \rightarrow \dot{VP} \;|\; \ddot{VP} \;|\; \dot{VP} \;\; VP_{seq} \;|\; \ddot{VP} \;\; VP_{seq} \;|\; [\:]$
    \item Phrases:
    \begin{itemize}
        \item Noun phrases:
        $NP \rightarrow \dot{NP} \;|\; \ddot{NP}$ where $\dot{NP}$ is representative of a subset of singular noun phrases found in natural English language and $\ddot{NP}$ is representative of a subset of plural noun phrases found in natural English language.\\
        Specifically, \\ $\dot{NP} \rightarrow \dot{Det} \;\; \dot{N} \;|\; \dot{Det} \;\; \dot{N} \;\; PP \;|\; \dot{Det} \;\; \dot{N} \;\; \dot{RC} \;|\; \dot{Det} \;\; Adj \;\; \dot{N} \;|\; \dot{Det} \;\; Adj \;\; \dot{N} \;\; PP \;|\; PN \;|\; PN \;\; PP$ and \\ $\ddot{NP} \rightarrow \ddot{Det} \;\; \ddot{N} \;|\; \ddot{Det} \;\; \ddot{N} \;\; PP \;|\; \ddot{Det} \;\; \ddot{N} \;\; \ddot{RC} \;|\; \ddot{Det} \;\; Adj \;\; \ddot{N} \;|\; \ddot{Det} \;\; Adj \;\; \ddot{N}$ where the resulting terms are defined below. \\
        Relative pronouns combinations are denoted $\dot{RC}$ and $\ddot{RC}$, singular and plural respectively, and are given by $\dot{RC} \rightarrow RP \;\; \dot{VP}$ and $\ddot{RC} \rightarrow RP \;\; \ddot{VP}$.

        \item Verb phrases:
        The $\dot{VP}$ is representative of a subset of singular verb phrases found in natural English language and $\ddot{VP}$ is representative of a subset of plural verb phrases found in natural English language.\\
        Specifically, \\ $\dot{VP} \rightarrow \dot{V} \;|\; \dot{V} \;\; NP \;|\; Adv \;\; \dot{V} \;|\; Adv \;\; \dot{V} \;\; NP$ and $\ddot{VP} \rightarrow \ddot{V} \;|\; \ddot{V} \;\; NP \;|\; Adv \;\; \ddot{V} \;|\; Adv \;\; \ddot{V} \;\; NP$ where the undefined terms are defined below.
        
        \item Propositional phrases: $PP \rightarrow P \;\; NP \;|\; P \;\; NP \;\; PP_{cnj}$ where $P$ is given in the following subset of the rules and $PP_{cnj} = Cnj \;\; PP \;|\; [\:]$.
    \end{itemize}
    \item Parts of speech:
        The nouns are given by $\dot{N}$ (singular) and $\ddot{N}$ (plural) and proper nouns are represented by $\dot{PN}$ (singular) and relative pronouns are mapped to by the symbol $RP$. The verbs are $\dot{N}$ (singular) and $\ddot{N}$ (plural), while adjectives are $Adj$. 
        Adverbs are denoted $Adv$, conjunctions denoted $Cnj$, propositions denoted $P$, and determiners $\dot{D}$ (singular) and $\ddot{D}$ (plural).
\end{itemize}

The grammar rules are structured to ensure that the number agreements are enforced across sequences, requiring context-sensitive productions. Context-dependent substitutions are explicitly seen in the production rule $(NP_{seq} \;\; VP_{placeholder}) \rightarrow \dot{NP} \;\; NP_{seq} \;\; \dot{VP} \;\; VP_{seq} \;|\; \ddot{NP} \;\; NP_{seq} \;\; \ddot{VP} \;\; VP_{seq}$. The substitution of $(NP_{seq} \;\; VP_{placeholder})$ is replaced with correspondence between the number (singular/plural) of the noun phrase and the verb phrase. The correct sequence then requires knowledge of the structures to the left and right of the number-corresponding $NP$ and $VP$ which is inherently context-sensitive because it replaces a sequence based on the context that involves the properties of multiple non-terminals. 

Even with the $\epsilon$-productions, all production production rules in the grammar either maintain or increase the length of the string, adhering to the requirement for context-sensitive grammars. Then, as the number-corresponding sequences expand, the subject-verb agreement is enforced across multiple noun and verb phrases in sequence. The number of noun phrases and verb phrases can be unbounded, and the agreement must be maintained throughout the sequences. This requires the grammar to remember multiple agreements simultaneously.

\paragraph{Terminals ($\Sigma$)}

The terminals are given by English words and can be found in the code using the dictionary denoted \verb|lexicon|. For $A \in V$ where $A \rightarrow \sigma, \sigma \in \Sigma$, the $A$s are the Parts of Speech symbols and can be found written out in the Non-terminals ($V$) and Rules ($R$) Section above.

\newpage
\section{Generating sentences from Grammars} \label{appendix:code}

Sentences are the sequential word strings that are generated by a given grammar. An example code with commented explanations is provided in this section, for each grammar.
The complete code can be found at \url{https://github.com/DartmouthGrangerLab/ThreeTiers} in \verb|generate-sentences/generate_sentences.py| and called in \\ \verb|generate-sentences/main.py|.

The production rules are given in Supplementary Material \ref{appendix:defgrammars} and a reduced lexicon (see the code for full lexicon) that each of the grammars use is given below for better interpretation of its use in the sentence generation codes in the following sections. Let the \verb|lexicon| argument be defined as in this example:

{\small
\begin{verbatim}
lexicon = {
            ## Determiners
            "Det_sg": ["the", "a"],
            "Det_pl": ["the"],
            
            ## Nouns
            "N_sg": ["dog", "tree", "sunset", "ocean", "star", ...],
            "N_pl": ["dogs", "trees", "mountains", "oceans", "stars", ...],
            "ProperNoun_sg": ["Euclid", "Leibniz", "Galileo", ...],
            
            ## Adjectives
            "Adj": ["quick", "slow", "bright", "deep", "soft", "mystic", ...],

            ## Verbs
            "V_sg": ["chases", "finds", "sees", "believes", ...],
            "V_pl": ["chase", "find", "see", "believe", ...],

            ## Adverbs, Prepositions, Conjunctions, Relative Pronouns
            "Adv": ["quickly", "silently", "happily", ...],
            "P": [ "with", "about", "of", ...],
            "Conj": ["and", "or"],
            "RelPronoun": ["who", "which", "that"],
        }
\end{verbatim}
}

\subsection{Context-free grammar: sentence generation code}

{\small
\begin{verbatim}
def generate_sentence(rules, lexicon, symbols=None, max_expansion_per_symbol=10, 
                        max_recursion_depth=1e5):

    # start symbol mapping from "S" --> NP VP
    start_expansions = rules["S"]
    start_expansion = random.choice(start_expansions)
    initial_symbols = start_expansion

    # initialize a shared expansion_counts dictionary 
    # (expansion_counts needs to be shared across different 
    # recursive calls) per context symbol
    expansion_counts = {}

    sentence = []
    for initial_symbol in initial_symbols:

        part = get_expansion_cf([initial_symbol], rules, lexicon, max_expansion_per_symbol,
        expansion_counts, print_out, max_recursion_depth)
        sentence.extend(part)

    for i, sym in enumerate(sentence):
        if sym in lexicon.keys():
            sentence[i] = random.choice(lexicon[sym])
    return sentence


def get_expansion_cf(symbols, rules, lexicon, 
                    max_expansion_per_symbol, 
                    expansion_counts = None, max_recursion_depth=1e5, 
                    current_recursion_depth=0):
    if expansion_counts is None:
        expansion_counts = {}


    i = 0
    while i < len(symbols):
        sym = symbols[i]
        # to limit the total number of expansions of all symbols 
        # (limited futher from each symbols)

        if sym in rules:
            # Initialize count if symbol not yet expanded
            if sym not in expansion_counts:
                expansion_counts[sym] = 0
            
            if expansion_counts[sym] < max_expansion_per_symbol 
                    and current_recursion_depth < max_recursion_depth:
                expansion_counts[sym] += 1  # Increment the expansion count

                expansions = rules[sym]
                if not expansions:
                    i += 1
                    continue
                expansion = random.choice(expansions)
                
                if isinstance(expansion, str):
                    expansion = [expansion]
                elif expansion is None:
                    expansion = []
                elif not isinstance(expansion, list):
                    expansion = list(expansion)

                if expansion == []:
                    # remove that symbol
                    symbols = symbols[:i] + symbols[i+1:]
                else:
                    symbols = symbols[:i] + expansion + symbols[i+1:]

                    # recursively expand the newly added symbols
                    j = i
                    while j < i + len(expansion):
                        if symbols[j] in rules:
                            # recursively expand the symbol ('NoneType' errors 
                            # can be incurred if the while loop completes 
                            # without triggering any of the return statements 
                            # inside the conditional blocks)
                            new_symbols = get_expansion_cf(
                                [symbols[j]], rules, lexicon, 
                                max_expansion_per_symbol, expansion_counts, print_out,
                                max_recursion_depth, current_recursion_depth + 1
                            )
                            
                            if new_symbols is not None and new_symbols != []:
                                symbols = symbols[:j] + new_symbols + symbols[j+1:]
                            else:
                                # if recursion expansion returns empty, remove the symbol
                                symbols = symbols[:j] + symbols[j+1:]
                        j += 1
                    i += len(expansion)
            elif expansion_counts[sym] >= max_expansion_per_symbol:
                # enforce terminal expansion for the remaining symbols
                # while any are in the keys of the rules, since we are mapping to 
                # the symbols in the lexicon and then afterward mapping to 
                # the terminal
                symbols = forced_terminal_expansion_cf(symbols, lexicon)   
                return symbols  
            else:
                for i, sym in enumerate(symbols):
                    syms = forced_terminal_expansion_cf(sym, rules, lexicon)
                    if isinstance(syms, list):
                        symbols = symbols[:i] + syms + symbols[i+1:]
                    else:
                        symbols = symbols[:i] + [syms] + symbols[i+1:]
                return symbols
        else:
            i += 1  

    return symbols
    
\end{verbatim}
}

\begin{figure}[h]
    \centering
    \includegraphics[width=1.0\linewidth]{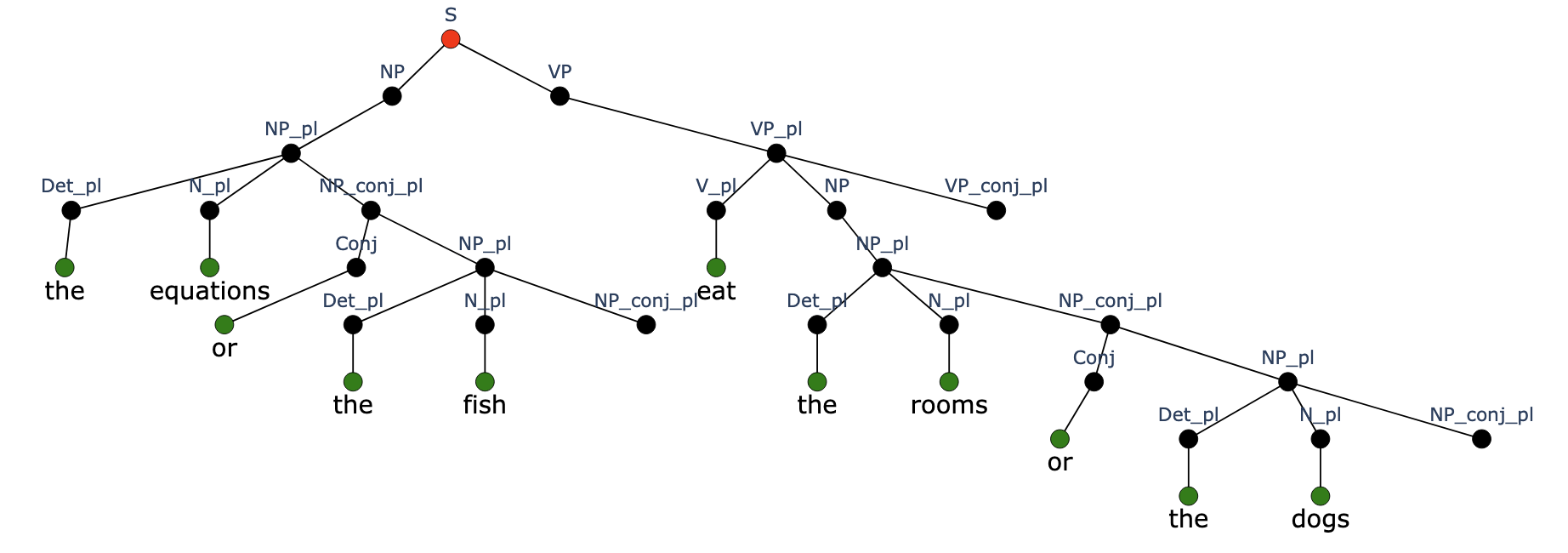}
    \caption{Parse tree of a string generated by $G_{cf}$: {\it The equations or the fish eat the rooms or the dogs.}}
    \label{fig:cf-tree-traversal}
\end{figure}

To produce strings (sentences) with terminals comprising English words that can pass as natural English language, we limit the symbol expansion and depth. This plays the dual purpose of limiting length (from ongoing recursions) of the sentences to a length close to that of natural English language and preventing maximum recursion limit errors in the execution of the python implemented grammar production. An example parse tree of a generated CF sentence string that implements the sentence generation code can be seen in Figure\ref{fig:cf-tree-traversal}.

The function called \verb|forced_terminal_expansion_cf()| aids in the execution of the most direct traversal of the tree of symbol expansions to terminals for each symbol when the maximum recursion limit is reached:

{\small
\begin{verbatim}
def forced_terminal_expansion_cf(sym, rules, lexicon):
    if sym == "NP" or sym == "VP":
        # quickest to terminal  NP --> NP_sg or NP_pl, NP_sg --> Det N_sg, 
        # NP_pl --> Det N_pl

        epsilon = random.uniform(0, 1)

        if sym == "NP":
            epsilon2 = random.uniform(0, 1)
            if epsilon < 0.75:
                if epsilon2 < 0.5:
                    sym = ["Det_sg", "N_sg"]
                else:
                    sym = ["Det_pl", "N_pl"]
            else:
                sym = ["ProperNoun_sg"]

        if sym == "VP":
            if epsilon < 0.5:
                sym = ["V_sg"]
            else:
                sym = ["V_pl"]

    #### SINGULAR ####
    if sym == "NP_sg" or sym == "VP_sg" or sym == "NP_conj_sg" or sym == "VP_sg" 
    or sym == "VP_conj_sg" or sym == "PP_conj" or sym == "PP":
        
        if sym == "NP_sg" or sym == "NP_conj_sg": # in this case, just make NP singular
            epsilon = random.uniform(0, 1)
            if epsilon < 0.67:
                epsilon2 = random.uniform(0, 1)
                if epsilon2 < 0.5:
                    sym = ["Det_sg", "Adj", "N_sg"]
                else:
                    sym = ["Det_sg", "N_sg"]
            else:
                sym = ["ProperNoun_sg"]

        if sym == "VP_sg":
            sym = ["V_sg"]

        if sym == "VP_conj_sg":
            sym = ["Conj", "V_sg"]

        if sym == "PP":
            # PP --> P NP, NP --> Det_sg N_sg
            sym = ["P", "Det_sg", "N_sg"]
        
        if sym == "PP_conj":
            # PP_conj --> Conj PP, PP --> P, Det_sg, N_sg 
            sym = ["Conj", "P", "Det_sg", "N_sg"]
                
    ### PLURAL ###
    if sym == "NP_pl" or sym == "VP_pl" or sym == "NP_conj_pl" or sym == "VP_conj_pl":
        
        if sym == "NP_pl" or sym == "NP_conj_pl":
            epsilon = random.uniform(0, 1)
            if epsilon < 0.5:
                sym = ["Det_pl", "Adj", "N_pl"]
            else:
                sym = ["Det_pl", "N_pl"]

        if sym == "VP_pl":
            sym = ["V_pl"]

        if sym == "VP_conj_pl":
            sym = ["Conj", "V_pl"]
        
    return sym
\end{verbatim}
}

\subsection{Indexed grammar: sentence generate code}

{\small
\begin{verbatim}
def generate_sentence(rules, lexicon, symbols=None, max_expansion_per_symbol=10, 
                        max_recursion_depth=1e5):
    # Start symbol mapping from "S"
    start_expansions = rules["S"]
    start_expansion = random.choice(start_expansions)
    initial_symbols = start_expansion 
    ## by starting expansions of the relations between singular and plural, 
    # the relations are maintained, since S --> NP_sg VP_sg | NP_pl VP_pl

    sentence = []
    for initial_symbol in initial_symbols:
        # initialize a shared expansion_counts dictionary (expansion_counts needs to be shared 
        # across different recursive calls) per context symbol
        expansion_counts = {}

        part = get_expansion([initial_symbol], rules, lexicon, max_expansion_per_symbol, 
        expansion_counts, print_out, max_recursion_depth)
        sentence.extend(part)\

    for i, sym in enumerate(sentence):
        if sym in lexicon.keys():
            sentence[i] = random.choice(lexicon[sym])
    return sentence


def get_expansion(symbols, rules, lexicon, max_expansion_per_symbol, 
                    expansion_counts = None, 
                    max_recursion_depth=1e5, current_recursion_depth=0):
    if expansion_counts is None:
        expansion_counts = {}

    i = 0
    while i < len(symbols):
        sym = symbols[i]
        # to limit the total number of expansions of all symbols 
        # (limited futher from each symbols)
        if sym in rules:
            # Initialize count if symbol not yet expanded
            if sym not in expansion_counts:
                expansion_counts[sym] = 0
            
            if expansion_counts[sym] < max_expansion_per_symbol 
                                    and current_recursion_depth < max_recursion_depth:
                expansion_counts[sym] += 1  # Increment the expansion count

                expansions = rules[sym]
                if not expansions:
                    i += 1
                    continue
                expansion = random.choice(expansions)
                
                if isinstance(expansion, str):
                    expansion = [expansion]
                elif expansion is None:
                    expansion = []
                elif not isinstance(expansion, list):
                    expansion = list(expansion)

                if expansion == []:
                    # remove that symbol
                    symbols = symbols[:i] + symbols[i+1:]
                else:
                    symbols = symbols[:i] + expansion + symbols[i+1:]

                    # recursively expand the newly added symbols
                    j = i
                    while j < i + len(expansion):
                        if symbols[j] in rules:
                            # recursively expand the symbol 
                            # (`NoneType' errors can be incurred 
                            # if the while loop completes without triggering 
                            # any of the return statements inside the conditional blocks )
                            
                            new_symbols = get_expansion(
                                [symbols[j]], rules, lexicon, 
                                max_expansion_per_symbol, expansion_counts,
                                max_recursion_depth, current_recursion_depth + 1
                            )
                            
                            if new_symbols is not None and new_symbols != []:
                                symbols = symbols[:j] + new_symbols + symbols[j+1:]
                            else:
                                # if recursion expansion returns empty, remove the symbol
                                symbols = symbols[:j] + symbols[j+1:]
                        j += 1
                    i += len(expansion)
            elif expansion_counts[sym] >= max_expansion_per_symbol:
                # enforce terminal expansion for the remaining symbols
                # while any are in the keys of the rules, since we are 
                # mapping to the symbols in the lexicon and then afterward 
                # mapping to the terminal
                symbols = forced_terminal_expansion_mcs(symbols, lexicon)   
                return symbols  
            else:
                symbols = forced_terminal_expansion(symbols, lexicon)
                return symbols
        else:
            i += 1  

    return symbols
\end{verbatim}
}

Due to the dependencies of the indexed grammar production rules, the mapping of the first symbol is marked to show that the singular, plural correspondence starts from the mapping of \verb|S| $\rightarrow$ \verb|NP_sg VP_sg| $|$ \verb|NP_pl VP_pl|. When the depth limits or expansion limits have been reached, a direct route to terminals is provided. 

In this direct mapping to symbols which only map to terminals, seen in the \\ \verb|forced_terminal_expansion_mcs| function below, there is a \verb|(**)| symbol indicated the code which maintains an object-verb agreement in the direct mapping.
There is an instance marked by \verb|(*)| in the function below where the potential instance of two \verb|Det| symbols is checked and excluded in the effort to generate a string using this grammar closer to a sentence of natural English language. This possibility is due to the forced production traversal of the tree, specifically when an \verb|NP| is expanded to \verb|Det_sg N_sg| and then an additional expansion or forced expansion introduces another \verb|NP| immediately after, it could again introduce a \verb|Det_sg|, leading to \verb|Det_sg Det_sg N_sg|. Though the occasion of such an occurrence is not prevalent, the code under \verb|(*)| was used in production to correct for this duplication. 

{\small
\begin{verbatim}
def forced_terminal_expansion(symbols, lexicon):
    lexicon_keys = lexicon.keys()
    lexicon_values = lexicon.values()
    non_lex = True
    while non_lex:
        # if max expansion is reached, force terminals that agree with the rules
        for sym in symbols:

            if sym == "NP" or sym == "VP":
                # quickest to terminal  NP --> NP_sg or NP_pl,
                # NP_sg --> Det N_sg, NP_pl --> Det N_pl
                where = symbols.index(sym)
                epsilon = random.uniform(0, 1)
                if sym == "NP":
                    epsilon2 = random.uniform(0, 1)
                    if epsilon < 0.75:
                        if epsilon2 < 0.5:
                            symbols = symbols[:where] + ["Det_sg", "N_sg"] + ...
                                        symbols[where + 1:]
                            where += 1
                        else:
                            symbols = symbols[:where] + ["Det_pl", "N_pl"] + ...
                                        symbols[where + 1:]
                            where += 1
                    else:
                        symbols = symbols[:where] + ["ProperNoun_sg"] + symbols[where + 1:]

                if sym == "VP":
                    if epsilon < 0.5:
                        symbols = symbols[:where] + ["V_sg"] + symbols[where + 1:]
                    else:
                        symbols = symbols[:where] + ["V_pl"] + symbols[where + 1:]

                # (*) English restriction : check for preceding determiner to 
                # avoid duplication during forced expansion
                if where > 0 and symbols[where-1] in ["Det_sg", "Det_pl"]:
                    continue  # skip insertion to prevent duplication

            #### SINGULAR ####
            if sym == "NP_sg" or sym == "VP_sg" or sym == "NP_conj_sg" or sym == "RC_sg" 
            or sym == "VP_sg" or sym == "VP_conj_sg" or sym == "PP_conj" or sym == "PP":
                where = symbols.index(sym)
                
                if sym == "NP_sg" or sym == "NP_conj_sg": # in this case, make NP singular
                    epsilon = random.uniform(0, 1)
                    if epsilon < 0.67:
                        epsilon2 = random.uniform(0, 1)
                        if epsilon2 < 0.5:
                            symbols = symbols[:where] + ["Det_sg", "Adj", "N_sg"] ...
                                        + symbols[where + 1:]
                            where += 2
                        else:
                            symbols = symbols[:where] + ["Det_sg", "N_sg"] + ...
                                        symbols[where + 1:]
                            where += 1
                    else:
                        symbols = symbols[:where] + ["ProperNoun_sg"] + symbols[where + 1:]
                    ### (**) for nested and cross-serial dependencies 
                    # (coordinated phrases) 
                    # check if the next symbol is a verb phrase, 
                    # if so, expand that now before continuing
                    if where + 1 < len(symbols):
                        if symbols[where + 1] == "Adv":
                            where += 1
                        if symbols[where + 1] == "VP_sg" or symbols[where + 1] == "VP":
                            if where + 2 < len(symbols):
                                symbols = symbols[:where + 1] + ["V_sg"] + ...
                                            symbols[where + 2:]
                            else:
                                symbols = symbols[:where + 1] + ["V_sg"] + ...
                                            symbols[where + 2:]

                if sym == "VP_sg":
                    symbols = symbols[:where] + ["V_sg"] + ...
                                symbols[where + 1:]

                if sym == "VP_conj_sg":
                    symbols = symbols[:where] + ["Conj", "V_sg"] + ...
                                symbols[where + 1:]

                if sym == "PP":
                    # PP --> P, Det_sg, N_sg
                    symbols = symbols[:where] + ["P", "Det_sg", "N_sg"] + ...
                                symbols[where + 1:]

                if sym == "PP_conj":
                    # PP_conj --> Conj PP, PP --> P, Det_sg, N_sg 
                    symbols = symbols[:where] + ["Conj", "P", "Det_sg", "N_sg"] ...
                                + symbols[where + 1:]

                if sym == "RC_sg":
                    symbols = symbols[:where] + ["RelPronoun", "V_sg"] + symbols[where + 1:]
                        
    

            ### PLURAL ###
            if sym == "NP_pl" or sym == "VP_pl" or sym == "NP_conj_pl" or sym == "RC_pl" 
            or sym == "VP_conj_pl":
                where = symbols.index(sym)
                
                if sym == "NP_pl" or sym == "NP_conj_pl":
                    epsilon = random.uniform(0, 1)
                    if epsilon < 0.5:
                        symbols = symbols[:where] + ["Det_pl", "Adj", "N_pl"] + ...
                                            symbols[where + 1:]
                        where += 2
                    else:
                        symbols = symbols[:where] + ["Det_pl", "N_pl"] + symbols[where + 1:]
                        where += 1
                    ### for nested and cross-serial dependencies (coordinated phrases) 
                    # check if the next symbol is a verb phrase, if so, 
                    # expand that now before continuing
                    if where + 1 < len(symbols):
                        if symbols[where + 1] == "Adv":
                            where += 1
                        if symbols[where + 1] == "VP_pl" or symbols[where + 1] == "VP":
                            if where + 2 < len(symbols):
                                symbols = symbols[:where + 1] + ["V_pl"] + ...
                                            symbols[where + 2:]
                            else:
                                symbols = symbols[:where + 1] + ["V_pl"] + ...
                                            symbols[where + 2:]

                if sym == "VP_pl":
                    symbols = symbols[:where] + ["V_pl"] + symbols[where + 1:]

                if sym == "VP_conj_pl":
                    symbols = symbols[:where] + ["Conj", "V_pl"] + symbols[where + 1:]

                if sym == "RC_pl":
                    symbols = symbols[:where] + ["RelPronoun", "V_pl"] + ...
                                symbols[where + 1:]
        non_lex = any(sym not in lexicon for sym in symbols)
    return symbols
\end{verbatim}
}

\subsection{Context-sensitive grammar: sentence generation code}

{\small
\begin{verbatim}
def generate_sentence(rules, lexicon, symbols=None, max_expansion_per_symbol=10, 
                        max_recursion_depth=1e5e):
    # Start symbol mapping from "S"
    start_expansions = rules["S"]
    start_expansion = tuple(start_expansions[0])
    initial_symbols = start_expansion

    # Initial expansion from context-sensitive production branches
    cs_rule = initial_symbols
    context_expansions = rules[initial_symbols]
    context_expansion = random.choice(context_expansions)
    context_symbols = context_expansion

    sentence = []
    for context_symbol in context_symbols:
        # initialize a shared expansion_counts dictionary 
        # (expansion_counts needs to be shared across different recursive calls) 
        # per context symbol
        
        expansion_counts = {}

        part = get_expansion([context_symbol], rules, lexicon, max_expansion_per_symbol, 
                            expansion_counts, max_recursion_depth)
        sentence.extend(part)

    for i, sym in enumerate(sentence):
        if sym in lexicon.keys():
            sentence[i] = random.choice(lexicon[sym])

    return sentence


def get_expansion(symbols, rules, lexicon, max_expansion_per_symbol,
                    expansion_counts = None,
                    max_recursion_depth=1e5, current_recursion_depth=0):
    if expansion_counts is None:
        expansion_counts = {}

    i = 0
    while i < len(symbols):
        sym = symbols[i]
        # to limit the total number of expansions of all symbols 
        # (limited futher from each symbols)
        
        if sym in rules:
            # Initialize count if symbol not yet expanded
            if sym not in expansion_counts:
                expansion_counts[sym] = 0
            
            if expansion_counts[sym] < max_expansion_per_symbol and
            current_recursion_depth < max_recursion_depth:
                expansion_counts[sym] += 1  # Increment the expansion count

                expansions = rules[sym]
                if not expansions:
                    i += 1
                    continue
                expansion = random.choice(expansions)
                
                if isinstance(expansion, str):
                    expansion = [expansion]
                elif expansion is None:
                    expansion = []
                elif not isinstance(expansion, list):
                    expansion = list(expansion)

                if expansion == []:
                    # remove that symbol
                    symbols = symbols[:i] + symbols[i+1:]
                else:
                    symbols = symbols[:i] + expansion + symbols[i+1:]

                    # recursively expand the newly added symbols
                    j = i
                    while j < i + len(expansion):
                        if symbols[j] in rules:
                            # recursively expand the symbol (`NoneType' errors can be 
                            # incurred if the while loop completes without triggering 
                            # any of the return statements inside the conditional blocks)
                            new_symbols = get_expansion(
                                [symbols[j]], rules, lexicon, 
                                max_expansion_per_symbol, expansion_counts, print_out,
                                max_recursion_depth, current_recursion_depth + 1
                            )
                            
                            if new_symbols is not None and new_symbols != []:
                                symbols = symbols[:j] + new_symbols + symbols[j+1:]
                            else:
                                # if recursion expansion returns empty, remove the symbol
                                symbols = symbols[:j] + symbols[j+1:]
                        j += 1
                    i += len(expansion)
            elif expansion_counts[sym] >= max_expansion_per_symbol:
                # enforce terminal expansion for the remaining symbols
                # while any are in the keys of the rules, since we are
                # mapping to the symbols in the lexicon which then always 
                # map to a terminal
                symbols = forced_terminal_expansion(symbols, lexicon, print_out)   
                return symbols  
            else:
                symbols = forced_terminal_expansion(symbols, lexicon, print_out)
                return symbols
        else:
            i += 1  

    return symbols
\end{verbatim}
}

Notice in the \verb|generate_sentence| function, in correspondence with the context-sensitive productions that enforce dependencies from the initial expansion of the start symbol \verb|S|, the context-sensitive generator takes into account both the expansion of the start symbol and then the choice of tense dependencies before recursively expanding each part of the selected sequence of context-sensitive symbols. To further adhere to a classic context-sensitive language, the canonical non-context-free languages $L = \{a^nb^nc^n \;|\; n \geq 1\}$ and $L = \{a^{n!} \;|\; n \geq 1 \}$, when depth or expansion limits are reached the current expansion and recursion depth parameters are specific (initialized) for each \verb|context_expansion| symbol.

The following provides the method of forcing a direct route to terminals for each symbol in the context-sensitive grammar. The relationship between the symbols is more extensive and so notice \verb|(**)| for a check of left-hand and right-hand agreement (requiring a \textit{look-ahead} before further expansions are checked) during the direct mapping to symbols that only lead to terminals:

{\small

\begin{verbatim}
def forced_terminal_expansion(symbols, lexicon, print_out=False):
    lexicon_keys = lexicon.keys()
    lexicon_values = lexicon.values()
    non_lex = True
    while non_lex:
        # if max expansion is reached, force terminals that agree with the rules
        for sym in symbols:
                # if max_expansion is minimal, then start from S --> NP_seq VP_seq
                # so, if sym == ["NP_sequence", "VP_placeholder"] then expand to either 
                # NP_sg, NP VP_sg or NP_pl NP VP_pl (let VP_seq = [] 
                # in this forced terminal case)
                
                if sym == ("NP_sequence", "VP_placeholder") or 
                sym == ["NP_sequence", "VP_placeholder"]: 
                # will be taken care of before this forced terminal section is 
                # reached, but incase the 
                initialization gets changed
                    where = symbols.index(sym)
                    epsilon = random.uniform(0, 1)
                    if epsilon < 0.5:
                        epsilon2 = random.uniform(0, 1)
                        epsilon2 = random.uniform(0, 1)
                        if epsilon2 < 0.5:
                            symbols = symbols[:where] + ["NP_sg", "NP_sg", "VP_sg"] + ...
                                        symbols[where + 1:]
                        else:
                            symbols = symbols[:where] + ["NP_sg", "NP_pl", "VP_sg"] + ...
                                        symbols[where + 1:]
                    elif epsilon >= 0.5:
                        epsilon2 = random.uniform(0, 1)
                        if epsilon2 < 0.5:
                            symbols = symbols[:where] + ["NP_pl", "NP_sg", "VP_pl"] + ...
                                        symbols[where + 1:]
                        else:
                            symbols = symbols[:where] + ["NP_pl", "NP_pl", "VP_pl"] + ...
                                        symbols[where + 1:]

                if sym == "NP_sequence": 
                    # (**) for context-dependence, if there is a "VP_sg" or a "VP_pl" 
                    # directly succeeding this symbol, change that first 
                    # so that the agreement is 
                    maintained before additional 
                    # expansions loses track of the corresponding verb phrase
                    
                    where = symbols.index(sym)
                    if where + 1 < len(symbols):
                        if symbols[where + 1] == "VP_sg":
                            if epsilon < 0.5:
                                if where + 2 < len(symbols):
                                    symbols = symbols[:where + 1] + ["V_sg"] + ...
                                                symbols[where + 2:]
                                else:
                                    symbols = symbols[:where + 1] + ["Adv", "V_sg"]
                        if symbols[where + 1] == "VP_pl":
                            if epsilon < 0.5:
                                if where + 2 < len(symbols):
                                    symbols = symbols[:where + 1] + ["V_pl"] + ...
                                                symbols[where + 2:]
                                else:
                                    symbols = symbols[:where + 1] + ["Adv", "V_pl"]

                    where = symbols.index(sym)
                    epsilon = random.uniform(0, 1)
                    if epsilon <= 0.5:
                        epsilon2 = random.uniform(0, 1)
                        if epsilon2 < 0.67:
                            epsilon3 = random.uniform(0, 1)
                            if epsilon3 < 0.5:
                                symbols = symbols[:where] + ["Det_sg", "Adj", "N_sg"] + ...
                                            symbols[where + 1:]
                            else:
                                symbols = symbols[:where] + ["Det_sg", "N_sg"] + ...
                                            symbols[where + 1:]
                        else:
                            symbols = symbols[:where] + ["ProperNoun_sg"] + ...
                                        symbols[where + 1:]
                    elif epsilon > 0.5:
                        if symbols[where - 1] == "N_pl":
                            symbols = symbols[:where] + ["Det_pl", "N_pl"] + ...
                                        symbols[where + 1:]
                        else:
                            symbols = symbols[:where] + ["Det_sg", "N_sg"] + ...
                                        symbols[where + 1:]


                #### SINGULAR ####
                if sym == "NP" or sym == "NP_sg" or sym == "VP_sg" or sym == "NPPrime_sg" 
                or sym == "VP_sequence" or sym == "VP_sg" or sym == "RC_sg":
                    # NP_seq --> NP , then have NP --> NP_sg , 
                    # then NP_sg --> Det N_sg, then select terminals of "Det" and "N"
                    
                    where = symbols.index(sym)
                    epsilon = random.uniform(0, 1)
                    if sym == "NP" or sym == "NP_sg": 
                    # in this case, we just make NP singular
                        if epsilon < 0.67:
                            epsilon2 = random.uniform(0, 1)
                            if epsilon2 < 0.5:
                                symbols = symbols[:where] + ["Det_sg", "Adj", "N_sg"] + ...
                                            symbols[where + 1:]
                            else:
                                symbols = symbols[:where] + ["Det_sg", "N_sg"] + ...
                                            symbols[where + 1:]
                        else:
                            symbols = symbols[:where] + ["ProperNoun_sg"] + ...
                                        symbols[where + 1:]

                    if sym == "VP_sequence" or  sym == "VP_sg":
                        symbols = symbols[:where] + ["V_sg"] + symbols[where + 1:]

                    if sym == "PP_conj":
                        # PP_conj --> conj PP
                        symbols = symbols[:where] + ["Conj", "P", "Det_sg", "N_sg"] + ...
                                    symbols[where + 1:]
                    
                    if sym == "NPPrime_sg":
                        # NPprime_sg --> PP , PP --> P, Det_sg, N_sg
                        symbols = symbols[:where] + ["P", "Det_sg", "N_sg"] + ...
                                    symbols[where + 1:]
                    
                    if sym == "RC_sg":
                        symbols = symbols[:where] + ["RelPronoun", "V_sg"] + ...
                                    symbols[where + 1:]
        

                ### PLURAL ###
                if sym == "NP_pl" or sym == "NPPrime_pl" or sym == "VP_pl" 
                or sym == "RC_pl":
                
                    where = symbols.index(sym)
                    if sym == "VP_pl":
                        # replace sym with "V_pl" or "Adv V_pl"
                        epsilon = random.uniform(0, 1)
                        if epsilon < 0.5:
                            symbols = symbols[:where] + ["Adv", "V_pl"] + ...
                                        symbols[where + 1:]
                        else: 
                            symbols = symbols[:where] + ["V_pl"] + ...
                                        symbols[where + 1:]

                    if sym == "NP_pl":
                        # NP_pl --> Det_pl N_pl or Det_pl Adj N_pl
                        epsilon = random.uniform(0, 1)
                        if epsilon < 0.5:
                            symbols = symbols[:where] + ["Det_pl", "Adj", "N_pl"] ...
                                        + symbols[where + 1:]
                        else:
                            symbols = symbols[:where] + ["Det_pl", "N_pl"] + ...
                                        symbols[where + 1:]

                    if sym == "NPPrime_pl":
                        # NPprime_pl --> PP , PP --> P, Det_pl, N_pl
                        symbols = symbols[:where] + ["P", "Det_pl", "N_pl"] + ...
                                    symbols[where + 1:]

                    if sym == "RC_pl":
                        symbols = symbols[:where] + ["RelPronoun", "V_pl"] + ...
                                    symbols[where + 1:]
    
                if sym == "PP" or sym == "PP_conj":
                    where = symbols.index(sym)
                    if sym == "PP_conj":
                        epsilon = random.uniform(0, 1)
                        if epsilon < 0.5:
                            # PP_conj --> Conj PP , pp --> P Det_pl N_pl
                            symbols = symbols[:where] + ["Conj", "P", "Det_pl", "N_pl"] ...
                                        + symbols[where + 1:]
                        else:
                            # PP_conj --> conj PP
                            where = symbols.index(sym)
                            symbols = symbols[:where] + ["Conj", "P", "Det_sg", "N_sg"] ...
                                        + symbols[where + 1:]
                    if sym == "PP":
                        epsilon = random.uniform(0, 1)
                        if epsilon < 0.5:
                            # PP --> P Det N_sg
                            symbols = symbols[:where] + ["P", "Det_pl", "N_pl"] + ...
                                        symbols[where + 1:]
                        else:
                            # PP --> P Det N_pl
                            symbols = symbols[:where] + ["P", "Det_sg", "N_sg"] + ...
                                        symbols[where + 1:]

        non_lex = any(sym not in lexicon for sym in symbols)
    return symbols

\end{verbatim}
}

\newpage
\section{Ongoing Errors}\label{appendix:ongoingerrors}

Language fluency and statistical pattern matching abilities are no small feat; however, this does not substitute for formal logic or adaptive reasoning abilities. In LMs, brittleness (e.g., inability to adapt to small deviations from memorized templates \cite{tamLetMeSpeak2024}; debilitating sensitivity to wording and phrasing \cite{McCoy2024llmsshapedbyproblem}), hallucinations \cite{huangSurveyHallucinationLarge2025, xuHallucinationInevitableInnate2024}, and weak generalizations, including the inability to perform on unfamiliar (even trivial) problems \cite{nezhurina2024alicewonderlandsimpletasks}, have been well documented and explored for years.

These shortfalls are indicative of the underlying computational class being one that does not support adaptive reasoning and novel applications of formal logic. The G-A classes do not provide a single model, but describe all models quantified with the computational abilities associated with that class. Recall that a model of a lower class is still made up of a FSM and can be broadened to be able to ``patch'' any mistakes caught in the model outputs; however, this patch does not solve the problem, as an infinite number of deviations of the kind that required the patch can be created. 

A canonical example seen often when ChatGPT (GPT-3) was first released to the public is the following: if you ask which is heavier, 10 kilos of steel or 1 kilo of feathers, the model would answer that they weigh the same. The response stems from the trick question ``which is heavier, 1 kilo of steel or 1 kilo of feathers?'' after inquiring about density, which is ubiquitous on the internet and, therefore, in the language model training data -- exemplifying the brittleness of simple pattern matching and the lack of abstraction required for generalizations and adaptive reasoning. 

\begin{figure}[htb]
    \centering
    \includegraphics[width=0.7\linewidth]{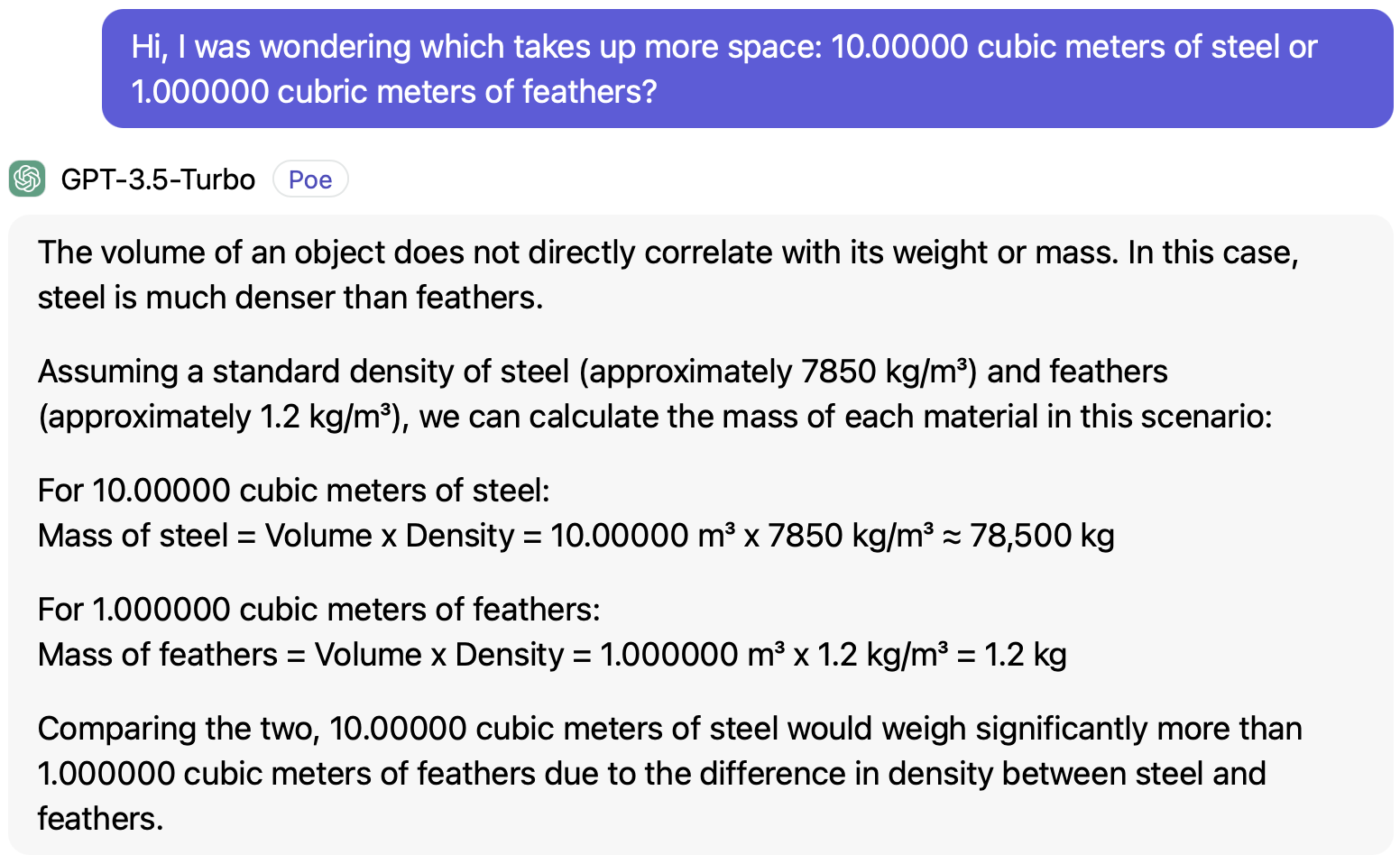}
    \caption{Screenshot 2025-03-11 at 19.55.57 of GPT-3.5-Turbo query and output in the Poe chatbot interface from 10 March 2025.}
    \label{fig:brittle-gpt}
\end{figure}

\begin{figure}[htb]
    \centering
    \includegraphics[width=0.7\linewidth]{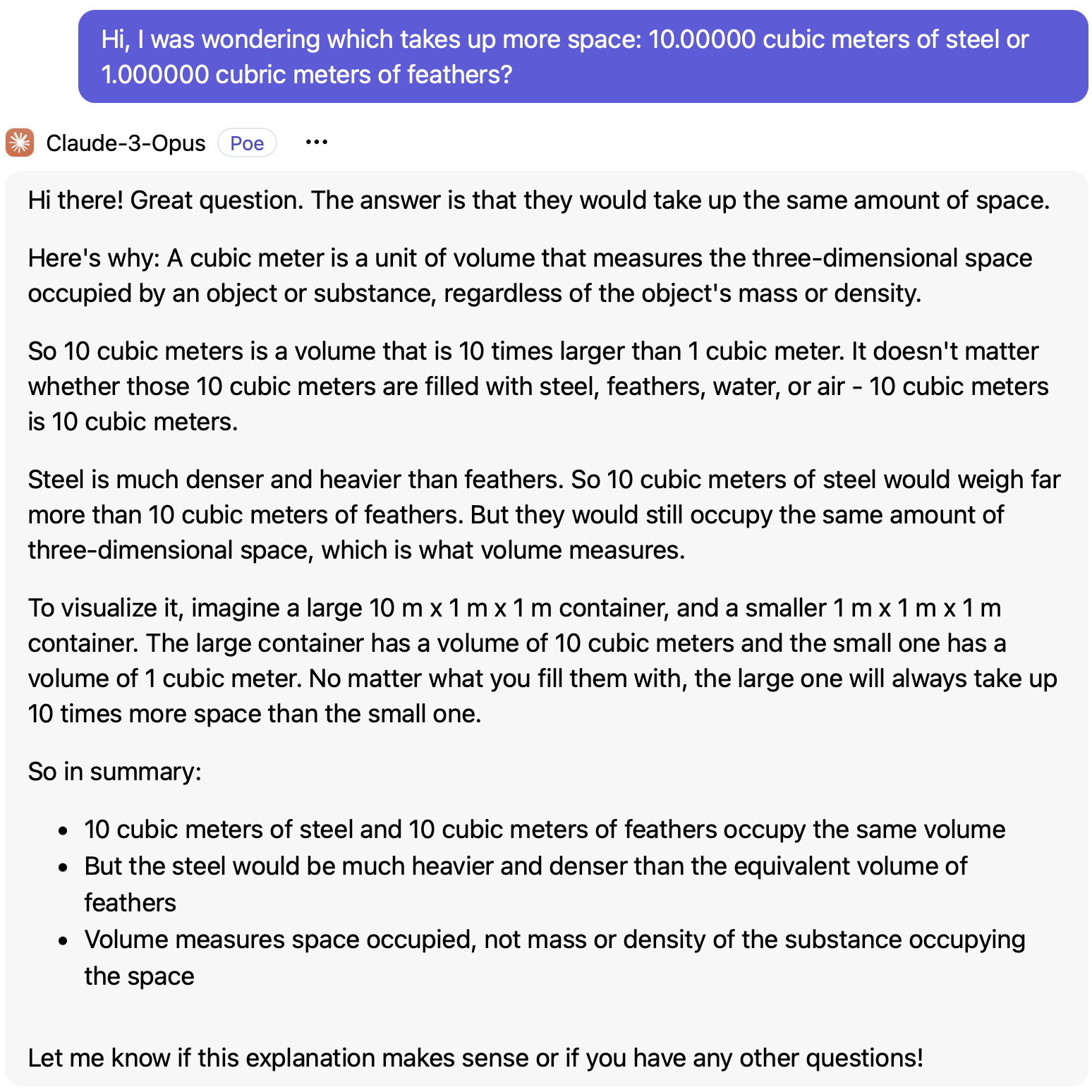}
    \caption{Screenshot 2025-03-11 at 19.55.35 of Claude 3 Opus query and output in the Poe chatbot interface from 10 March 2025.}
    \label{fig:brittle-claude}
\end{figure}

This problem was identified and patched in ChatGPT and will likely not appear in any models released since then, due to the developers of the models being aware of this query and being able to fine-tune or retroactively correct it with human feedback to avoid the mistaken response. To demonstrate that the patch is in no way a solution, we have decided to test a slight variant of the gist of this problem, and we see that both a small GPT LM, GPT-3.5-Turbo, and the largest LLM, Claude 3 Opus we tested (which was indicated to fall in the IXG/HOPDA tier), both were unable to answer correctly. See Figure \ref{fig:brittle-gpt} and Figure\ref{fig:brittle-claude} for the queried question and the models' respective outputs.

\newpage

\newpage
\section{Outside of the formal computational perspective}\label{appendix:phil}

Our empirical findings are purely at the computational level, not at the algorithmic \cite{MarrVisionComputationalInvestigation1982}. By this we mean that just because a model recognizes strings generated from grammars of a particular computational class,

A model's ability to process strings from grammars of a particular computational tier is agnostic with respect to any specific algorithmic method (e.g., the operational use of stacks, tapes, etc.)  No specific algorithm is implied, solely an equivalence class of such methods: i.e., any method(s) that can achieve the specified computational ability.

The formal characterizations of automata in the hierarchy are abstract control systems (finite state machines) combined with very specific idealized external memories (particular stack or tape organizations).  These define the formal system without reference to the myriad ways that such abstract engines could be specifically implemented (which may or may not actually include specific stack, queue, tape, or other mechanism).

We nonetheless remain extremely interested in understanding and specifying both biological circuitry (implementations) and mechanistic operational descriptions (algorithms) of processes that may underlie the computational abilities at given tiers.  The thalamocortical loop characterization (Section 5.4) is part of this endeavor, and the details of its operation are proposed as specific hypothesized mechanisms. The remainder of the present manuscript is focused specifically on the computation, not algorithm level.  We emphasize the strong distinction between the characterizations of computational powers of systems, versus the identification of specific mechanisms that may underlie these identified computational equivalence classes. Both endeavors are of interest.

\subsection{Work relating abstract machines at other analytical levels}

Section 4 presented empirical evaluations of zero-shot responses of already-trained language models: these identify the computational capacity of a system while remaining agnostic to any particular algorithms or mechanisms by which the systems' responses are produced.  

Related work in the field often includes investigations of algorithms involved in learning and producing sequences from grammars, including artificial grammars.  \citet{ferrignoHumansUsePushDown2025} compared two candidate memory architectures (pushdown stacks and queues) in humans who had been trained on center-embedded and cross-serial sequences, finding (via several evaluation methods) no evidence of the use of a last-in-first-out stacks but instead results consistent with first-in-first-out queues; this is an instance of direct testing of particular algorithm-level analyses, as opposed to our benchmark questions that ask which classes of generated strings are accepted by existing (trained) language models, without distinguishing among stacks, queues, content-addressable representations, or other potentially computationally (but not algorithmically) equivalent procedures.  

Terminological issues also arise.  The $A^n B^n$ grammar in \citet{ferrignoHumansUsePushDown2025} is indexed in the sense that corresponding $A_i$--$B_i$ base pairs are individually identified by subscripts but is not to be conflated with the formal class of indexed grammars in the grammar hierarchy (and is unrelated to the particular tree-adjoining grammar work developed here in Section 3 and used for empirical LLM evaluations in Section 4.  In general, even where a stack is a method sufficient for, or formally characteristic of, a grammar class, other memory organizations may be responsible for task responses in any particular algorithm for the given grammar class.  

The empirical evaluations in this work (Section 4) focuses on evaluating the zero-shot responses of already-trained language models (language foundation models), not on the training process itself. Moreover, our results concern computational capacity rather than the particular algorithms or physical mechanisms by which the models produce their responses.

\subsection{Forms of learning}

Many open distinctions exist among different forms of learning, such as the well-studied distinction between learning facts vs. learning skills, which engage doubly dissociable brain mechanisms \cite{cohenMemoryItemsMemory1997, eichenbaumMemorySystems2010}.  Open questions remain about the relationships between natural language acquisition and the learning of artificial languages.  Evidence exists that knowledge can be implicitly acquired through artificial-language training \cite{dienesTheoryImplicitExplicit1999, rebuschatMeasuringImplicitExplicit2013, Ellis2022}; such findings are consistent with a range of possible mechanisms, and may not resolve the differences that may exist between artificial and natural language learning.  Perhaps most importantly, natural language acquisition is ubiquitous among all neurologically intact humans, in contrast to the substantial variability of most learned human abilities.

Pre-transformer artificial systems (such as RNNs and LSTMs) exhibited substantial language abilities including machine translation, but to group them together with transformers under the broad label of "artificial neural networks" is to obscure highly notable distinctions, and does not explain the marked increase in language performance that arose with the introduction of the novel transformer architecture \cite{vaswani2017}.  Transformers and previous architectures are not easily collapsed into a single category while retaining explanatory power.

\subsection{Program induction from formal language strings}

\citet{yangOneModelLearning2022} address grammar induction via a Bayesian program-induction model that searches over compositions of domain-general primitives, to infer programs that generate observed strings.  The model learns examples of regular, context-free, and context-sensitive languages from positive evidence, often from relatively few strings.  Despite being described as ``one model for the learning of language'', the system should not be confused with the architectures or operations of the broad range of extant transformer-based language models, nor any model trained over a natural-language corpus; it is, rather, an idealized model of induction over formal symbol sequences.  Their goal is not to reproduce broad linguistic performance per se, but to test whether positive evidence can, in principle, support the induction of generative computations without dependence on a narrowly language-specific hypothesis space.  The program-induction framework makes candidate rules explicit and permits direct comparison of the descriptions inferred for different formal languages.  A contemporary large language model would instead introduce many additional variables and, perhaps notably, produces fluency in natural languages.  The present work is different in kind: to evaluate the empirical computational capabilities of models fluent in natural languages, as opposed to the use of an induction model to infer grammars that generate strings.

\subsection{A note on the computational perspective of natural phenomena}
It is also worth briefly noting that our approach is concordant with extensive work in ``natural computation'' and ``biological computation,'' in which natural phenomena are understood in terms of computational analyses (see, e.g., \citet{mitchellUbiquitySymposiumBiological2011, denningComputingNaturalScience2007, chuComputationNaturalSystems2018}, as well as \citet{dennettBrainstormsPhilosophicalEssays1986} and \citet{laforteWhyGodelsTheorem1998} refutations of claims from \citet{penroseEmperorsNewMind1989} on computationalism and computational tractability.

\newpage

\section{Parse} \label{appendix:parse}

As described in supplementary  Section~\ref{appendix:experiments}, the probes of the tested language systems were in the form of intentionally simple (and relatively inexpensive) binary yes/no queries asking whether or not a given sentence was grammatical, without any incentives or objectives associated with their responses.

Questions especially arise as a function of the difficulty of the data: are the most challenging sets of sentences actually being successfully parsed? 
In particular, were there any cases in which the system  responded ``yes'' 
but was not in fact parsing the sentence? To further examine this issue, we ran combined tests of  a recently released language model that we would categorize as an ATT system: GPT 5.5, asking the system first to produce a binary yes/no response to given sentences,  and then to produce an actual proposed parse of the sentence. Thus we can check that a ``yes'' response aligns with the production of a parse whereas a ``no'' aligns with an attempted but failed parse.   
  
\subsection{Model configuration and queries}
The GPT 5.5 model evaluated is the April 23, 2026 snapshot: \texttt{gpt-5.5-2026-04-23}. The model was configured with \texttt{system message} ``You are a helpful assistant. Follow the user's instructions carefully and provide accurate, relevant responses in the requested format.'', \texttt{reasoning} of \texttt{effort: high}, \texttt{verbosity} set to \texttt{low}, and the temperature set to $0.2$.

First, the original prompt was reused:
\begin{quote}
    \it
    Disregarding any lack of punctuation or sensicality, please answer the following with only ``Yes'' or ``No'': If the following sentence grammatically complete?
\end{quote}

Next, the model was prompted with a follow-up query requesting a parse of the string:
\begin{quote}
        \it
        Parse the sentence exactly as written. Use a bracketed format such as:
        \begin{center}
            $[\text{S}\; [\text{NP} \,\dots] \, [\text{VP} \,\dots]\,]$
        \end{center}
        Return only the bracketed parse.
\end{quote}

The follow-up query was accompanied by the following input into the \texttt{instructions} GPT 5.5 configuration:
\begin{quote}
    \it Parse the sentence exactly as written without correcting, rewriting, or normalizing it. The grammaticality was determined by your previous response. If it is ungrammatical, provide the best bracketed parse possible and mark any ill-formed or unattached material with an ERR label. Return only the bracketed parse, without Markdown, explanation, or commentary.
\end{quote}

\subsection{Parse responses}

We first show a valid parse tree of a sample string,  in terms of the grammar that was used here to generate the string (Figures A.\ref{fig:G_ix-parse-string}, A.\ref{fig:G_cs-parse-string-6}, and A.\ref{fig:G_cs-parse-string-36}).  

We then show corresponding parse trees of GPT-5.5 responses to three strings: 
an IXG-generated string (see Figure\ref{fig:G_ix-parse-string-GPT}), a CSG-generated string that was accepted by the model (see Figure\ref{fig:G_cs-parse-string-GPT5.5-accept}), and a CSG-generated string that was not accepted by the model (Figure\ref{fig:G_cs-parse-string-GPT5.5-reject}).


\newpage
\begin{figure}[p]
    \centering
    \vspace{2cm}
    \begin{adjustbox}{max width=\linewidth}
    \begin{forest} parse
    [S
      [$\dot{\text{NP}}$
        [$\dot{\text{PN}}$ [Leibniz]]
        [$\dot{\text{RC}}$
          [RP [who]]
          [$\dot{\text{VP}}$
            [$\dot{\text{V}}$ [invites]]
            [NP
              [$\dot{\text{NP}}$
                [$\dot{\text{PN}}$ [Riemann]]
                [$\dot{\text{RC}}$
                  [RP [who]]
                  [$\dot{\text{VP}}$
                    [$\dot{\text{V}}$ [conjectures]]
                    [NP
                      [$\dot{\text{NP}}$
                        [$\dot{\text{PN}}$ [Galileo]]
                        [$\dot{\text{RC}}$
                          [RP [who]]
                          [$\dot{\text{VP}}$
                            [$\dot{\text{V}}$ [rests]]
                          ]
                        ]
                      ]
                    ]
                  ]
                ]
              ]
            ]
          ]
        ]
      ]
      [$\dot{\text{VP}}$
        [$\dot{\text{V}}$ [solves]]
        [NP
          [$\ddot{\text{NP}}$
            [$\ddot{\text{Det}}$ [the]]
            [$\ddot{\text{N}}$ [mountains]]
            [$\ddot{\text{RC}}$
              [RP [that]]
              [$\ddot{\text{VP}}$
                [$\ddot{\text{V}}$ [eat]]
                [$\ddot{\text{VP}}_{cnj}$
                  [Cnj [or]]
                  [$\ddot{\text{VP}}$
                    [$\ddot{\text{V}}$ [fall]]
                  ]
                ]
              ]
            ]
          ]
        ]
      ]
    ]
    \end{forest}
    \end{adjustbox}
    \caption{A valid parse of the following string, in terms of its generator $G_{ix}$ (defined in Section~\ref{appendix:defgrammars}): \\ \textit{Leibniz who invites Riemann who conjectures Galileo who rests solves the mountains that eat or fall.}}
    \label{fig:G_ix-parse-string}
\end{figure}

\begin{figure}[p]
    \centering
    \vspace{2cm}
        \begin{adjustbox}{max width=\linewidth}
        \begin{forest} parse
        [S
          [NP
            [N [Leibniz]]
            [CP [WHNP [who]]
              [S [VP [V [invites]]
                [NP [N [Riemann]]
                  [CP [WHNP [who]]
                    [S [VP [V [conjectures]]
                      [NP [N [Galileo]]
                        [CP [WHNP [who]]
                          [S [VP [V [rests]]]]
                        ]
                      ]
                    ]]
                  ]
                ]
              ]]
            ]
          ]
          [VP [V [solves]]
            [NP [Det [the]] [N [mountains]]
              [CP [WHNP [that]]
                [S [VP [VP [V [eat]]] [Conj [or]] [VP [V [fall]]]]]
              ]
            ]
          ]
        ]
        \end{forest}
        \end{adjustbox}
    \caption{GPT-5.5 parse response for the string: \\ \textit{Leibniz who invites Riemann who conjectures Galileo who rests solves the mountains that eat or fall.}}
    \label{fig:G_ix-parse-string-GPT}
\end{figure}

\begin{figure}[h]
    \centering
    \vspace{2cm}
    \begin{adjustbox}{max width=\linewidth}
        \begin{forest} parse
        [S
          [$\text{NP}_{\text{seq}}$
            [$\dot{\text{NP}}$
              [$\dot{\text{Det}}$ [a]]
              [Adj [bright]]
              [$\dot{\text{N}}$ [tree]]
              [PP
                [P [by]]
                [NP
                  [$\ddot{\text{NP}}$
                    [$\ddot{\text{Det}}$ [the]]
                    [$\ddot{\text{N}}$ [equations]]
                    [$\ddot{\text{RC}}$
                      [RP [who]]
                      [$\ddot{\text{VP}}$
                        [Adv [happily]]
                        [$\ddot{\text{V}}$ [fade]]
                      ]
                    ]
                  ]
                ]
              ]
            ]
            [$\text{NP}_{\text{seq}}$
              [NP
                [$\ddot{\text{NP}}$
                  [$\ddot{\text{Det}}$ [the]]
                  [$\ddot{\text{N}}$ [rooms]]
                ]
              ]
            ]
          ]
          [$\text{VP}_{\text{placeholder}}$
            [$\dot{\text{VP}}$
              [Adv [brilliantly]]
              [$\dot{\text{V}}$ [conjectures]]
            ]
            [$\text{VP}_{\text{seq}}$
              [$\ddot{\text{VP}}$
                [$\ddot{\text{V}}$ [study]]
              ]
            ]
          ]
        ]
        \end{forest}
    \end{adjustbox}
    \caption{A valid parse of the following string, in terms of its generator $G_{cs}$ (defined in Section~\ref{appendix:defgrammars}): \\ \textit{A bright tree by the equations who happily fade the rooms brilliantly conjectures study.}}
    \label{fig:G_cs-parse-string-36}
\end{figure}

\begin{figure}[h]
    \centering
    \vspace{2cm}
    \begin{adjustbox}{max width=\linewidth}
    \begin{forest} parse
    [S
      [NP
        [Det [A]]
        [Adj [bright]]
        [N [tree]]
        [PP
          [P [by]]
          [NP
            [Det [the]]
            [N [equations]]
            [CP
              [Rel [who]]
              [VP
                [Adv [happily]]
                [V [fade]]
                [NP
                  [Det [the]]
                  [N [rooms]]
                ]
                [Adv [brilliantly]]
              ]
            ]
          ]
        ]
      ]
      [VP
        [V [conjectures]]
        [NP
          [N [study]]
        ]
      ]
      [. [.]]
    ]
    \end{forest}
    \end{adjustbox}
    \caption{GPT-5.5 parse response for the string: \\ \textit{A bright tree by the equations who happily fade the rooms brilliantly conjectures study.}}
    \label{fig:G_cs-parse-string-GPT5.5-accept}
\end{figure}
\newpage

\begin{figure}[h]
    \centering
    \vspace{2cm}
    \begin{adjustbox}{max width=\linewidth}
    \begin{forest} parse
    [S
      [$\text{NP}_{\text{seq}}$
        [$\dot{\text{NP}}$
          [$\dot{\text{Det}}$ [the]]
          [$\dot{\text{N}}$ [bird]]
        ]
        [$\text{NP}_{\text{seq}}$
          [NP
            [$\dot{\text{NP}}$
              [$\dot{\text{Det}}$ [the]]
              [$\dot{\text{N}}$ [bird]]
              [PP
                [P [about]]
                [NP
                  [$\ddot{\text{NP}}$
                    [$\ddot{\text{Det}}$ [the]]
                    [$\ddot{\text{N}}$ [rooms]]
                    [PP
                      [P [by]]
                      [NP
                        [$\ddot{\text{NP}}$
                          [$\ddot{\text{Det}}$ [the]]
                          [Adj [elegant]]
                          [$\ddot{\text{N}}$ [theorems]]
                          [PP
                            [P [by]]
                            [NP
                              [$\dot{\text{NP}}$
                                [$\dot{\text{PN}}$[Ramanujan]]
                              ]
                            ]
                          ]
                        ]
                      ]
                    ]
                  ]
                ]
              ]
            ]
          ]
        ]
      ]
      [$\text{VP}_{\text{placeholder}}$
        [$\dot{\text{VP}}$
          [$\dot{\text{V}}$[saves]]
        ]
        [$\text{VP}_{\text{seq}}$
          [$\dot{\text{VP}}$
            [$\dot{\text{V}}$ [eat]]
            [NP
              [$\dot{\text{NP}}$
                [$\dot{\text{PN}}$ [Riemann]]
              ]
            ]
          ]
        ]
      ]
    ]
    \end{forest}
    \end{adjustbox}
    \caption{A valid parse of the following string, in terms of its generator $G_{cs}$ (defined in Section~\ref{appendix:defgrammars}): \\ \textit{The bird the bird about the rooms by the elegant theorems by Ramanujan saves eat Riemann.}}
    \label{fig:G_cs-parse-string-6}
\end{figure}

\begin{figure}[h]
    \centering
    \vspace{2cm}
    \begin{adjustbox}{max width=\linewidth}
    \begin{forest} parse
    [S
      [NP
        [Det [The]]
        [N [bird]]
        [CP
          [NP
            [Det [the]]
            [N [bird]]
            [PP
              [P [about]]
              [NP
                [Det [the]]
                [N [rooms]]
                [PP
                  [P [by]]
                  [NP
                    [Det [the]]
                    [Adj [elegant]]
                    [N [theorems]]
                    [PP
                      [P [by]]
                      [NP
                        [N [Ramanujan]]
                      ]
                    ]
                  ]
                ]
              ]
            ]
          ]
          [VP
            [V [saves]]
          ]
        ]
      ]
      [VP
        [{\color{red}{ERR}}
          [V [eat]]
        ]
        [NP
          [N [Riemann]]
        ]
      ]
    ]
    \end{forest}
    \end{adjustbox}
    \caption{GPT-5.5 parse response for the string, where ``ERR'' (\textit{red}) was the instructed symbol to mark any ``ill-formed or unattached material'': \\ \textit{The bird the bird about the rooms by the elegant theorems by Ramanujan saves eat Riemann.}}
    \label{fig:G_cs-parse-string-GPT5.5-reject}
\end{figure}

\clearpage
\bibliographystyle{ACM-Reference-Format}
\bibliography{bibfile}

\end{document}